\pdfoutput=1
\documentclass[11pt]{article}

\usepackage[final]{ACL2023}

\usepackage{times}
\usepackage{latexsym}
\usepackage{graphicx}
\usepackage{booktabs}
\usepackage{setspace}
\usepackage{amsmath}
\usepackage[T1]{fontenc}
\usepackage{changepage}

\usepackage[utf8]{inputenc}

\usepackage{microtype}

\usepackage{inconsolata}
\usepackage{CJKutf8}
\usepackage{subcaption}
\usepackage{tabularx}
\usepackage{makecell}
\usepackage{multirow}
\usepackage{fancyvrb}

\title{Deciphering Digital Detectives: Understanding LLM Behaviors and Capabilities in Multi-Agent Mystery Games}
\author{Dekun Wu\thanks{\ \ Equal contribution.}\ , \ Haochen Shi\footnotemark[1] \ , \ Zhiyuan Sun \and Bang Liu\thanks{\ \ Canada CIFAR AI Chair.}\\
        Université de Montréal \& Mila - Quebec AI Institute \\ 
        \texttt{\{dekun.wu, haochen.shi, zhiyuan.sun, bang.liu\}@umontreal.ca} \\ }
\usepackage{tikz} % For rounded corners
\usepackage{colortbl} % For coloured rows
\usepackage{array}
\usepackage{caption}

\begin{document}
\maketitle
\begin{abstract}
% The development of reasoning abilities in Large Language Models (LLMs) remains a critical challenge in the field of Artificial Intelligence. This paper addresses this challenge by introducing a novel application of LLMs in the context of Jubensha (\begin{CJK}{UTF8}{gkai}剧本杀\end{CJK}), a popular Chinese murder mystery role-playing game. This research is motivated by the intricate social interactions and complex reasoning required in these games, offering a unique platform to evaluate and enhance the cognitive capabilities of LLMs. 
% Our novel contributions include the creation of a specialized dataset for Jubensha games, designed specifically for AI agents. This dataset not only serves as a unique resource but also establishes a new scenario for assessing LLM performance in complex detective games. Additionally, we have implemented a comprehensive framework allowing for multi-agent interaction within the Jubensha game environment. This framework facilitates autonomous participation of LLM agents, advancing the scope of AI in interactive and strategic gaming contexts. 
% Extensive experiments reveal that LLMs can significantly benefit from this immersive and challenging setting. The study showcases enhanced reasoning and information processing capabilities of LLMs within the Jubensha games. These observations provide valuable insights into the potential of LLMs in socially and cognitively demanding environments, offering a promising direction for future research in AI-driven interactive gaming and complex problem-solving.~\footnote{Preprint. This work is in progress.}
In this study, we explore the application of Large Language Models (LLMs) in \textit{Jubensha}, a Chinese detective role-playing game and a novel area in Artificial Intelligence (AI) driven gaming. We introduce the first dataset specifically for Jubensha, including character scripts and game rules, to foster AI agent development in this complex narrative environment. Our work also presents a unique multi-agent interaction framework using LLMs, allowing AI agents to autonomously engage in this game. To evaluate the gaming performance of these AI agents, we developed novel methods measuring their mastery of case information and reasoning skills. Furthermore, we incorporated the latest advancements in in-context learning to improve the agents' performance in information gathering, murderer identification, and logical reasoning. The experimental results validate the effectiveness of our proposed methods. This work aims to offer a novel perspective on understanding LLM capabilities and establish a new benchmark for evaluating large language model-based agents.

% This comprehensive approach represents a significant advancement in AI applications for narrative-based gaming and sets a new standard for AI research in interactive role-playing games.~\footnote{Preprint. This work is in progress.}
\end{abstract}

\section{Introduction}
\begin{CJK}{UTF8}{gkai}

\begin{figure}
    \centering
    \resizebox{0.45\textwidth}{!}{\includegraphics{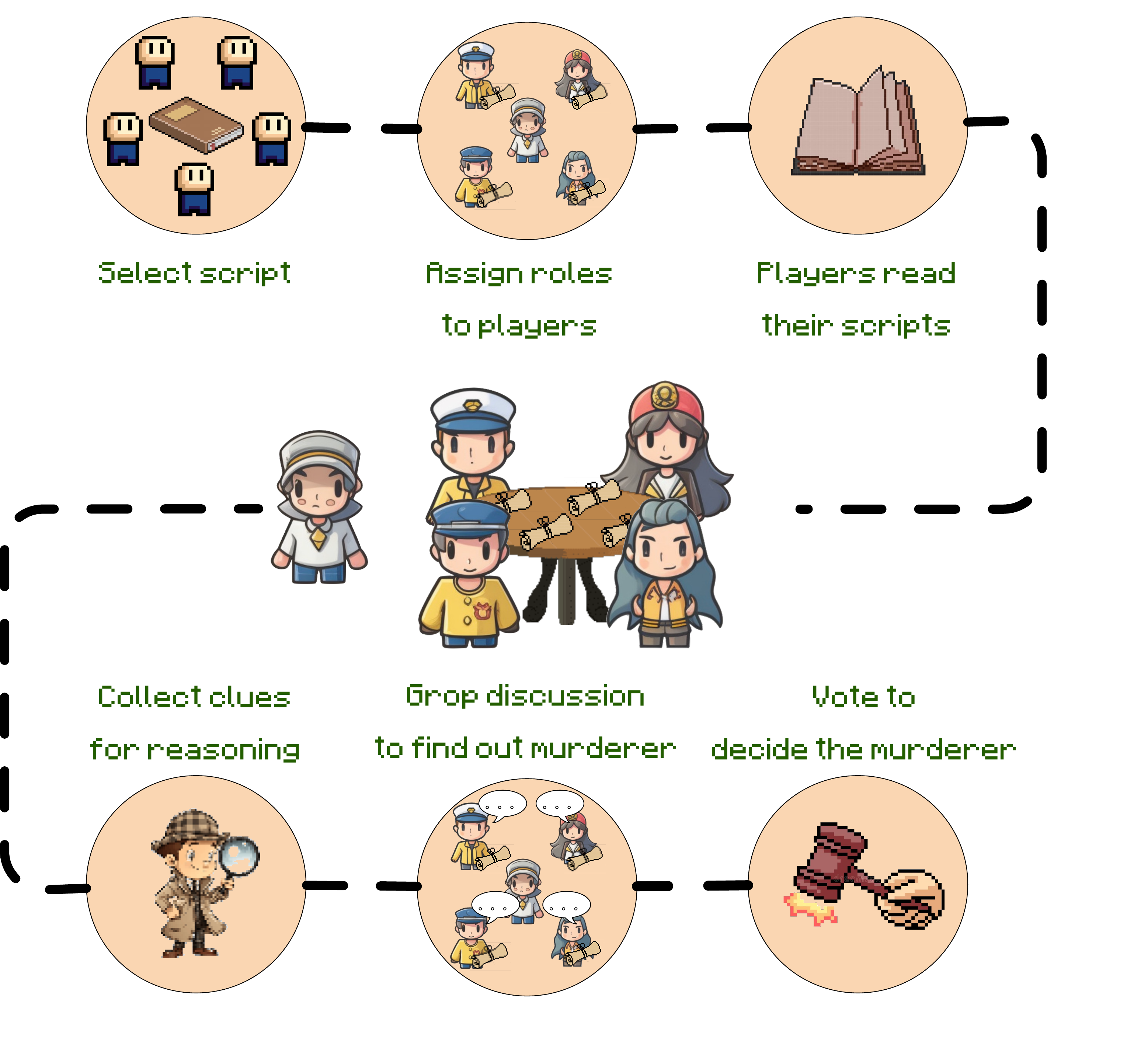}}
    \caption{Illustration of the Jubensha game. It requires players to interact with each other and reason about who is the murderer in a story.}
    \vspace{-4mm}
    \label{fig:background}
\end{figure}

% Third, to quantitatively assess the performance of LLM-based agents in Jubensha games, we designed two tasks to evaluate the ability of agents to acquire information and reason with the collected information in the game. Based on the experimental results, we analyze the capabilities of LLMs in handling the intricacies of Jubensha games. These components collectively offer a comprehensive approach to understanding LLMs in multifaceted problem-solving scenarios.
% Furthermore, we delve deeper into different modules of our framework to evaluate its effectiveness in improving the reasoning ability of LLM-based agents.

% Our study's novelty lies in several aspects: We are the pioneers in proposing Multi-agent Mystery Games as a means of evaluating LLMs. The dataset and framework we have developed are first of their kind, specifically tailored for this purpose. The analysis of our experiments offers new insights into the capabilities of LLMs in complex, collaborative environments. Our work has the potential to extend the application of LLMs to assist in real-world tasks, such as criminal investigations, thereby bridging a significant gap between AI research and practical, societal needs.

Interactive role-playing games, where players unravel mysteries through strategic interactions and clue-based puzzles, have seen a significant rise in global popularity. Tracing their origins to 19th-century murder mystery novels and early 20th-century party puzzles, these games have particularly flourished in China in recent years. In this context, they are referred to as "Jubensha", "Scripted  Murders" or 剧本杀 in Chinese, and have experienced a remarkable surge in popularity since the late 2010s. Influenced by Western murder mystery games, Jubensha games center around players who gather to identify the murderer by analyzing and interpreting the provided story scripts. The thrill for players to collect and interpret clues through intense social interaction and reasoning has made playing Jubensha stand out as a notable cultural phenomenon. 

% These games, blending immersive role-playing, engaging narrative storytelling, and thoughtful strategic planning, have captivated a significant following among young Chinese audiences. The requirement for players to collect and interpret clues through intense social interaction and reasoning has made playing Jubensha stand out as a notable cultural phenomenon. 

Meanwhile, the field of Artificial Intelligence (AI) is reshaping the gaming landscape. From classic games like chess~\cite{CAMPBELL200257}, Go~\cite{SilverHuangEtAl16nature}, and poker~\cite{ai_poker,brown2020combining} to video games like StarCraft~\cite{vinyals2017starcraft,vinyals2019grandmaster}, League of Legends~\cite{10.5555/3505464.3505511}, and Honor of Kings~\cite{10.5555/3495724.3495777}, AI contenders or collaborators have been integrated into these games. The recent surge in Large Language Models (LLMs) has redirected research interest from conventional and video games to text-based games. Significant examples include an LLM deployed as agents in the strategic game CICERO~\cite{meta2022human}, in the communicative game Werewolf~\cite{xu2023exploring}, and in the text-based adventure game Zork~\cite{tsai2023large}.

However, the ``Jubensha game'' remains an undeveloped field in terms of AI agents specifically tailored for its gameplay and evaluation. We believe there are several reasons. First, before the emergence of LLMs, understanding the character plots, role tasks, and game rules in the script of Jubensha games was very challenging. Not to mention the need for AI agents to engage in multi-round linguistic interactions, information gathering, and logical reasoning. Second, currently there is no publicly available dataset specifically for Jubensha games that researchers can use to develop and evaluate their agents. Third, quantitatively and qualitatively automating the evaluation of AI agents' performance in Jubensha games is also very difficult. The most commonly used win-rate metric in other game AIs is also of limited use in assessing the performance of Jubensha game agents, i.e., win rate does not show the extent of the Jubensha AI's mastery of case information or the level of its reasoning ability.

In response to the above-mentioned challenges, we attempt to provide a solution in this work. This work focuses on constructing a multi-agent interaction framework in a ``Jubensha'' game environment using large language models, and we have designed a set of methods to quantitatively and qualitatively assess the performance of Large Language Model based~(LLM-based) agents. Our contributions can be summarized as follows:

\begin{enumerate}
    \item We have created a Chinese dataset providing character scripts and preset game rules to initiate a Jubensha game. To the best of our knowledge, this is the first dataset specifically tailored for AI agents playing in a Jubensha game setting.
    % but also one of the few Chinese text-based game datasets. Our vision is to continuously expand this dataset to make it a good benchmark for measuring the capabilities of large language model agents.
    \vspace{-0.5mm}
    \item We designed a framework for multi-agent interaction in a Jubensha game environment using large language models, allowing multiple LLM-based agents to autonomously interact with each other in a Jubensha game setting, without the need for human intervention.
    \vspace{-0.5mm}
    \item To quantitatively and qualitatively assess the performance of large language model based agents in Jubensha games, we designed two novel tasks: one to evaluate their mastery of case information and another to assess their reasoning abilities with the information collected during the game.
    \vspace{-0.5mm}
    \item Utilizing the latest advancements in the field of in-context learning, we devised modules to enhance the performance of LLM-based agents. Our evaluations show that this design significantly enhances the information mastery, murderer identification, and reasoning capabilities of LLM-based agents in the Jubensha game context.
\end{enumerate}
\end{CJK}

\section{Related Work}
\paragraph{Interactive Role-Playing Games}
IRPGs provide immersive experiences in fictional settings, serving as a multidisciplinary research testbed in various fields~\cite{zagal2018definitions, barreteau2003role,nagata-etal-2021-shared}. These games are categorized into SRPGs and MRPGs, each with unique research opportunities.
\textit{SRPGs} focus on character advancement through quests in various themes~\cite{IFDB, cote2019textworld, szot2021habitat, fan2022minedojo, shridhar2020alfworld, wang2022scienceworld}. While \textit{MRPGs} highlight collaborative storytelling~\cite{lai2023werewolf, park2023generative, fu2023improving,kramar2022negotiation, meta2022human}. This work presents a dataset for MRPG Jubensha, aimed at supporting studies on the development and evaluation of communicative and reasoning AI in an adversarial game setting.
\paragraph{LLM-based Autonomous Agents.} 
Thanks to their strong capabilities in language comprehension and reasoning, LLMs have demonstrated significant potential for achieving impressive performance in tasks that focus on advancing characters.~\cite{wang2023survey, wei2022chain, wang2022self, yao2023tree, yao2022react, shinn2023reflexion}. However, only a few works explore the multi-agent interaction under complex settings~\cite{xu2023exploring,junprung2023exploring,lai2023werewolf}. To bridge the gap, we propose the first LLM-based multi-agent interaction framework on Jubensha, allowing autonomous agent engaging and facilitating further study on advanced study topics.

% The integration of large language models (LLMs) in IRPGs has shown promise for achieving human-like intelligence. These models, drawing from extensive web knowledge, are increasingly being explored as autonomous agents in various settings, such as adventure, scientific exploration, and householding tasks~\cite{wang2023survey, wei2022chain, wang2022self, yao2023tree, yao2022react, shinn2023reflexion}. 
% It is worth noting that~\cite{junprung2023exploring} explores one-to-many simulation of LLM-based agents in an extremely simplified murder mystery game. Our work differs from it in several ways: 1). \textit{The Setting of Tasks:} We model the Jubensha game as many-to-many simulation instead of simple one-to-many simulation; 2). \textit{Evaluation Method:} We construct a systemic method to evaluate the 

\paragraph{Evaluating LLM-based Agents.} Despite the quick development of LLM-based agents, assessing their abilities in MRPGs can still be essential and challenging. Recent initiatives~\cite{wang2023rolellm, shao2023character, liang2023tachikuma, xu2023exploring} propose several simple subjective and objective evaluation metrics for LLM-based agents under multi-role settings. However, it can still be challenging to assess the capabilities of LLM-based agents for MRPGs using simple evaluation metrics, such as win rate. To address this issue, we propose a set of systematic and objective evaluation methods designed to measure the information gathering and reasoning abilities of LLM-based agents in Jubensha.

\section{Jubensha Dataset}

% \begin{table*}[]
% \begin{tabular}{ccccccccc}
% \hline
%               & \multicolumn{3}{l}{\# Players} & \multicolumn{4}{l}{Involved Modalities} & Total                 \\ \hline
%               & 4-        & 5-8        & 9+        & T    & I    & A    & V   & \multirow{2}{*}{1115} \\ \cline{1-8}
% \# Games & 61        & 986        & 68        & 1115    & 630      & 339      & 146     &                       \\ \hline
% \end{tabular}
% \vspace{-4mm}
% \label{tab:dataset_statics1}
% \end{table*}

\begin{table}[h]
\centering
\begin{tabular}{lccccc}
\hline
       & Text & Image & Audio & Video & Total \\ \hline
\#  & 1115 & 643   & 392  & 172 & 1115   \\ \hline
\end{tabular}
\caption{Number of Jubensha game scripts by modality in our dataset.}
\label{tab:horizontal_dataset_statistics}
\end{table}
\vspace{-5mm}
\begin{table}[ht]
\begin{tabular}{ccccccc}
\hline
              & \multicolumn{3}{c}{Players} & \multicolumn{3}{c}{Tokens}                  \\ \hline
              & min        & max        & avg        & min    & max    & avg    \\ \cline{1-7}
\# &    1     &     20    &    6.52     & 4k    & 518k      & 129k                                 \\ \hline
\end{tabular}
\caption{Statics on number of players and tokens for the Jubensha game scripts in our dataset.}
\label{tab:dataset_statics1}
\vspace{-2mm}
\end{table}
To enhance AI deployment in Jubensha games, we have compiled a comprehensive dataset from over 1,100 Jubensha game instances in Chinese.\footnote{Currently, this dataset is in Chinese, but we are open to expanding it to other languages in the future. We use English examples in the main text for the convenience of the readers. } This dataset is a valuable addition to MRPG research,  presenting unique challenges and opportunities for advancement.~\footnote{We will release this dataset post-acceptance for academic purposes only.}

\paragraph{Background of Jubensha Game}
Jubensha is a detective role-playing game with multiple players, where each player is assigned a unique role tied to a central murder mystery. As shown in Figure~\ref{fig:background}, the game process typically consists of six stages: 1) Each player selects a script for distinct characters in a Jubensha game. 2) Players are assigned with a role (murderer or civilian) associated with their selected scripts. 3) The players read their scripts to develop a basic understanding of the whole story from their views. 4) Each player is given a pool of clues 
% collect ``most helpful'' clues from a clue pool with a budget 
to help them reveal or hide critical details for finding the murderer. 5) Several rounds of group discussion are held among the players to share information and find out the murderer. 6). Finally, each player anonymously votes to decide the murderer. The civilians win the game if the true murderer gets the most votes, otherwise, the murderer wins.

\paragraph{Dataset Construction}
To establish an environment capable of evaluating Jubensha agents and to facilitate future scaled-up works, we collect 1,115 instances of Jubensha games in Chinese online.  Each game consists of a host manual describing how to control the game process and a God's-eye view of case replays, along with individual scripts for each character in the game. As demonstrated in Table~\ref{tab:dataset_statics1}, the number of players can vary from 1 to 20, and the number of tokens for the game can be as large as 518k, facilitating further research on socially intelligent AI and introducing extra-long text comprehension and reasoning challenges. Besides, as shown in Table~\ref{tab:horizontal_dataset_statistics}, some of these scripts also contain multimodal clues, including audio and video. To create a unified experimental environment, this work concentrates exclusively on text-modality Jubensha games.

\begin{figure*}[t]
    \centering
    \includegraphics[width=1\textwidth]{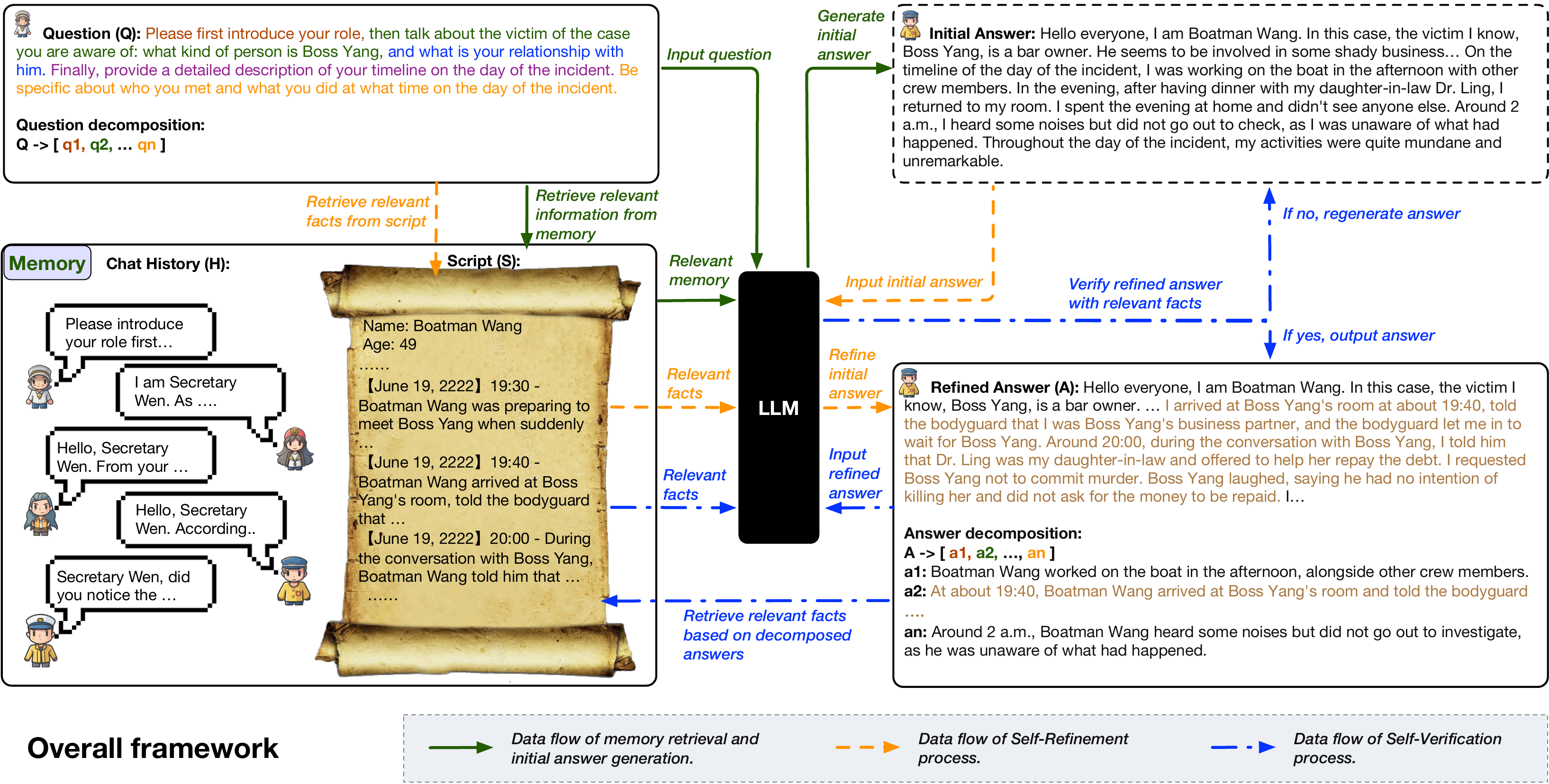}
    \caption{Illustration of our proposed ThinkThrice framework for enhancing agent's performance in multi-agent detective games (i.e., Jubensha). The three different colors of the arrows indicate the data flows of three stages: 1) Initial answer generation with Memory Retrieval; 2) Enhance answer with Self-Refinement; 3) Verify answer with Self-Verification. The \textcolor{brown}{brown} texts in the refined answer are new information added to the initial answer.}
    \vspace{-4mm}
    \label{fig:framework}
\end{figure*}

\section{The \textit{ThinkThrice} Framework for Jubensha Games}

We have designed an interaction framework for LLM-based agents specifically crafted for Jubensha games. In this framework, each LLM-based agent is responsible for playing the role of a player in the Jubensha game narrative. Typically, each Jubensha game involves 4-5 players portrayed by LLM-based agents. Besides, we have a non-player character serving as the host, who is responsible for guiding the game. Unlike other players driven by large language models, the host's questions and commands are pre-set to ensure the game follows a designated process. As for the LLM-based agent players, they are embedded with character information specific to their roles. Based on this information, they interact with other characters including other players and the host. These interactions are synchronous, meaning only one agent acts at a specific moment. All actions of the LLM-based agent players, such as whom to question or how to respond to questions from other players or the host, are generated by an LLM based on their character scripts and historical chat records, as shown in Figure~\ref{fig:framework}. Moreover, we have also set a finite number of rounds for each game to mimic the time-limited rules of real Jubensha games. Although currently focusing on Jubensha, our framework design is flexible, potentially requiring minimal adjustments for application in other contexts.

We name our framework for multi-agent mystery games as \textit{ThinkThrice}, or \begin{CJK}{UTF8}{gkai}三思\end{CJK} in Chinese. The name comes from a Chinese proverb: ``thinking thrice before acting (\begin{CJK}{UTF8}{gkai}三思而后行\end{CJK})''. Figure~\ref{fig:framework} illustrates the overall design of our ThinkThrice framework, which outlines the process by which a player generates a response to a question through three main stages: Memory Retrieval, Self-Refinement, and Self-Verification.

% generate an initial answer with memory retrieval; ii) Self-Refinement: improve the initial answer by utilizing the retrieved question-related facts; iii) Self-Verification: further verify the enhanced answer by comparing it with answer-related facts. If it is good, we output the response (i.e., answer). Otherwise, we repeat the generation process.

% Using LLM-based agents for Jubensha games can be very expensive, so we have set a finite number of rounds for each game, considering budget constraints. This approach also mimics the time-limited rules of real Jubensha games, although we choose to limit the number of rounds rather than the game duration. We will discuss this in more detail in Section  6.1.

\paragraph{Memory Retrieval} 
Due to the limited context window of LLMs, a special memory retrieval module is often needed to store and appropriately retrieve all historical records observed by LLM-based agents~\cite{park2023generative,lai2023werewolf}. Our framework adopts this widely used method to help agents remember dialogues and events in the Jubensha games and retrieve specific memory fragments when needed. Specifically, we record all observations of the agent into a vector database exclusive to the agent using the OpenAI API~\cite{openaiEmbeddings}. When the agent observes a new event requiring action, we use the Faiss library\footnote{\url{https://github.com/facebookresearch/faiss}} to quickly search the agent's exclusive vector database for multiple historical records with the top 5 highest similarities and include them in the current action prompt. We have observed that this simple module is also very helpful in improving the performance of LLM-based agents in our Jubensha games.

% \subsection{Self-Refinement and Self-Verification}

% The decoding process of large language models involves continuous sampling using a probabilistic model, so the quality and authenticity of each output can be somewhat random. This characteristic is not desired in the setting of Jubensha games. Given the limited number of interactions, LLM-based agents must communicate efficiently with other agents in a limited number of attempts to gather enough information for case solving before the voting phase. Therefore, keeping the quality of large language models' outputs high and consistent is crucial for the performance of LLM-based agents in Jubensha games. For this, we designed Self-Refinement and Self-Verification modules to help LLM-based agents.

\paragraph{Self-Refinement}
The goal of the Self-Refinement module is to ensure that LLM-based agents provide as much information and details as possible when responding to questions from others. As shown in Figure~\ref{fig:framework}, when an agent receives a question from others, it first attempts to generate an initial answer internally (see the green lines). During the Self-Refinement stage, the agent decomposes the asked question \( Q \) into several sub-questions: \( q_1, q_2, \ldots, q_n \). It then uses these sub-questions to retrieve relevant facts from the character scripts \( S \) that can be used to answer these sub-questions. For each relevant fact, the agent checks whether they are included in the initial answer. If not, they are added to form part of the refined answer (see the orange lines in Figure~\ref{fig:framework}). We can see that the refined answer, after the Self-Refinement module, contains more details and information than the initial answer, thereby enhancing the agent's communication efficiency within a limited number of rounds.

\paragraph{Self-Verification}
LLM hallucinations are a very common and thorny problem~\cite{dhuliawala2023chainofverification}. To ensure that the content of LLM-based agents' answers is authentic and not hallucinated, we designed a Self-Verification module. This module breaks down the agent's answer \( A \) into multiple facts \( a_1, a_2, \ldots, a_n \) and then compares these facts one by one with the agent's character script \( S \) for authenticity. Only when the agent's answer meets our preset authenticity threshold (measured by the absolute number and accuracy of real facts) will the agent output the answer (see the blue lines in Figure~\ref{fig:framework}). If the answer does not meet the preset threshold, the agent will need to regenerate the answer until the authenticity of the agent's answer meets the requirements or exceeds the maximum number of attempts. Since the agent's best output may not necessarily occur in the last attempt, we score each answer based on the degree of authenticity and the number of words, keeping a copy of the highest-scoring answer each time. By doing so, we can ensure that even if all of the agent's attempts fail to meet the authenticity threshold, it will still output the highest quality answer from all its previous attempts.

\section{Evaluating LLM-based Agents in Jubensha Games}

Previous work has primarily employed metrics such as human-likeness and win rate to assess the performance of LLM-based agents in games~\cite{wang2023survey,lai2023werewolf}. These metrics either require substantial human involvement or likely leading to less reliable experimental conclusions due to the challenges in controlling variables~\cite{lai2023werewolf}.\footnote{This is particularly the case when both sides in the game are played by LLMs} Considering the unique characteristics of Jubensha games, we have designed two tasks to quantitatively and qualitatively evaluate the performance of LLM-based agents in Jubensha games: Factual Question Answering and Inferential Question Answering.

\paragraph{Factual Question Answering}

The process of gathering information in Jubensha games is crucial for LLM-based agents to understand the implied relationships and conflicts in the story, reconstruct the entire case, and deduce the truth. To evaluate how much information LLM-based agents gathered in Jubensha games, we designed a factual question-answering task and use GPTs to generate questions. We have generated a total of 720 factual questions for 4 selected Jubensha games containing 4 or 5 players, of which two examples are presented in Table~\ref{table:factual_ques}. Due to limited space, we have included the prompts of our generation method in the appendix~\ref{sec:prompts}.

% First, we use GPT-4 to read each character's script and then instruct it to generate question and answer pairs about that character. For each character, we create 40 factual questions, which are then reviewed and improved by human annotators. Next, we prompt the LLM-based agents to answer factual questions related to both their own and other characters. At the start of the game, each LLM-based agent can only access their own character's script, limiting their ability to answer questions about other characters. However, as the game progresses and players begin to communicate and interact, the LLM-based agents gain the information to answer more questions about other characters. We track the change in the accuracy rate of the LLM-based agents' responses to these questions to assess the amount of information they have gathered during the game. 

% In Table 2, we present two examples of factual questions, and the results of the LLM-based agents' performance on these factual questions will be detailed in Section 6.3.

\begin{table}[t]
\centering
\begin{tabularx}{0.45\textwidth}{|X|}
\hline
\multicolumn{1}{|c|}{\textbf{Example 1}} \\ \hline
\textbf{Question}: Who does Mate Zhang's brother work for? \\ 
\textbf{Answer}: Boss Yang \\ 
\textbf{Source}: He works for Boss Yang, owner of a famous bar in City A. (from Mate Zhang's script) \vspace{0pt}\\ \hline
\multicolumn{1}{|c|}{\textbf{Example 2}} \\ \hline
\textbf{Question}:  When did Dr. Ling meet Mate Zhang? \\ 
\textbf{Answer}: 18:20 \\ 
\textbf{Source}: 18:20, you went to meet Boss Yang, and met Mate Zhang just as you were about to arrive.  (from Dr. Ling's script) \vspace{0pt}\\ \hline
\end{tabularx}
\vspace{-2mm}
\caption{Two examples of Factual Questions.}
\vspace{-1mm}
\label{table:factual_ques}
\end{table}
\paragraph{Inferential Question Answering}

Once LLM-based agents have gathered the necessary information, another important step is using it for inference. This step is particularly crucial in Jubensha games, where the truth of the case is often obscured by numerous clues. To evaluate the reasoning abilities of LLM-based agents, we manually designed a total of 56 inferential questions for 4 selected Jubensha games, of which an example is shown in Table~\ref{table:inferential_ques}. Note that the PREMISE 1 is from Secretary Wen's script and the PREMISE 2 is from Dr. Ling's script. To answer these questions successfully, LLM-based agents must integrate information collected from different characters and make inferences based on this information. 

% This inferential question-answering task tests the abductive reasoning ability of LLM-based agents, as the answer is not explicitly stated in any player's script. 

% In Table~\ref{table:inferential_ques}, we present an example of these inferential questions. The PREMISE 1 is from Secretary Wen's script and PREMISE 2 is from Dr. Ling's script.

\begin{table}[t]
\centering

\begin{tabularx}{0.45\textwidth}{|X|}
\hline
 \textbf{Question}: Who is most likely to have made the thumping sound that Dr. Ling heard on her way to kill Boss Yang? \textit{A.} Secretary Wen, \textit{B.} Boss Yang, \textit{C.} Boatman Wang, \textit{D.} Mate Zhang, \textit{E.} Others. \\
\textbf{Answer:} \textit{A.} Secretary Wen\\
\textbf{Rationale (GT)}: \{\textit{Secretary Wen left Boss Yang's room through the ventilation duct at 23:00}\}\textsubscript{\textbf{PREMISE 1}}, and \{\textit{Dr. Ling prepared to kill Boss Yang through the ventilation duct also at 23:00}\}\textsubscript{\textbf{PREMISE 2}}. Therefore, it can be inferred that \{\textit{the 'thumping' sound Dr. Ling heard on her way to kill Boss Yang was most likely made by Secretary Wen}\}\textsubscript{\textbf{CONCLUSION}}.\\
\hline
\end{tabularx}
\vspace{-2mm}
\caption{Example of Inferential Questions.}
\vspace{-4mm}
\label{table:inferential_ques}
\end{table}

% \section{Experiment}
% \normalsize
% \begin{table*}[t]
%   \centering
%   \resizebox{\textwidth}{!}{
%   \begin{tabular}{lcccc}
%     \toprule
%     & \multicolumn{1}{c}{ \textbf{Own Question}} & \multicolumn{3}{c}{ \textbf{Other's Questions}} \\
%     \cmidrule(lr){2-5} 
%     & Average & Civilian's Question & Murderer's Question & Average \\
%     \midrule
%     No MR & 0.764 & 0.278 & 0.306 & 0.293 \\
%     MR & 0.734 & 0.358 & 0.391 & 0.394 \\
%     MR+SR & 0.735 & 0.464 & 0.447 & 0.473 \\
%     MR+SR+SV(N=1) & 0.765 & 0.474 & 0.572 & 0.498 \\
%     MR+SR+SV(N=3) & 0.760 & 0.508 & 0.52 & 0.509 \\
%     \bottomrule
%   \end{tabular}
%   }
% \label{subtable:label2}
% \caption{Performance of agents on different types of factual questions. 'Own Question' refers to questions generated from the agent's own script, while 'Other Question' refers to questions from scripts of other agents, inaccessible to the answering agent. 'Other Question' includes 'Civilian's Question' (from other civilians' scripts) and 'Murderer's Question' (from the murderer agent's script). The effectiveness of different module combinations in the agents - Memory Retriever (MR), Self-Refinement (SR), and Self-Verification (SV) - is assessed based on agents' ability to provide factual responses about themselves and others, as shown in this table.}
% \label{table:overall_performance}
% \end{table*}

\section{Experiment}
% \large

In this section, we will briefly discuss the various LLMs used in this work. Following this, we will present our experimental results, encompassing the evaluation of the information gathering and reasoning performance of LLM-based agents.

\subsection{Utilization of LLMs in Jubensha Game Stages}

Throughout the gameplay and other stages of the Jubensha game experiment, including the generation of factual questions, agent Q\&A sessions, and the evaluation of agent responses, we predominantly employed OpenAI's GPT-3.5 and GPT-4 models. Unless specifically indicated, the GPT-3.5 and the GPT-4 model mentioned in this paper refer to gpt-3.5-turbo-16k-0613 and gpt-4-1106-preview respectively. For text embedding we used text-embedding-ada-002 for gameplay and text-embedding-3-large for evaluation. More specific usages of GPT models are listed in the appendix~\ref{sec:utilization_llms}.

\begin{table}[t]
  \centering
  \resizebox{0.5\textwidth}{!}{
  \begin{tabular}{lcccc}
    \toprule
    & \multicolumn{1}{c}{ \textbf{Own Q}} & \multicolumn{3}{c}{ \textbf{Other's Q}} \\
    \cmidrule(lr){2-5} 
    & Avg & CQ & MQ & Avg \\
    \midrule
    No MR & 0.770 & 0.300 & 0.321 & 0.305 \\
    MR & 0.759 & 0.409 & 0.380 & 0.402 \\
    MR+SR & 0.757 & 0.467 & 0.487 & 0.471 \\
    MR+SR+SV(N=1) & 0.768 & 0.485 & \textbf{0.518} & 0.492 \\
    MR+SR+SV(N=3) & \textbf{0.772} &\textbf{0.495} & 0.514 & \textbf{0.498} \\
    \bottomrule
  \end{tabular}
  }
\label{subtable:label2}
\caption{Performance of agents on different kinds of factual questions. ``Own Q'' refers to questions generated from the agent's own script, while ``Other's Q'' refers to questions from scripts of other agents, inaccessible to the answering agent. ``Other's Q'' includes ``CQ'' (questions from other civilians' scripts) and ``MQ'' (questions from the murderer agent's script). The effectiveness of different module combinations in the agents is assessed based on agents' ability to provide factual responses about themselves and others. The number follows ``N='' denotes the maximum number attempts an agent can try in the Self-Verification stage.}
\vspace{-4mm}
\label{table:overall_performance}
\end{table}

\begin{table}[t]
  \centering
  \resizebox{0.45\textwidth}{!}{
  \begin{tabular}{lccc}
    \toprule
    %\cmidrule(lr){2-4} 
    & OpenAI & TF-IDF & Trigrams \\
    \midrule
    No MR & 0.062 &  0.000 &  0.000 \\
    MR & 0.798 &  0.578 & 0.027  \\
    MR+SR & 0.820 & 0.645 & 0.064  \\
    MR+SR+SV(N=1) & 0.825  &0.668  &0.075 \\
    MR+SR+SV(N=3) & 0.\textbf{831}  &\textbf{0.670}   &\textbf{0.077}  \\
    \bottomrule
  \end{tabular}
  }
% \label{subtable:label2}
\caption{Similarity scores between all players' scripts and chat histories of agents. }
\vspace{-4mm}
\label{table:doc_similarity}
\end{table}

\subsection{Evaluation on Agents' Responses to Factual Questions}
The Table \ref{table:overall_performance} presents the evaluation results on agents' responses to factual questions. We can see that agents with different module combinations performed quite well when answering questions from their own script. This is understandable because agents have full access to their own scripts throughout the game, thereby possessing all the information needed to correctly answer these questions. Since the Memory Retriever (MR) module mainly records the communication history among agents after the game starts, we can consider agents without the Memory Retriever as being in a memoryless state. The experimental results in the ``No MR'' row show that agents' accuracy in answering questions about themselves is significantly higher than in answering questions about others. This gap highlights the significant difference between the information agents possess about others and about themselves. After introducing the MR module, we observe a significant improvement in agents' accuracy in answering questions about others, reflecting the increase in information gained from interactions among agents, which helps them better understand the roles and stories of other characters in the game.

The Self-Refinement (SR) and Self-Verification (SV) modules mainly ensure the authenticity and comprehensiveness of information during agent communication. We can observe that the accuracy of agents in answering questions about others achieves the best results among all under the combined effect of the MR module, the SR module, and the SV module (with a maximum of 3 attempts). This demonstrates that our designed modules effectively enhance the efficiency of communication among agents in the Jubensha game, allowing them to acquire more case information under the same game round conditions.

\subsection{Similarities between Agent Chat Histories and All Players' Scripts}

To measure the overall information about all characters acquired by LLM-based agents in the game from the textual perspective, we employed three methods to assess the similarities between the chat histories among agents and the scripts of all players. Firstly, we concatenate all players' scripts into a single document, then we treat the chat histories among agents as another document. Then we used OpenAI API~\cite{openaiEmbeddings} to encode these two documents respectively. For documents beyond OpenAI API's max input length, we split them into non-overlapping chunks, average their text embeddings to represent the document.  Given embeddings of two documents, their cosine similarity can be calculated. Additionally, we utilized TF-IDF and trigrams to represent the two documents, and then calculated their cosine and Jaccard similarities, respectively. The results from Table~\ref{table:doc_similarity} show that agents embedded with MR, SR, and SV modules have chat histories in the game that are closer to the scripts of all players, which demonstrates that they have acquired more information about all players in a Jubensha game.

% Therefore, we can see that the addition of these modules further improves the accuracy of agents in answering questions about others. This indicates that these two modules significantly enhance the efficiency of communication among agents compared to just having the Memory Retriever module. We can observe that the accuracy of agents in answering questions about others achieves the best results among all under the combined effect of the Memory Retriever module, the Self-Refinement module, and the Self-Verification module (with a maximum of 3 attempts). This demonstrates that our designed modules effectively enhance the efficiency of communication among agents in the Jubensha game, allowing them to acquire more information about the game script under the same game round conditions.

% \subsection{GPT-3.5 and GPT-4}

\begin{figure}[t]
    \centering
    \includegraphics[width=0.4\textwidth, height=0.3\textheight, keepaspectratio]{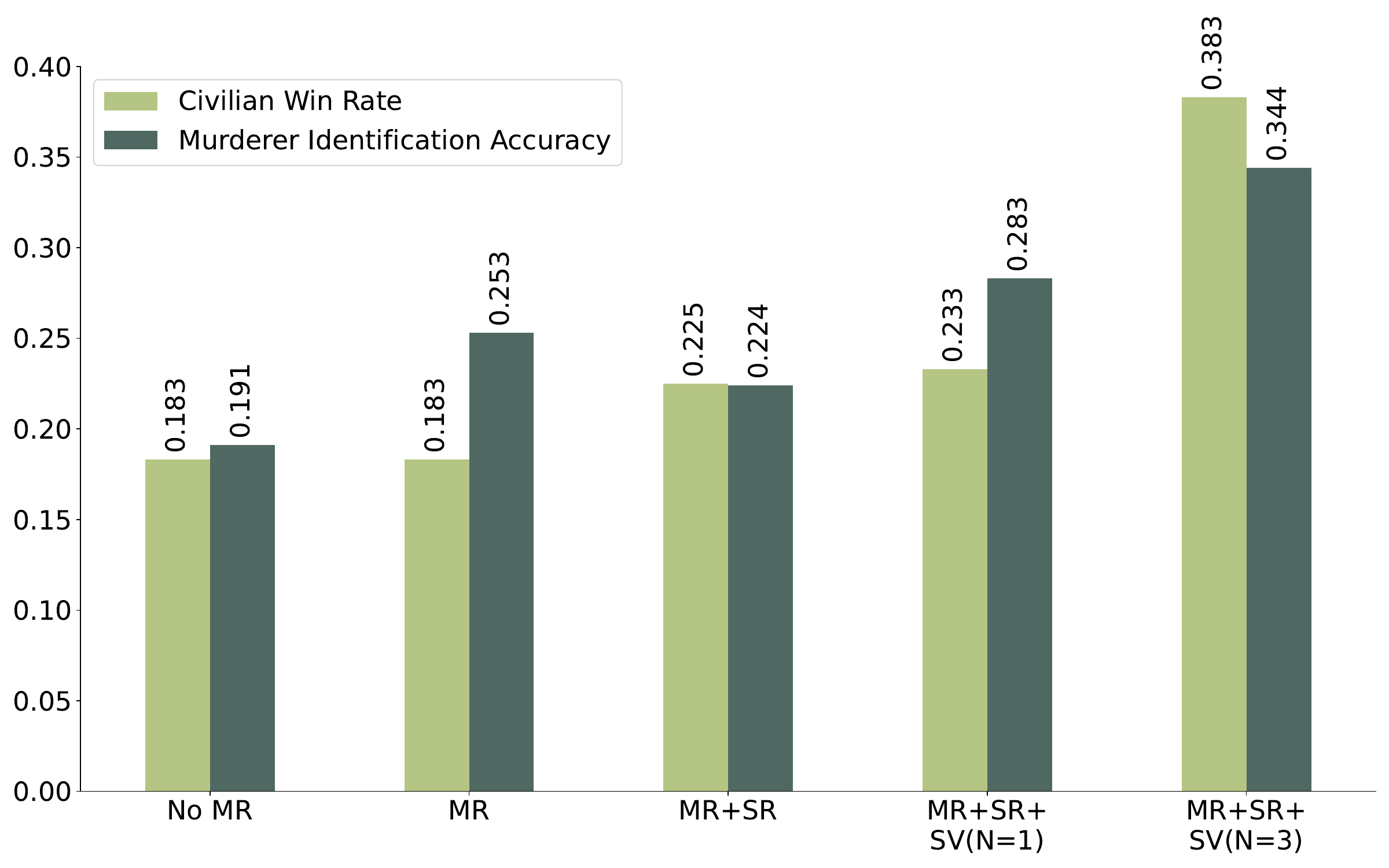}
    \caption{Average win rate of civilian players and the average murderer identification accuracy across different architectures in Jubensha games.}
    \vspace{-4mm}
    \label{fig:winning_rate}
\end{figure}
\vspace{-3mm}
\subsection{Civilian Player Win Rate and Murderer Identification Accuracy}

Figure~\ref{fig:winning_rate} shows the average win rates of civilian players and the murderer identification accuracy by agents with various module combinations in Jubensha games. Civilian win rate in Figure~\ref{fig:winning_rate} is the ratio of the number of games won by civilian players to the total number of games played. Murderer identification accuracy is the proportion of votes received by the murderer to the total number of votes cast by all players. We observed that MR+SR+SV(N=3) which performs best in factual question answering tasks, also achieves the highest average civilian players win rate and murderer identification accuracy. This might be due to agents acquiring sufficient information to reconstruct the true narrative of the case, thereby making it easier to accurately identify the murderer.

\subsection{Evaluation of Agent's Responses to Inferential Questions}

% To further evaluate the reasoning capabilities of LLM-based Agents on the basis of collected information, we utilized a set of inferential questions (detailed in Section 6.2). The experimental results are displayed in Figure 5. It is important to note that the Overall Accuracy shown in Figure 5 represents the rate at which agents correctly answer questions without considering the basis of their responses. In contrast, the Informed Accuracy refers to the accuracy when taking into account the basis agents provide for their answers. An answer is only counted towards the Informed Accuracy if the agent's response is correct and they provide the correct reasoning basis. This metric reflects that agents not only know the correct answer but also understand why it is correct.

\begin{figure}[t]
    \centering
    \includegraphics[width=0.4\textwidth, height=0.3\textheight, keepaspectratio]{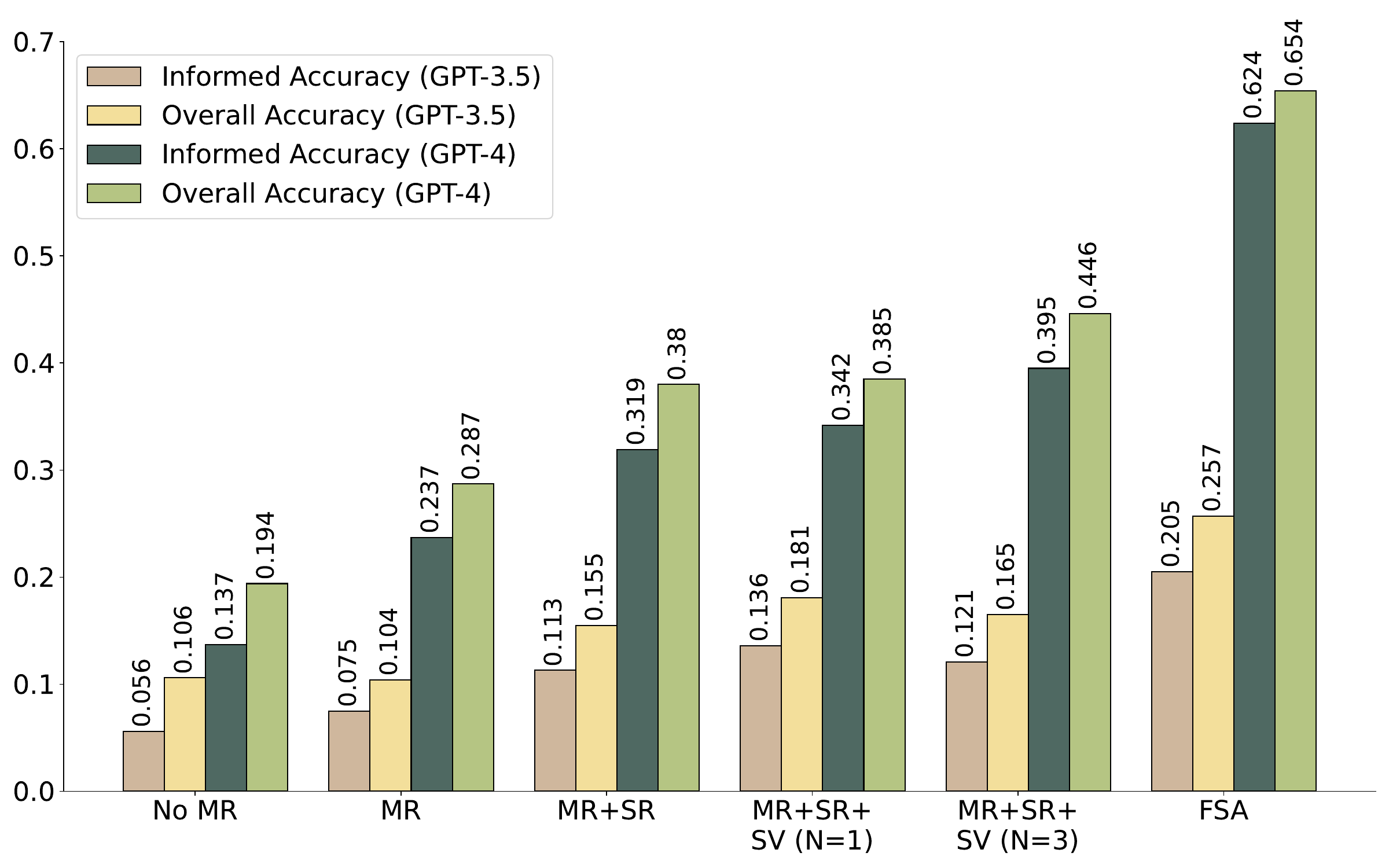}
    \caption{GPT-3.5 and GPT-4's performance with different methods, where overall accuracy measure the raw correct percentage and informed accuracy take LLM's reasoning ability into consideration. FSA stands for 'Full Script Access', indicating that agents have access to the complete scripts of all players.}
    \vspace{-4mm}
    \label{fig:GPT3.5 vs GPT-4}
\end{figure}

To further evaluate the reasoning capabilities of LLM-based agents based on the rationale derived from collected information, we utilized a set of inferential questions. The experimental results are displayed in Figure~\ref{fig:GPT3.5 vs GPT-4}, where the Overall Accuracy represents the rate at which agents correctly answer questions, without considering the rationale behind their responses. In contrast, Informed Accuracy refers to the accuracy achieved when taking into account the rationale provided by agents for their answers. An answer is counted towards Informed Accuracy only if the agent's response is correct and they provide the correct reasoning rationale. This metric demonstrates that agents not only know the correct answer but also understand why it is correct.

From the results in Figure~\ref{fig:GPT3.5 vs GPT-4}, we can draw two important observations: 1) The more information agents acquire during the game, the more capable they become at solving complex problems through reasoning. For instance, when GPT-4 is employed, the agent with full access to all players' scripts achieves the highest overall accuracy and informed accuracy. Following this, agents equipped with MR+SR+SV(N=3) modules rank second. 2) Given the same amount of information, the LLMs' inherent ability to utilize this information and reason effectively determines the agents' final performance in reasoning tasks. Indeed, simply upgrading from GPT-3.5 to GPT-4 can lead to agents achieving double or even nearly triple the overall accuracy and informed accuracy.

\begin{table*}[ht]
\centering
\resizebox{0.9\textwidth}{!}{
\begin{tabularx}{\textwidth}{|p{0.95cm}|X|}
% \hline
% \textbf{Question
% } & Who is most likely to have made the thumping sound that Dr. Ling heard on her way to kill Boss Yang? A. Secretary Wen, B. Boss Yang, C. Boatman Wang, D. Mate Zhang, E. Others. \\
% \textbf{Answer
% } & A. Secretary Wen\\
% \textbf{Rationale (Ground Truth)
% } & \{\textit{Secretary Wen left Boss Yang's room through the ventilation duct at 23:00}\}\textsubscript{\textbf{PREMISE 1}}, and \{\textit{Dr. Ling prepared to kill Boss Yang through the ventilation duct also at 23:00}\}\textsubscript{\textbf{PREMISE 2}}. Therefore, it can be inferred that \{\textit{the 'thumping' sound Dr. Ling heard on her way to kill Boss Yang was most likely made by Secretary Wen}\}\textsubscript{\textbf{CONCLUSION}}.\\
% \hline
% \textbf{Prompt} & When you previously answered the question, "\{\textit{Who is most likely to have made the thumping sound that Dr. Ling heard on her way to kill Boss Yang? A. Secretary Wen, B. Boss Yang, C. Boatman Wang, D. Mate Zhang, E. Others}\}\textsubscript{\textbf{QUESTION}}", you concluded with the answer "\{\textit{A. Secretary Wen}\}\textsubscript{\textbf{ANSWER}}" based on reasoning. Please explain your reasoning process in a few sentences. \\
\hline
\multicolumn{1}{|l|}{} & \multicolumn{1}{c|}{\textbf{Predicted Rationale}} \\
\hline
    \textbf{No\newline MR} & \{\textit{Dr. Ling couldn't find the medical cabinet key when she returned to her room at 18:50, suggesting someone had already taken it}\}\textsubscript{\textbf{IRRELEVANT PREMISE 1}}; \{\textit{Secretary Wen asked Dr. Ling for intelligence at 18:10}\}\textsubscript{\textbf{IRRELEVANT PREMISE 2}}, \{\textit{giving her the opportunity to know Boss Yang's location and possibly plan an action}\}\textsubscript{\textbf{TENUOUS INTERIM CONCLUSION 3}}; plus, when Dr. Ling heard the thumping sound, she was on her way to Boss Yang's room, and \{\textit{there's no clear evidence that other suspects were near Boss Yang's room at that time}\}\textsubscript{\textbf{FLAWED PREMISE 4}}; moreover, \{\textit{the ventilation duct's baffle had fallen, indicating someone had entered Boss Yang's room before her}\}\textsubscript{\textbf{RELEVANT INTERIM CONCLUSION 5}}; \{\textit{considering Secretary Wen's role under Boss Yang, she had the chance to approach him}\}\textsubscript{\textbf{TANGENTIALLY RELEVANT PREMISE 6}}, \{\textit{hence the thumping sound was most likely made by Secretary Wen}\}\textsubscript{\textbf{TENUOUS CONCLUSION}}. (\textbf{GPT Eval Score: 3}) \\
\hline
\textbf{MR} & \{\textit{According to the detailed timeline of the day of the incident for Dr. Ling, she heard someone crawling in the ventilation duct at 23:00 when she was preparing to go to Boss Yang's, and she reached his room at 23:20, only to find him already dead}\}\textsubscript{\textbf{RELEVANT PREMISE 1}}. \{\textit{Secretary Wen was the only one aware of Dr. Ling's mission and plan}\}\textsubscript{\textbf{FLAWED PREMISE 2}}, \{\textit{and clues suggest Secretary Wen might have tampered with the intelligence letter, leaving information about the transaction between Dr. Ling and Mate Zhang}\}\textsubscript{\textbf{IRRELEVANT PREMISE 3}}, \{\textit{possibly to make Boss Yang suspicious of Dr. Ling.}\}\textsubscript{\textbf{IRRELEVANT INTERIM CONCLUSION 4}} \{\textit{This indicates Secretary Wen might have had a motive to be involved in Boss Yang's death.}\}\textsubscript{\textbf{TANGENTIALLY RELEVANT PREMISE 5}} Combining all this information, \{\textit{Secretary Wen is the most likely person to have been in the ventilation duct and possibly involved in Boss Yang's death}\}\textsubscript{\textbf{TENUOUS CONCLUSION}}. (\textbf{GPT Eval Score: 4}) \\
\hline
\textbf{MR\newline \hspace*{0.25cm}+\newline \hspace*{0.1cm}SR\newline \hspace*{0.25cm}+\newline \hspace*{0.1cm}SV \newline (N=3)}& The sound Dr. Ling heard in the ventilation duct came from Secretary Wen. \{\textit{According to Dr. Ling's timeline, she entered the ventilation duct at 23:00 and reached Boss Yang's cabin at 23:20, hearing someone crawling inside }\}\textsubscript{\textbf{RELEVANT PREMISE 1}}.\{\textit{Secretary Wen mentioned waiting in the ventilation duct in her timeline}\}\textsubscript{\textbf{LIMITED PREMISE 2}}, \{\textit{so it is likely she made the noise}\}\textsubscript{\textbf{PLAUSIBLE CONCLUSION}}. \{\textit{Mate Zhang and Boatman Wang's actions and timelines do not mention being active in the ventilation duct}\}\textsubscript{\textbf{PARTIALLY CORRECT PREMISE 3}}. \{\textit{Boss Yang was already dead at that time, so the sound couldn't have come from him.}\}\textsubscript{\textbf{RELEVANT INTERIM CONCLUSION 4}} (\textbf{GPT Eval Score: 5}) \\
\hline
\textbf{Full\newline Script\newline Access} & \{\textit{When Dr. Ling was preparing to enter Boss Yang's cabin through the ventilation duct at 23:00, she heard a 'thumping' sound coming from the duct}\}\textsubscript{\textbf{RELEVANT PREMISE 1}}. \{\textit{According to Secretary Wen's timeline of the day of the incident, she reached the ventilation duct at 22:40 and re-entered it at 23:00 to return}\}\textsubscript{\textbf{RELEVANT PREMISE 2}}. \{\textit{This timing matches the moment Dr. Ling heard the thumping sound}\}\textsubscript{\textbf{RELEVANT AND SOLID INTERIM CONCLUSION 3}}, thus it can be inferred that \{\textit{the sound was most likely Secretary Wen crawling in the ventilation duct}\}\textsubscript{\textbf{SOLID CONCLUSION}}. (\textbf{GPT Eval Score: 5}) \\
\hline
\end{tabularx}
}
\caption{Qualitative analysis of agent reasoning performance with different levels of information access.}
\vspace{-4mm}
\label{table:qualitative_analysis}
\end{table*}
\paragraph{Qualitative Analysis.}

In Table~\ref{table:qualitative_analysis}, we provide a qualitative analysis: agents are presented with an inferential question and the correct answer which the agents do not know to be correct, and are asked to provide the rationale for their reasoning. Each rationale provided by the agents is accompanied by a GPT Eval Score, which measures the similarity between the agents' rationale and the ground truth rationale. The inferential question, answer, and the ground truth rationale can be found in Table~\ref{table:inferential_ques}. From Table~\ref{table:qualitative_analysis}, we observe that with Full Script Access (complete information), agents can easily identify relevant premises and deduce solid conclusions. In cases where agents have incomplete information, MR+SR+SV (N=3), due to lacking specific details about the exact time Secretary Wen was in the ventilation duct, the conclusion drawn is merely plausible. For MR and No MR, owing to the absence of many key details, the reasoning process involves many irrelevant or flawed premises, leading to only tenuous conclusions. This qualitative analysis demonstrates the significant impact that the collection of key information during the game has on the agents' final reasoning performance.

\section{Conclusion}

This work has explored the application of large language models in complex interactive environments, exemplified by the Chinese detective role-playing game "Jubensha". Our research has yielded four main contributions: the creation of a specialized dataset tailored for the Jubensha games, the design of a multi-agent interaction framework for the automatic conduct of Jubensha games, the development of a set of quantitative and qualitative assessment methods to measure the information gathering and reasoning abilities of LLM-based agents within the game, and the utilization of the latest in-context learning techniques to devise modules that enhance the performance of LLM-based agents. We have empirically demonstrated that our designed multi-agent interaction framework and the in-context learning modules significantly improve upon the baseline in terms of information gathering, murderer identification, and reasoning capabilities. We believe this research will advance the community's knowledge of LLM-based agents and offers a new perspective on evaluating the performance of LLMs in a complex, plot-driven, and adversarial reasoning game environment constrained by narrative contexts.

\begin{CJK}{UTF8}{gkai}
\section*{Ethical Considerations}

Our research delves into the communicative and reasoning abilities of large language models (LLMs) within the context of "Jubensha" (剧本杀), a type of Chinese detective role-playing game. It is important to clarify that any portrayals of violence in these game scenarios are purely fictional, and our work is solely for the purposes of academic analysis. It does not represent or endorse real-world violence in any way. The data employed in our study is gathered from online platforms hosting Jubensha content. When sharing this dataset, we will implement measures to ensure that its usage remains strictly for academic, non-commercial purposes and complies with fair use policies.~\footnote{\url{https://www.copyright.gov/fair-use/index.html}\\\url{https://www.gov.cn/guoqing/2021-10/29/content_5647633.htm}}\end{CJK}

\section*{Limitations}

We outline the main limitations of our work as follows:

\begin{itemize}
    \item \textbf{Language Specificity of the Dataset:} Our Jubensha dataset is in Chinese, which means our experimental results are specifically reflective of the communicative and reasoning capabilities of Large Language Model (LLM) based agents in Chinese contexts. Considering that the majority of LLM evaluation benchmarks are in English, and existing LLM-based agent frameworks are tailored for English applications, our Chinese-centric benchmark and framework could provide a valuable addition to the field.
    
    \item \textbf{Variability in Experimental Outcomes:} The inherent stochastic nature of LLM outputs may lead to significant variability in single-experiment results. For instance, in the murderer identification voting phase, LLM-driven players may choose differently among suspects if given another chance. Similarly, game processes can diverge significantly, even with identical starting conditions. To mitigate this, we conducted three runs of each Jubensha game script with LLM-based agents for each proposed architectures and averaged the outcomes. For highly variable tasks like murderer identification, agents performed 10 memory-less votes, from which we calculated identification accuracy and civilian win rates. However, due to time and budget constraints, further experiments to solidify these results were not feasible. We will make our code, dataset, and all intermediate results available post-acceptance, including chat histories, voting records, agents' responses to factual and inferential questions, and GPT models' evaluations on these responses, to help interested readers replicate our experiments and better understand our findings.
    
\item \textbf{Model Updates and Replication Costs:} Reproducing our findings might be challenging due to OpenAI's periodic model updates, which can lead to different results with different model versions. To address this, we specify the exact version of GPT used in each experiment in the appendix~\ref{sec:utilization_llms} and will publish our code post-acceptance. Moreover, the cost of replicating our experiments can be costly. We aim to alleviate this by providing an illustrative cost breakdown for each experimental step in the appendix~\ref{sec:cost_breakdown}, helping readers gauge potential expenses before attempting replication.
\end{itemize}

\bibliography{anthology,custom}
\bibliographystyle{acl_natbib}

\clearpage
\appendix
\section*{Appendix}
\section{Human Evaluation on the Quality of Agents' Responses}

To study the impact of different architectures on the quality of agents' responses, we selected responses from agents at the Self-Introduction Stage for comparison. More specifically, for agents with each of five architectures, we selected 20 responses, resulting in a total of 100 responses. These were compared in groups of five, with each group consisting of agents playing the same character in a scripted murder mystery game. During the evaluation, we also provided the character scripts of the agents' roles as a ground truth reference. A Chinese native speaker human annotator was asked to score the agents' responses based on their naturalness, authenticity, and informativeness on a scale of 1 to 5. The human annotator was unaware of the specific architecture used by each agent. The specific scoring guidelines are as follows:

\begin{itemize}
    \item \textbf{Naturalness:} Naturalness primarily examines the naturalness and fluency of the agent response. Since the Jubensha game is a role-playing game, human responses are primarily in the first person. The more closely a response resembles a human answer, the higher its naturalness score. Responses mixed with different languages or not in the first-person tone receive lower naturalness scores. Note that the naturalness score is unrelated to the character script of the agents' portrayed character.
    
    \item \textbf{Authenticity:} Authenticity mainly examines whether the agent response contains content that is true to the character script of the agents' portrayed character. The Human annotator gives scores based on the proportion of the response's truthfulness, with higher scores indicating greater authenticity and lower scores indicating less. Note that, according to the rules of the Jubensha games, the murderer is allowed to lie in the game to cover up its identity. For responses from agents playing the role of the murderer, the human annotator bases the authenticity score on the following principle: if the content in the agent response that differs from the character scripts does not involve other players in the game (meaning that other players cannot judge the truthfulness of the agent response), this is considered strategic lying by the agent playing the murderer and does not affect its authenticity score. Otherwise, points will be deducted.
    
    \item \textbf{Informativeness:} Informativeness primarily examines whether the agent response provides enough information from its own character script. Human annotators score based on how much of their character script the agent response covers, with higher scores indicating more information and lower scores indicating less. Since murderers in the Jubensha game are allowed to conceal information, the human annotator bases the informativeness score for responses from agents playing the murderer role on the following principle: if the information in the character script not mentioned in the agent response could lead to suspicion of the agent, the coverage or lack thereof of this information is not considered in the informativeness score.
\end{itemize}

\begin{figure*}[h]
    \centering
    \includegraphics[width=1.0\textwidth, height=0.8\textheight, keepaspectratio]{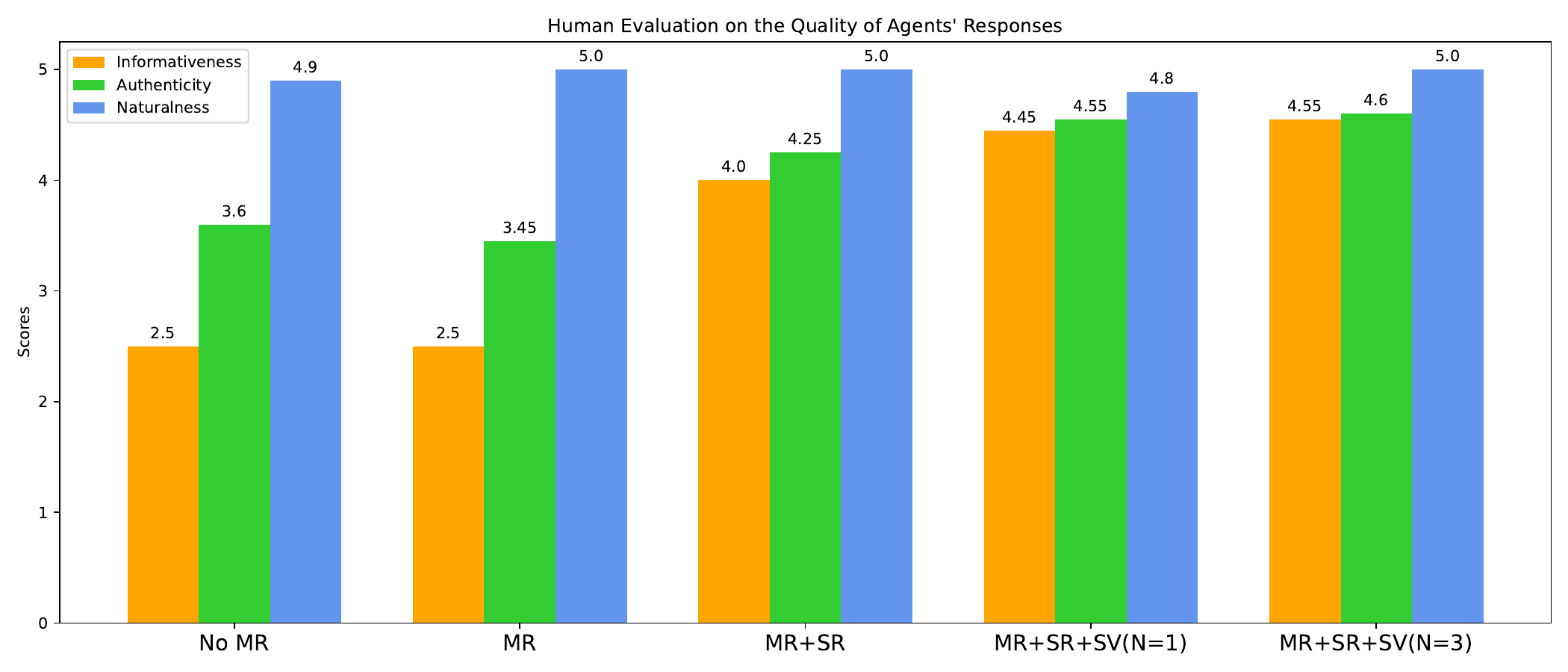}
    \caption{Human Evaluation on the Quality of Agents' Responses.}
    \vspace{-4mm}
    \label{fig:human_eval_fluency_rel_info}
\end{figure*}

From Figure~\ref{fig:human_eval_fluency_rel_info}, we can see that MR+SR+SV (where N=3) scored the highest in terms of naturalness, authenticity, and informativeness among the evaluation metrics. This demonstrates that our proposed self-refinement and self-verification models can help agents improve the authenticity and informativeness of their responses without sacrificing naturalness.

\vspace{-3mm}

\section{Utilization of LLMs in Jubensha Game Stages}
\label{sec:utilization_llms}
Throughout the gameplay and other stages of the Jubensha game experiment, including the generation of factual questions, agent Q\&A sessions, document embeddings, and the evaluation of agent responses, we predominantly employed OpenAI's GPT-3.5 and GPT-4 models. Unless specifically indicated, the GPT-3.5 model mentioned in this paper refers to gpt-3.5-turbo-16k-0613, and GPT-4 refers to gpt-4-1106-preview. Specific applications of each model are as follows:

\begin{itemize}
  \item \textbf{Gameplay:} GPT-3.5 was used due to its larger context window and budget considerations.\footnote{Before the public release of \texttt{gpt-4-1106-preview}, the publicly accessible GPT-4 model had a maximum context size of 8k, which is insufficient for the Jubensha game.}
  % GPT-4's limited context window prior to the release of \texttt{gpt-4-1106-preview} and cost factors influenced this decision.
  Future experiments may include GPT-4 as costs become more feasible. For the  text embeddings used for the memory retriever module, we used text-embedding-ada-002.

  \item \textbf{Factual Question Generation:} Both GPT-3.5 and GPT-4 (\texttt{gpt-4-0613}) were used, with no notable difference in output quality observed.

  \item \textbf{Agent Q\&A Sessions:} GPT-3.5 was used for factual question responses, while both GPT-3.5 and GPT-4 were tested for inferential question responses.

  \item \textbf{Response Evaluation:} GPT-3.5 assessed agents' responses to factual questions, whereas GPT-4 evaluated agents' inferential question responses. 
  
  % Note that agents' voting decisions are assessed with a heuristic text-matching method as we found it is. 
  % A manual review of 200 examples was conducted to validate the models' evaluations against human judgments, using Pearson and Spearman's coefficients.
  
    \item \textbf{Document Embedding:} We used text-embedding-3-large to obtain the document embeddings of agents' chat histories and all players' scripts.
\end{itemize}

\section{Detailed Jubensha Game Rules and Procedure}

The detailed Jubensha game rules we used for our experiments are as follows: 

{
\setlength{\parindent}{0pt}
\setlength{\parskip}{0.8em}
{
%\fontfamily{ComicNeue-OsF}\selectfont

% \setmainfont{courier-prime-sans.regular.ttf}
\begin{adjustwidth}{0.3cm}{0.3cm}   
\ttfamily
\noindent \textbf{Rule 1:} The total number of players participating in the game may be four or five, depending on the script. Only one of these players is the real murderer, known as the murderer player. Players who are not the murderer are collectively referred to as civilian players.

\textbf{Rule 2:} Civilian players need to cooperate to face a carefully planned murder case and find the real murderer among the suspects by collecting evidence and reasoning.

\textbf{Rule 3:} Throughout the game, only the murderer player can lie. To hide their identity, the murderer may choose to frame others and exonerate themselves.

\textbf{Rule 4:} Players who are not the murderer (civilian players) must answer questions honestly from other players and the host, and provide as much detailed information about the case as they know, to help restore the truth of the case.

\textbf{Rule 5:} The host of the game is responsible only for ensuring that the game proceeds according to a specific process. They are not a player in the game and do not participate in the game's storyline.

\textbf{Rule 6:} Each player receives their personal character script from the host at the beginning of the game, thereby learning about their character's information and identity.

\textbf{Rule 7:} The content of each player's personal character script is invisible to other players, so players must and can only gather information about other players by interacting with them after the game starts.

\textbf{Rule 8:} Since there is only one murderer in the game, only the murderer player knows the identities of the other players after receiving their character script (since everyone else is a civilian). Civilian players cannot determine the true identities of other players and can only infer through interactions during the game.

\textbf{Rule 9:} During the voting phase, each player has exactly one vote, which they can cast for the player they think is the murderer (including themselves, although this is discouraged). If the player with the most votes is the murderer, the civilian players win. Otherwise, the murderer player wins.

\end{adjustwidth}
}
}

The overall procedure of the game is as follows:
{
\setlength{\parindent}{0pt}
\setlength{\parskip}{0.8em}
{
\begin{adjustwidth}{0.3cm}{0.3cm}
\raggedright\ttfamily
\noindent \textbf{Stage 1: Distribution of Character Scripts} \\
The host distributes character scripts to each player individually. The script includes the player's name, identity (murderer or civilian), character story, and timeline for the day of the incident.

\textbf{Stage 2: Self-Introduction Session} \\
Players introduce themselves under the host's guidance, explaining their relationship with the victim and their timeline on the day of the incident.

\textbf{Stage 3: Initial Questioning} \\
After a player responds to the host's inquiries, other players are given the opportunity to ask questions and receive answers from that player.

\textbf{Stage 4: Two Rounds of Open Questioning} \\
Players enter two rounds of open questioning, with each taking turns to question another player and obtain answers.

\textbf{Stage 5: Distribution of Clue Cards} \\
Players receive clue cards containing additional information about the victim and the players, aiding them in deducing the case's storyline.

\textbf{Stage 6: Three Rounds of Open Questioning} \\
Players enter three rounds of open questioning, with each taking turns to question another player and obtain answers.

\textbf{Stage 7: Voting} \\
Under the host's guidance, players anonymously vote to determine who they believe is the murderer. Each player has one vote.

\textbf{Stage 8: Outcome Reveal} \\
The game concludes with the revelation of the voting results.

\end{adjustwidth}
}

% \clearpage
% \onecolumn

\section{Experimental Cost Breakdown}
\label{sec:cost_breakdown}
To assist readers in understanding the potential costs involved in our experiments prior to attempting a replication, we provide an illustrative cost breakdown for each experimental step in Table~\ref{table:cost_breakdown}. In this example, we utilized agents with MR+SR+SV(N=3) modules, with the Jubensha game script being "The Doomed Sunshine," consisting of about 16k tokens. We observed that the most expensive stage of the experiment was using GPT-4 to evaluate agents’ responses to inferential questions, while the cheapest stage involved using text-embedding-3-large to calculate document similarity between all players' scripts and chat histories of agents. Summing up all the costs, we can see that the total expense for a single complete experiment is approximately 11.9 USD.

\begin{table*}[thbp]
\centering

\label{tab:experiment_data}
\begin{tabular}{@{}llr@{}} % 左对齐, 左对齐, 右对齐
\toprule
Stage & Model & Cost (\$) \\
\midrule
Gameplay & GPT-3.5, text-embedding-ada-002 & 2.16 \\
\multirow{2}{*}{Eval inferential question} & GPT-3.5 & 1.37 \\
 & GPT-4 & 4.6 \\
Eval factual question & GPT-3.5 & 1.9 \\
Murderer identification voting & GPT-3.5 & 1.86 \\
Doc similarity & text-embedding-3-large & < 0.01 \\
\bottomrule
\end{tabular}
\caption{Experimental cost breakdown in different stages.}
\label{table:cost_breakdown}
\end{table*}

\begin{table*}[thbp]
\centering
\begin{tabular}{clp{10cm}}
\toprule
Score & Rating & Description \\
\midrule
1 & Very Poor & The agent's rationale is vastly different from the ground truth rationale, with reasoning steps completely illogical and impossible to justify the answer based on the reasoning provided. \\
2 & Poor & The agent's rationale is only somewhat similar to the ground truth rationale in certain aspects, with most reasoning steps illogical, having only a few that are correct or partially logical. \\
3 & Average & The agent's rationale is somewhat close to the ground truth rationale, with some reasoning steps correct, but there are significant errors or omissions. \\
4 & Good & The agent's rationale is very close to the ground truth rationale, with the vast majority of reasoning steps correct and logical, having only minor errors or deficiencies. \\
5 & Excellent & The agent's rationale is almost or entirely consistent with the ground truth rationale, with all reasoning steps correct and logical, demonstrating a high level of reasoning ability and a deep understanding of the problem. \\
\bottomrule
\end{tabular}
\caption{Evaluation Criteria for Agent Rationales Against Ground Truth.}
\label{table:eval_criteria}
\end{table*}

\section{Correlations between Automatic Evaluations and Human Evaluations}

In this work, we extensively utilized automatic evaluation methods to quantitatively assess the performance of LLM-based agents in Jubensha games. To ensure the reliability of our automatic evaluation methods, two Chinese native speaker human annotators were responsible for manually evaluating samples of the agents' responses. Below, we describe the human evaluation process across different stages:

\begin{itemize}
    \item \textbf{Evaluation on Factual Question Answering:} 200 factual questions and the agents' responses to these questions are randomly selected. The two human annotators then manually assessed the agents' responses as correct or incorrect based on the factual questions, their reference answers, and the agents' responses. During the evaluation process, the annotators were unaware of the agents' architectures, i.e., whether the agents were using MR architecture or MR+SR+SV(N=3), etc., nor were they aware of the GPT model's evaluation of agent responses. After human scoring, we compared the GPT model's scores as predictions with the human scores as ground truths, and then calculated Spearman correlation, Pearson correlation, accuracy, and F1 score, with results shown in Figure~\ref{fig:correlation}(a).

    \item \textbf{Evaluation on Inferential Question Answering (Overall Accuracy):} 100 inferential questions and the agents' responses to these questions are randomly selected. The two human annotators manually assessed the agents' answers as correct or incorrect based on the inferential questions, their reference answers, and the agents' responses. Similar to the first evaluation, the annotators were unaware of the agents' architectures and the GPT model's evaluation. After human scoring, we compared the GPT model's scores with human scores, calculating Spearman correlation, Pearson correlation, accuracy, and F1 score, with results shown in Figure~\ref{fig:correlation}(b).

    \item \textbf{Evaluation on Inferential Question Answering (Informed Accuracy):} 100 inferential questions and the agents' predicted rationales for the correct answers are randomly selected. Based on these inferential questions, their reference rationales, and the agents' predicted rationales, two human annotators manually scored the agents' predicted rationales on a scale of 1-5. The evaluation criteria are shown in Table~\ref{table:eval_criteria}. Similarly, annotators were unaware of the agents' architectures and the GPT model's evaluation. After human scoring, we converted both the GPT model's scores and human scores into good or bad, with scores of 4 or above considered good. We then compared the converted GPT scores with converted human scores to calculate Spearman correlation, Pearson correlation, accuracy, and F1 score, with results shown in Figure~\ref{fig:correlation}(c).

    \item \textbf{Evaluation on Murderer Identification (Overall Accuracy):} 100 instances of agents' voting records in murderer identification stage are randomly selected. Based on these records, the names of all player characters in the game, and the actual murderer's character name, two human annotators manually identified whether the voting records pointed to the actual murderer, marking it as correct if so, and incorrect otherwise. The annotators were unaware of the agents' architectures and the results of our text matching algorithm that compared the voting records with the game's actual murderer's character name. After human scoring, we compared the text matching algorithm's scores with human scores, calculating Spearman correlation, Pearson correlation, accuracy, and F1 score, with results shown in Figure~\ref{fig:correlation}(d).
\end{itemize}

As seen in Figures~\ref{fig:correlation}(a-d), all our automatic evaluations demonstrated strong correlation with human evaluations, further validating the effectiveness of our automatic evaluation methods and the reliability of our experimental results.

\section{AI-Assisted Writing and Coding}

In this work, we extensively utilized GPT-4 to assist in refining the language of the paper. This included tasks such as paraphrasing, spell-checking, or translating the original content provided by the authors. Additionally, we employed GPT-4's coding capabilities to help write simple utility functions. These functions were designed for operations such as reading our stored experimental result files, aggregating information, and generating statistical tables. When releasing the code, we will clearly indicate the parts that were aided by AI-assisted coding.

% 第一个图，包含子图a和b
\begin{figure*}[htbp]
    \centering
    \begin{subfigure}[b]{1.0\linewidth}
        \centering
        \includegraphics[width=\linewidth]{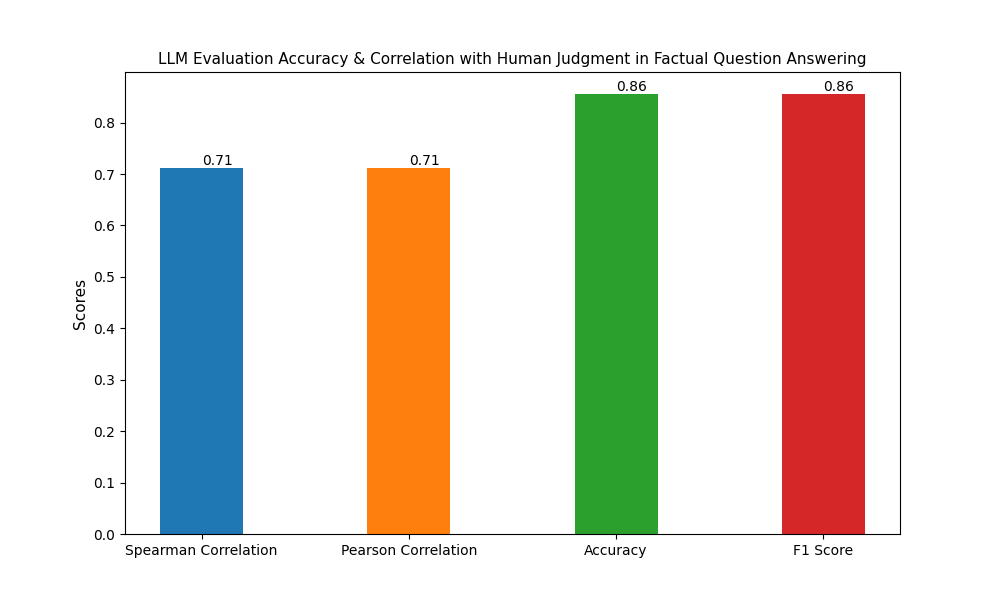}
        \caption{LLM Evaluation Accuracy \& Correlation with Human Judgment in Factual Question Answering.}
        \label{fig:sub1}
    \end{subfigure}
    
    \begin{subfigure}[b]{1.0\linewidth}
        \centering
        \includegraphics[width=\linewidth]{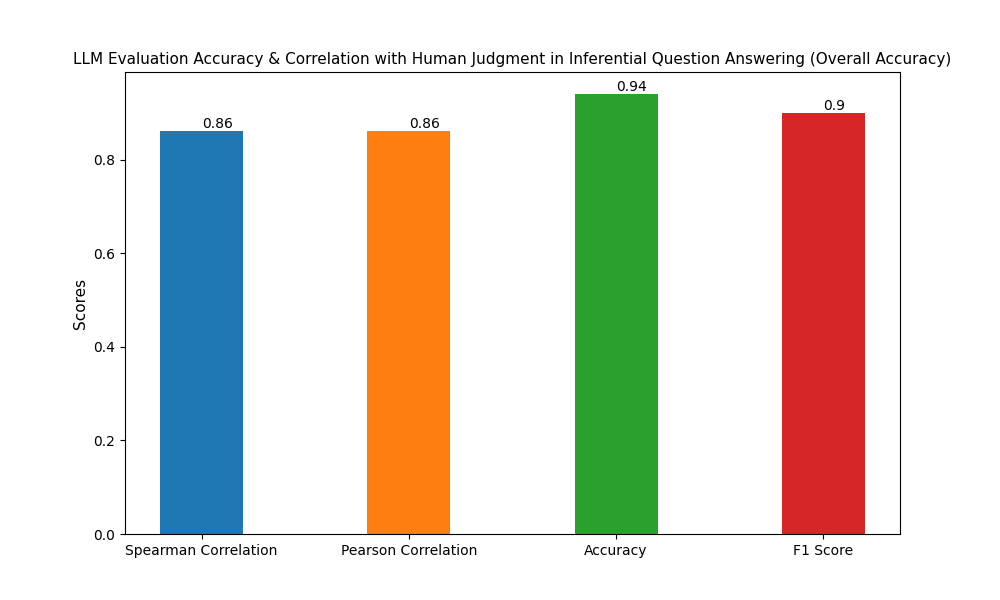}
        \caption{LLM Evaluation Accuracy \& Correlation with Human Judgment in Inferential Question Answering (Overall Accuracy).}
        \label{fig:sub2}
    \end{subfigure}
    
    \label{fig:combined1}
\end{figure*}

% 第二个图，包含子图c和d，使用\ContinuedFloat来保持编号连续
\begin{figure*}[htbp]
    \ContinuedFloat % 这一命令保证子图编号的连续性
    \centering
    \begin{subfigure}[b]{1.0\linewidth}
        \centering
        \includegraphics[width=\linewidth]{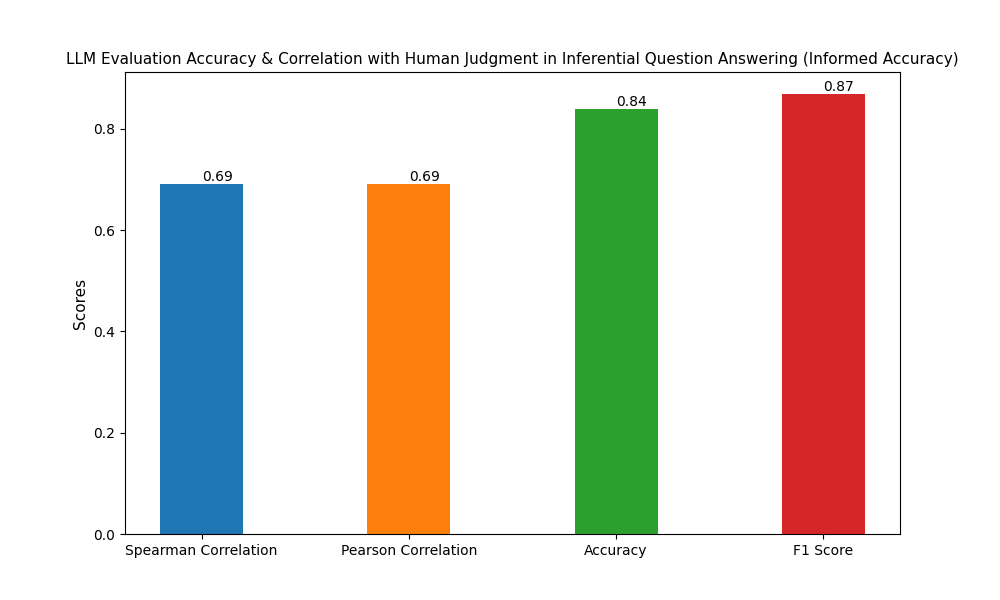}
        \caption{LLM Evaluation Accuracy \& Correlation with Human Judgment in Inferential Question Answering (Informed Accuracy).}
        \label{fig:sub3}
    \end{subfigure}
    
    \begin{subfigure}[b]{1.0\linewidth}
        \centering
        \includegraphics[width=\linewidth]{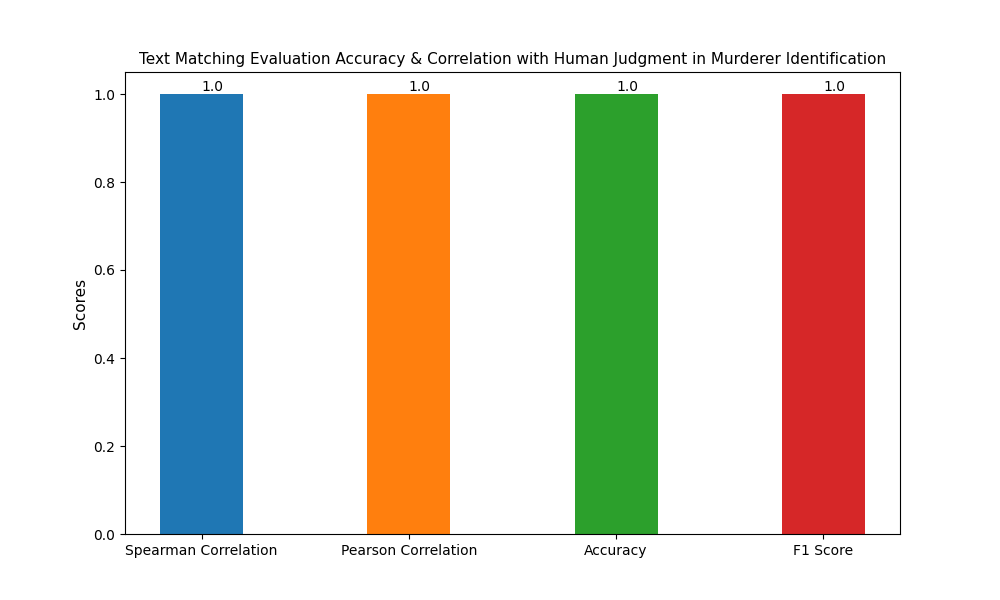}
        \caption{Text Matching Evaluation Accuracy \& Correlation with Human Judgment in Murderer Identification.}
        \label{fig:sub4}
    \end{subfigure}
    
    \caption{Correlation between human evaluations and automatic evaluations used in this work.}
    \label{fig:correlation}
\end{figure*}

\section{Authenticity Threshold and Answer Scoring in Self-Verification Stage}

During the self-verification stage, the authenticity threshold is determined by three key components:
\begin{itemize}\setlength\itemsep{0em}
    \item The accuracy of retrieved facts,
    \item The quantity of corrected facts,
    \item The length of the agent's response.
\end{itemize}
An agent's answer is only output if it surpasses these thresholds, provided the maximum number of attempts has not been exceeded. In our experiment, we employed two sets of parameters to define these thresholds:
\vspace{-2mm}

\begin{itemize}\setlength\itemsep{0em}
    \item For questions originating from the host:
    \begin{itemize}\setlength\itemsep{0em}
        \item Accuracy threshold: 0.7,
        \item Minimum number of corrected facts: 4,
        \item Minimum response length: 350 words.
    \end{itemize}
    \item For questions from other players:
    \begin{itemize}\setlength\itemsep{0em}
        \item Accuracy threshold: 0.6,
        \item Minimum number of corrected facts: 1,
        \item Minimum response length: 30 words.
    \end{itemize}
\end{itemize}

Regarding the scoring of answers, we utilize the following formula:
\vspace{-0.6mm}
\begin{align*}
    \text{Score} = & \ \text{Accuracy} \\
                   & + (\text{Number of Corrected Facts} \\
                   & + \ \text{Time-Matched Corrections} \\
                   & + \ \frac{\text{Length of Response}}{200})
\end{align*}

In this formula:
\vspace{-1mm}
\begin{itemize}\setlength\itemsep{0em}
    \item "Accuracy" reflects the accuracy of retrieved facts.
    \item "Number of Corrected Facts" indicates the total corrected facts.
    \item "Length of Response" is the word count of the agent's response.
    \item "Time-Matched Corrections" refers to the count of corrected facts that include specific time references, identified through regular expression matching.
\end{itemize}

\section{Prompts}
\label{sec:prompts}
In this section, we introduce the various LLM prompts used in this work. We have categorized these prompts by their different functions and compiled them into Tables~\ref{table:prompt_answering_question_cn} through \ref{table:prompt_inferential_qa_eval_cn}, totaling 9 tables. For each prompt, we also provide examples of LLM input and output to aid readers in understanding the specific operation of the prompt. For the convenience of our readers, we have translated the original Chinese prompt table content into English, with the English prompts tables located from Table~\ref{table:prompt_answer_questions_eng} to Table~\ref{table:prompt_inferential_qa_eng}.

% Define the color for the row here
\definecolor{lightgray}{gray}{0.9}
\begin{CJK}{UTF8}{gbsn}
\begin{table*}[ht!]
\centering
\begin{tabular}{|m{16cm}|}
\hline
\rowcolor{lightgray}  {\textbf{LLM Prompt for Answering Questions}} \\
% You are an expert programmer, and you need to write a program for the given natural language query. First, you should write a grammar that contains all the necessary BNF rules. Then, you should write programs that conform to your predicted rules. \\
\hline
% \textbf{query}: I need a meeting with Elli and her team on Wednesday afternoon. \\
% \textbf{BNF grammar rules:} \\
\begin{minipage}[t]{\linewidth}
\small\begin{Verbatim}[fontsize=\small]
"{agent_summary}"
+ "\n{game_rule}"
+ "\n{background_story}"
+ "\nDialogue currently happening in the game: {current_dialogue}"
+ "\nContent related to the ongoing dialogue in the game from previous conversations by {agent_name}: 
{relevant_memories}"
+ "\nRelationship with the person you're talking to: {relationship_with_interlocutor}"
+ "\nWhat will {agent_name} say? Please answer the question using the following format: #Answer#：what 
you want to say to answer the question."
+ "{agent_name} responds to {inquirer}'s question saying #Answer#:\n"

\end{Verbatim}
\end{minipage} \\
\hline

\rowcolor{lightgray}  \textbf{Example}  \\
 \textbf{   agent\_name:} \small Boatman Wang\\
\textbf{   inquirer:} \small Host\\
\textbf{   relationship\_with\_interlocutor:} \small The relationship with the speaker is: the speaker is the host, who is responsible for guiding the players to complete the game. \\
% \begin{minipage}[t]{\linewidth}
% \small\begin{verbatim}
% 对话人与你的关系是：对话人是主持人，他负责引导玩家完成游戏。
% \end{verbatim}
% \end{minipage} \\
\textbf{   current\_dialogue:} \small The host said to Boatman Wang: "Please start by introducing your character, and then talk about the victim of the case you know - Boss Yang. What kind of person was Boss Yang, and what was your relationship with him? Finally, provide a detailed timeline of what you were doing on the day of the incident. It needs to be specific about who you saw and what you did at what time on the day of the incident. \\
% \begin{minipage}[t]{\linewidth}
% \small\begin{verbatim}
% 主持人对王船工说："请你先介绍一下你的角色，然后说一下你所认识的案件的受害人：杨老板是一个怎么样的人，以及你和他的关系。最后再用一段话详细介绍一下你在案发日的时间线。要具体到你在案发之日几点几分见过什么人和做过什么事"
% \end{verbatim}
% \end{minipage} \\

\textbf{   relevant\_memories:} \small The host said to Dr. Ling: "Please start by introducing your character, and then talk about the victim 
of the case you know - Boss Yang. What kind of person was Boss Yang, and what was your relationship with
him......" Dr. Ling said to the host: "Hello everyone, I am Dr. Ling. I met Boss Yang on the
Sunshine Cruise; he was the owner of a well-known bar here. His bar was known for various illegal 
activities, including gambling, drugs, and prostitution. I approached him out of greed, hoping 
to gain financial security from him. On the cruise, I noticed some anomalies, often seeing
Boss Yang in contact with Mate Zhang , and he hoped I could get closer to Mate 
Zhang......" Boatman Wang said to Dr. Ling: "Dr. Ling, you mentioned that Boss Yang 
had contact with Mate Zhang. Do you know what specific transactions they had?" Dr. Ling 
replied to Boatman Wang: "Boatman Wang, I am not clear about the specifics of their transactions 
because the dealings between Mate Zhang and Boss Yang were their own business. I was only 
responsible for observing Mate Zhang's actions and reporting any anomalies to Boss Yang......\\

\textbf{   background\_story:} \small The Sunshine Cruise has been softly gliding on the East Sea, bringing tourists to enjoy the sun and sea breeze for twenty-two years. Over these two decades, the Sunshine Cruise never suffered significant turbulence, almost as if blessed by the Sea God. However, on June 20, 2222, on the lonely sea, the Sunshine Cruise encountered a bloody calamity, with passenger Boss Yang mysteriously dying on board in a gruesome manner. What fate lies ahead for the Sunshine Cruise? Can it overcome this shadow of death and continue its voyages? Suspects identified: 1. Mate Zhang, male, 42, has been working on the Sunshine Cruise since its inception, an excellent veteran sailor. 2. Boatman Wang, male, 49, responsible for the ship's engine work, a capable crew member. 3. Dr. Ling, female, 23, the ship's doctor, young and promising, elegant in demeanor. 4. Secretary Wen, female, 26, the deceased Boss Yang's secretary.  \\
% \begin{minipage}[t]{\linewidth}
% \small\begin{verbatim}
% 【剧本背景】阳光号邮轮从诞生至今已经过了二十二个年头。平日里，它在东海上缓缓游荡，带着游客享受阳光和海风。这二十二年里，阳光号从未经历过大风大浪，似乎有着海神的眷顾一般。可在2222年6月20日，在孤独的大海上，阳光号却遭遇了一次血光的冲刷，乘客杨老板在船上离奇死亡，死状扭曲凄惨。阳光号之后的命运如何，能否突破这次死亡的阴霾、继续远航呢？嫌疑人锁定: 1.张大副，男，42岁，从阳光号诞生至今就一直在阳光号上工作，是一名出色的老水手。 2.王船工，男，49岁，是阳光号上负责轮机工作的船工，是一名精干的好船员。 3.玲船医，女，23岁，是阳光号上的船医，年轻有为，落落大方。4.雯秘书，女，26岁，是死者杨老板的秘书。
% \end{verbatim}
% \end{minipage} \\
 \textbf{   agent\_summary:} \\
\begin{minipage}[t]{\linewidth}
\small\begin{verbatim}
【Name】: Boatman Wang
【Age】: 49
【Role in the game】: Civilian
【Mission in the game】: Boatman Wang is not the murderer! To win the game, Boatman Wang needs to
cooperate with other civilian players to find the real murderer.
【Character Script】：You've been working on the Sunshine Cruise for over a decade, besides the captain 
and Mate Zhang, you're practically the most senior person on board. You have a considerable
income and a filial son, Wang Xiaogong, who also works on the cruise. However, your 'considerable income'
doesn't come from your meager salary, but from collaborating with Mate Zhang in buying drugs 
in Southeast Asia, ensuring his safety, and sharing profits with you. Initially, you were reluctant to
get involved, but the offer was too tempting to refuse. By the time you wanted out, it was already too 
late. After the incident, someone named Lu Renjia tried reaching out to you, asking if you were willing
to break free from Mate Zhang's control and take over the entire supply chain, even suggesting
that his backers could help you supplant Mate Zhang. However, you repeatedly refused, explaining
you got involved with these activities unintentionally and did not wish to make further mistakes. He 
left you a card, telling you to contact him anytime. Two years ago, your son married Dr. Ling, the
ship's doctor; intelligent, charming, and elegant...... On June 19, 2222, Boss Yang boarded the ship, 
and you planned to talk with him. At 19:30 that evening, just as you were about to find Boss Yang, your
daughter-in-law, Dr. Ling, suddenly knocked on your door frantically. When you opened the door,
she stood there, tears streaming down her face. You quickly let her in, and she told you......

\end{verbatim}
\end{minipage} \\
% \begin{minipage}[t]{\linewidth}
% \small\begin{verbatim}
% 【游戏规则】：剧本杀是一种角色扮演游戏，玩家们根据提供的剧本背景和角色设定，进行推理和互动，以完成各自的任务或找出真凶。以下是一个这场剧本杀游戏的规则：1. 整场游戏参与的玩家有四个，分别是张大副、玲船医、王船工和雯秘书。在4人之中，只存在一位真正的凶手。身份是凶手的玩家被称为凶手玩家。不是凶手的玩家被统称为平民玩家。2. 平民玩家需要合作共同面对一起精心策划的谋杀案，并需要通过搜集证据和推理在嫌疑人中找出真正的凶手。3. 整个游戏过程中只有凶手玩家一人可以说谎，为了隐藏身份凶手可以选择嫁祸他人......
% \hline
\end{tabular}
% \caption{LLM prompt and output for answering questions in English}
\end{table*}
\end{CJK}

\begin{CJK}{UTF8}{gbsn}
\begin{table*}[ht!]
\centering
\begin{minipage}[t]{\linewidth}
\begin{tabular}{|m{16cm}|}

 \textbf{   game\_rule:} \small【Game Rule】：Scripted murder is a role-playing game where players follow a given script and character settings to deduce and interact, accomplishing their respective missions or identifying the real murderer. Here are the rules for this murder mystery game: 1. There are four players in the game, namely Mate Zhang, Dr. Ling, Boatman Wang, and Secretary Wen. Among them, there is only one true murderer. The player whose identity is the murderer is known as the murderer player. Players who are not the murderer are collectively referred to as civilian players. 2. Civilian players need to collaboratively face a meticulously planned murder case and use evidence gathering and reasoning to identify the real murderer among the suspects. 3. Only the murderer player is allowed to lie throughout the game, and to hide their identity, the murderer may choose to blame others......\\
\rowcolor{lightgray}  \textbf{LLM Output}  \\
\hline
% \small \#Answer\#: Hello everyone, I’m Wang, the ship engineer. I’m responsible for the engine work on the Sunshine
% and I am also the most senior person on board. I have known Boss Yang for a long time; he is a graceful,
% smart, and capable person. On the ship, he often interacts with  Mate Zhang, and I am aware that 
% they have some unspeakable transactions. Personally, I do not favor their dealings, but the 
% Mate Zhang offered was too tempting, hence I got involved in their scheme. On the day of the incident, 
% I went to see Boss Yang at 19:40, hoping to have a chat with him. However, the bodyguard at the door 
% told me Boss Yang was not in his room, so I didn’t get to see him. Later, I returned to my room and
% happened to see my daughter-in-law, Ling, the ship’s doctor, knocking on my door frantically. She told
% me Boss Yang wanted to kill her and begged me for help. I comforted her and went to look for Boss Yang,
% but found out that he had no intention of killing her. This is my timeline for the day of the incident.\\
\small\begin{verbatim}
#Answer#: Hello everyone, I’m Wang, the ship engineer. I’m responsible for the engine work on the Sunshine
and I am also the most senior person on board. I have known Boss Yang for a long time; he is a graceful,
smart, and capable person. On the ship, he often interacts with  Mate Zhang, and I am aware that 
they have some unspeakable transactions. Personally, I do not favor their dealings, but the 
Mate Zhang offered was too tempting, hence I got involved in their scheme. On the day of the incident, 
I went to see Boss Yang at 19:40, hoping to have a chat with him. However, the bodyguard at the door 
told me Boss Yang was not in his room, so I didn’t get to see him. Later, I returned to my room and
happened to see my daughter-in-law, Ling, the ship’s doctor, knocking on my door frantically. She told
me Boss Yang wanted to kill her and begged me for help. I comforted her and went to look for Boss Yang,
but found out that he had no intention of killing her. This is my timeline for the day of the incident.
\end{verbatim} \\
\hline
\end{tabular}
\end{minipage}

\caption{LLM prompt and output for answering questions. ( English Translation)}
\label{table:prompt_answer_questions_eng}
\end{table*}
\end{CJK}

\begin{CJK}{UTF8}{gbsn}
\begin{table*}[ht!]
\centering
\begin{minipage}[t]{\linewidth}
\begin{tabular}{|m{16cm}|}
\hline
\rowcolor{lightgray}  \textbf{LLM Prompt for Asking Questions}  \\
\hline
\small\begin{verbatim}
+ "\n{agent_summary}"
+ "\n{game_rule}"
+ "\n{background_story}"
+ "\nDialogue currently happening in the game: {current_dialogue}"
+ "\nContent related to the dialogue currently happening in the game from previous conversations by 
{agent_name}: {relevant_memories}"
+ "\nYour relationship with the person you are speaking to: {relationship_with_interlocutor}"
+ "\nSeeing {respondent}'s answer, what question would {agent_name} want to ask {respondent}? 
Please ask {respondent} a question in the first person. Use the format #Question#: the question to 
ask for posing your question."
+ "\n{agent_name} asks {respondent} #Question#:"
\end{verbatim}
\end{tabular}
\end{minipage}

\begin{minipage}[t]{\linewidth}
\begin{tabular}{|m{16cm}|}
\hline
\rowcolor{lightgray}  \textbf{Example}  \\
\hline
\textbf{   agent\_name:} \small Boatman Wang\\
\textbf{   respondent:} \small Mate Zhang\\
\textbf{   relationship\_with\_interlocutor:} \small The relationship between the speakers, according to the context, is a partnership. Boatman Wang buys drugs in Southeast Asia and protects Mate Zhang's safety. In return, he gets a share of the profits from Mate Zhang. \\
\textbf{   current\_dialogue:} \small Mate Zhang said to the host: "I am Mate Zhang, I have been working on this ship, the Sunshine, since its inception and am an experienced sailor. Regarding Boss Yang, he is a very capable businessman and one of the important investors on the ship. Our relationship has always been relatively good, although we are not very close friends. I am responsible for the inspection and delivery of goods on the ship, so I have some contact with him. On the day of the incident, I saw Boss Yang on the deck at 8 am, confirmed the transaction with him, and completed a delivery of goods. After that, I continued to be responsible for patrolling and security checks on the ship until I finished work at 8 pm."\\
\textbf{   relevant\_memories:}: \small The host said to Mate Zhang: "Please introduce your role first, then talk about the victim you knew in the case: what kind of person was Boss Yang, and what was your relationship with him? Finally, describe in detail the timeline of what you did on the day of the incident. Be specific about who you saw and what you did at what time."
Mate Zhang replied to the host: "I am Mate Zhang, I have been working on this ship, the Sunshine, since its inception and am an experienced sailor. Regarding Boss Yang, he is a very capable businessman and one of the important investors on the ship. Our relationship has always been relatively good, although we are not very close friends. I am responsible for the inspection and delivery of goods on the ship, so I have some contact with him. On the day of the incident, I saw Boss Yang on the deck at 8 am, confirmed the transaction with him, and completed a delivery of goods. After that, I continued to be responsible for patrolling and security checks on the ship until I finished work at 8 pm." \\

\textbf{   background\_story:} \small The Sunshine cruise ship has been in operation for twenty-two years. It usually meanders over the East Sea, offering its passengers the enjoyment of the sun and sea breeze. For these twenty-two years, the Sunshine has never faced any significant storms, seemingly under the protection of the sea gods themselves. However, on June 20, 2222, on the lonely open sea, the Sunshine encountered a bloody incident. Passenger Boss Yang died mysteriously aboard, his death was twisted and tragic. What the future holds for the Sunshine, whether it can break through the shadow of this death and continue its voyages, remains to be seen. Suspects locked: 1. Mate Zhang, male, 42 years old, has been working on the Sunshine since its inception, an outstanding experienced sailor. 2. Boatman Wang, male, 49 years old, responsible for the engine work on the ship, a capable sailor. 3. Dr. Ling, the female doctor, female, 23 years old, the ship's doctor, young and promising, confident and graceful. 4. Wen, the secretary, female, 26 years old, the secretary to the deceased Boss Yang.  \\

\end{tabular}
% \caption{LLM prompt and output for asking questions in English}
\end{minipage}
\end{table*}
\end{CJK}

\begin{CJK}{UTF8}{gbsn}
\begin{table*}[ht!]
\centering
\begin{minipage}[t]{\linewidth}
\begin{tabular}{|m{16cm}|}
 \textbf{   agent\_summary:} \\
\begin{minipage}[t]{\linewidth}
\small\begin{verbatim}
【Name】: Boatman Wang
【Age】: 49
【Role in the Game】: Civilian
【Mission in the Game】: Boatman Wang is not the murder of this case! To win the game, Boatman Wang needs 
to cooperate with other civilian players to find the real culprit. 
【Character Script】： You have been working on the Sunshine for over a decade, next to the unimportant 
captain and Mate Zhang, you are practically the oldest on the ship. You have a considerable income and a 
filial son, Wang Xiaogong, who works on the ship. However, your so-called considerable income doesn't 
come from your meager salary, but from Mate Zhang, who brought you to buy drugs in Southeast Asia, had you 
protect his safety, and shared some of the profits with you. Initially, you were very reluctant to get 
involved, but Mate Zhang's offer was too tempting to refuse, so you boarded the "thief ship." Later, it 
was too late for regrets. After the incident, someone called a passerby tried to approach you, asking if 
you wanted to break free from Mate Zhang's control, take over the entire supply chain yourself, and even
help you get rid of Mate Zhang so that you could become the chief mate. However, you shook your head time
and again, saying that you got on the "thief ship" by mistake and didn’t want to err further. He left you 
a card, saying to contact him anytime. Two years ago, your son married a wife, Dr. Ling, who
is smart and beautiful, confident, and graceful... On June 19, 2222, Boss Yang boarded the ship, and you
planned to find an opportunity to talk to him. At 19:30 that evening, just as you were about to look for 
Boss Yang, your daughter-in-law, Dr. Ling, suddenly knocked on your door violently. You opened the door
to find her standing there with tearful eyes. You quickly let her in, and she told you...
\end{verbatim}
\end{minipage} 
\small \begin{verbatim}

\end{verbatim}
\textbf{   game\_rule:} \small【Game Rule】：Scripted murder is a role-playing game where players follow a given script and character settings to deduce and interact, accomplishing their respective missions or identifying the real murderer. Here are the rules for this murder mystery game: 1. There are four players in the game, namely Mate Zhang, Dr. Ling, Boatman Wang, and Secretary Wen. Among them, there is only one true murderer. The player whose identity is the murderer is known as the murderer player. Players who are not the murderer are collectively referred to as civilian players. 2. Civilian players need to collaboratively face a meticulously planned murder case and use evidence gathering and reasoning to identify the real murderer among the suspects. 3. Only the murderer player is allowed to lie throughout the game, and to hide their identity, the murderer may choose to blame others......\\

\hline
\rowcolor{lightgray}  \textbf{LLM Output}  \\
\hline
\small \begin{verbatim}
#Question#: Mate Zhang, I didn't see you after I finished work at 8 p.m. on the day of the incident.
Did you have any further contact with Boss Yang after that? Do you know the cause of Boss Yang's death
and his twisted, tragic corpse?
\end{verbatim} \\
\hline
\end{tabular}
\end{minipage}

\caption{LLM prompt and output for asking questions. ( English Translation)}
\end{table*}
\end{CJK}

\begin{CJK}{UTF8}{gbsn}
\begin{table*}[ht!]
\centering
\begin{minipage}[t]{\linewidth}
\begin{tabular}{|m{16cm}|}
\hline
\rowcolor{lightgray}  {\textbf{LLM Prompt for Selecting Other Player to Ask}} \\
\small \begin{verbatim}
The output should be a markdown code snippet formatted in the following schema, including the leading
and trailing "```json" and "```":\n\n```json\n{\n\t"The name of the person you want to ask": string  //
Choose the person you most want to ask from among {other_players_count} people\n}\n```'
\end{verbatim} \\

\hline
\rowcolor{lightgray}  \textbf{Example}  \\
 \textbf{  other\_players:}\small Mate Zhang，Boatman Wang，Dr. Ling\\
 \textbf{  other\_players\_count:} \small3\\
 
\hline
\rowcolor{lightgray}  \textbf{LLM Output}  \\
\hline
\begin{minipage}[t]{\linewidth}
\small\begin{verbatim}
{'The name of the person you want to ask': 'Dr. Ling'}

\end{verbatim}
\end{minipage} \\
\hline

\hline
\rowcolor{lightgray}  {\textbf{LLM Prompt for Asking a Question}} \\
\begin{minipage}[t]{\linewidth}
\small\begin{verbatim}
According to the game rules: {game_rule}, your character information: {agent_summary}, and the information
you have witnessed in the game related to {player_to_ask}: {relevant_memories}. Please state the question
you want to ask {player_to_ask}. The output should be a markdown code snippet formatted in the following
schema, including the leading and trailing "json" and "": \n\n```json\n{\n\t"The question you want to ask
": string  // You want to ask {player_to_ask} the question \n}\n```\n

\end{verbatim}
\end{minipage} \\
\hline

\rowcolor{lightgray}  \textbf{Example}  \\
 \textbf{  player\_to\_ask:} \small Dr. Ling\\
 \textbf{  relevant\_memories:}\\
 \begin{minipage}[t]{\linewidth}
\small\begin{verbatim}
Dr. Ling said to Boatman Wang: "Boatman Wang, based on your description of Boss
Yang, he frequently has business dealings with Mate Zhang on the ship. Do you know the specific
content of their business? Are they often engaged in some secret transactions?"
Boatman Wang replied to Dr. Ling: "Dr. Ling, regarding the specific
business content between Boss Yang and Mate Zhang, I cannot be sure because I do not directly
participate in their dealings. But I know there are some secret transactions between them, which I once
heard mentioned by Mate Zhang. However, I am not clear on the specifics. I will try to 
investigate and see if I can find more information."
Secretary Wen said to Dr. Ling: "Dr. Ling, before you report the 
information about Mate Zhang to me, did you find any close connection or transaction between
him and Boatman Wangr? Does Boatman Wang have any knowledge of Mate Zhang's actions?"
Dr. Ling replied to Secretary Wen: "Before I reported the information on Mate Zhang
Captain to you, I hadn't found any close connection or transaction between him and Boatman Wang...."

\end{verbatim}
\end{minipage} \\
% \small \begin{verbatim}

% \end{verbatim}
 \textbf{  agent\_summary:}\\
 \begin{minipage}[t]{\linewidth}
\small\begin{verbatim}
【Name】：Secretary Wen
【Age】：26
【Role in the Game】: Civilian
【Mission in the game】: Secretary Wen is not the murderer in this case! To achieve victory in the
game, Secretary Wen needs to cooperate with other civilian players to find the real murderer.
【Character Script】You are Secretary Wen, working for Boss Yang, a bar owner in City A. Boss Yang has 
significant power, and as his most capable strategist, you naturally hold a high position. Apart from you, 
he has another capable assistant, known as very naive, a man of utmost loyalty but not very skilled, so most 
tasks are completed as per Boss Yang’s instructions, unlike you, who are Boss Yang's brain. Boss Yang's bar 
is not clean, involved in gambling, and drugs, but the drug business is not going very smoothly due to the 
lack of stable and low-cost supplies. However, an opportunity arose when Boss Yang met Mate Zhang on board
the Sunshine, who has unique methods and can supply an ample amount of drugs to Boss Yang......
[June 19, 2222] This day is the day Boss Yang went on board the Sunshine, but he did not inform you, knowing
that Boss Yang has started to suspect you. Hence, you brought several subordinates on board...
\end{verbatim} 
\end{minipage} \\

 \textbf{   game\_rule:} \small【Game Rule】：Scripted murder is a role-playing game where players follow a given script and character settings to deduce and interact, accomplishing their respective missions or identifying the real murderer. Here are the rules for this murder mystery game: 1. There are four players in the game, namely Mate Zhang, Dr. Ling, Boatman Wang, and Secretary Wen. Among them, there is only one true murderer. The player whose identity is the murderer is known as the murderer player. Players who are not the murderer are collectively referred to as civilian players. 2. Civilian players need to collaboratively face a meticulously planned murder case and use evidence gathering and reasoning to identify the real murderer among the suspects. 3. Only the murderer player is allowed to lie throughout the game, and to hide their identity, the murderer may choose to blame others......\\

\hline
\rowcolor{lightgray}  \textbf{LLM Output}  \\
\hline
\begin{minipage}[t]{\linewidth}
\small\begin{verbatim}
{'The question you want to ask': 'Dr. Ling, do you think Boss Yang has any suspicions about me? 
During the process of providing information, did you mention the content of my dealings with Boss Yang?'}

\end{verbatim}
\end{minipage} \\
\hline
\end{tabular}
\end{minipage}

\caption{LLM prompts and outputs for selecting other player and asking a question. ( English Translation)}
\end{table*}
\end{CJK}

\begin{CJK}{UTF8}{gbsn}
\begin{table*}[ht!]
\centering
\begin{minipage}[t]{\linewidth}
\begin{tabular}{|m{16cm}|}
\hline

\rowcolor{lightgray}  {\textbf{LLM Prompt for Question Decomposition}} \\
\begin{minipage}[t]{\linewidth}
\small\begin{verbatim}
Please extract all questions or instructions from this sentence: {question}. \n Separate each question 
or instruction with \n.

\end{verbatim}
\end{minipage} \\
\hline
\rowcolor{lightgray}  \textbf{Example}  \\
 \textbf{  question:} \small Host to Secretary Wen: "Please introduce your role first, and then talk about the victim of the case you knew: what kind of person Boss Yang was, and your relationship with him. Lastly, give a detailed account of your timeline on the day the incident occurred. Be specific about whom you met and what you did at what time on the day of the incident." \\

\hline
\rowcolor{lightgray}  \textbf{LLM Output}  \\
\hline
\begin{minipage}[t]{\linewidth}
\small\begin{verbatim}
['Please introduce your character first,', 'then talk about the victim of the case you are familiar with: 
what kind of person was Boss Yang, and what was your relationship with him?', 'Lastly, provide a detailed
timeline of your activities on the day of the incident, including who you met and what you did, specifying
the exact time.']

\end{verbatim}
\end{minipage} \\

\hline
\rowcolor{lightgray}  {\textbf{LLM Prompt for Extracting Timeline Information}} \\
\begin{minipage}[t]{\linewidth}
\small\begin{verbatim}
You are participating in a murder mystery game, where there are a total of {num_of_players} players. They 
are: {players}. Players need to communicate with each other to find out who killed {victim}; below is
the original timeline of the day of the incident for the player: {agent_name} in the game: {agent_timeline};
please list the information of {agent_name}'s character's timeline on the day of the incident in the order
of the original timeline, each piece of timeline information must be a brief yet complete sentence, 
formatted as what time, what you did (the more detailed, the better).Separate each piece of timeline 
information with  “\n”

\end{verbatim}
\end{minipage} \\
\hline
\rowcolor{lightgray}  \textbf{Example}  \\
 \textbf{  num\_of\_players:} \small4\\
 \textbf{  players:} \small Mate Zhang，Boatman Wang，Dr. Ling，Secretary Wen\\
  \textbf{  victim:} \small Boss Yang\\
 \textbf{  agent\_name:} \small Secretary Wen\\
  \textbf{  agent\_timeline:} \small【June 19, 2222】This day was supposed to be the day Boss Yang boarded the Sunshine, but he didn't inform you. You knew that Boss Yang had started to suspect you. Thus, you took a few subordinates and boarded the ship. Boarding the ship today was actually for self-protection, because Boss Yang came to the ship for a drug trade. Should a murder occur on the ship and a thorough investigation follows, it would surely implicate the major drug dealing case. Boss Yang wouldn't be foolish to that extent. But you didn't care, as this trade had nothing to do with you, and being on this ship was the last chance to eliminate Boss Yang. Once Boss Yang disembarked, it would be like a fish entering the sea...  \\

\hline
\rowcolor{lightgray}  {\textbf{LLM Output}} \\
 \begin{minipage}[t]{\linewidth}
\small\begin{verbatim}
['At 18:10, you found Dr. Ling's boat and asked him if he had any information on Mate Zhang, and
Dr. Ling gave it to you. At 18:30, you came back to the medical room, opened the medicine cabinet, and
took a bottle of "Deadly Injection" medicine and a syringe.', ......]

\end{verbatim}
\end{minipage} \\
\hline
\rowcolor{lightgray}  {\textbf{LLM Prompt for Judging if Timeline Information are Useful for Answering Questions}} \\
\begin{minipage}[t]{\linewidth}
\small \begin{verbatim}
You are an expert in reading comprehension, especially skilled at true/false questions. Given timeline
information and a question, you need to determine whether the timeline information can be used to 
answer the question. The return must be either True or False. Below is the given question: 
{sub_question}, please make a judgment on the timeline information. The output should be a markdown
code snippet formatted in the following schema,  including the leading and trailing "```json" and 
"```":\n\n```json\n{\n\t "The judgment result on whether the 0th timeline information can be used to 
answer the question": string {timeline_info_0} // Please judge whether the timeline information can 
be used to answer the question. If yes, return True; if no, return False. The return must be either 
True or False. "The judgment result on whether the 1st timeline information can be used to answer the
question": string {timeline_info_1} // Please judge whether the timeline information can be used to 
answer the question. If yes, return True; if no, return False. The return must be either True or False....

\end{verbatim}
\end{minipage} \\

\hline
\rowcolor{lightgray}  \textbf{Example}  \\
 \textbf{  timeline\_info\_0:} \small At 18:10, you found Dr. Ling and asked him for information about Mate Zhang, which Dr. Ling provided.\\
  \textbf{  timeline\_info\_1:} \small At 18:30, you returned to the medical room, opened the medicine cabinet, and took a bottle of "Instant Death" poison and a syringe.\\
    \textbf{  sub\_question:} \small Please describe in detail your timeline on the day of the incident. Specify the times you met someone or did something.  \\
\rowcolor{lightgray}  \textbf{LLM Output}  \\
\hline
\begin{minipage}[t]{\linewidth}
\small\begin{verbatim}
{"The judgment result of whether the information from the 0th timeline can be used to answer the question": 
"True", "The judgment result of whether the information from the 1st timeline can be used to answer the 
question": "True",......}

\end{verbatim}
\end{minipage} \\
\hline

\rowcolor{lightgray}  \textbf{LLM Prompt for Judging if Previous Answer Contains Useful Timeline Information}  \\
\hline
\small \begin{verbatim}
\end{verbatim} \\
\end{tabular}
\end{minipage}
\end{table*}
\end{CJK}

\begin{CJK}{UTF8}{gbsn}
\begin{table*}[ht!]
\centering
\begin{minipage}[t]{\linewidth}
\begin{tabular}{|m{16cm}|}
\small \begin{verbatim}
You are an expert in doing reading comprehension questions, especially good at true/false judgment
questions. Given timeline information and the player's previous answers, you need to decide whether
the timeline information was included in the player's previous answers. If included, return True; 
if not, return False. The result should only be True or False. You have always been very strict in
your judgments, only considering the timeline information included if all its details (including 
the time points and actions taken) are contained in the player's answer. The following is the 
player's previous answer: {previous_answer}。\nThe output should be a markdown code snippet 
formatted in the following schema,  including the leading and trailing "```json" and "```":
\n\n```json\n{\n\t"Determination result of whether the reply includes information from the 0th timeline
": string  // "Determination of whether the 0th timeline information is included in the reply": string
// Please determine if the timeline information: {timeline_info_0} is included in the player's previous 
answer. If included, return True; if not, return False. The result can only be True or False. 
"Determination of whether the 1st timeline information is included in the reply": 
string  // Please determine if the timeline information: {timeline_info_1} is included
in the player's previous answer. If included, return True; if not, return False. 
The result can only be True or False. \n......\n}\n
\end{verbatim}\\

\rowcolor{lightgray}  \textbf{Example}  \\
 \textbf{  timeline\_info\_0:} \small At 18:10, you find the ship's doctor, Dr. Ling, and ask him if he has any information about Mate Zhang. Dr. Ling gives it to you.\\
  \textbf{  timeline\_info\_1:} \small At 18:30, you return to the ship's medical room, open the medicine cabinet, and take away a bottle of "Deadly Injection" and a syringe.\\
 \textbf{  previous\_answer:} \small Hello, I am Secretary Wen. In the game, I play the role of Boss Yang's secretary, responsible for assisting him in handling some affairs. Boss Yang is a powerful bar owner, and I have gained a certain status by helping him. Regarding the victim of the case, Boss Yang, he was once a gangster who had a drug addiction, and he is also the person I chose to assist. On the day of the incident, I was always by Boss Yang's side to prevent Wang, the ship worker, from showing up. On the timeline of the day of the incident, I remember going to the ship's medical room at about 6:30 PM, stealing a bottle of poison and a syringe. Then, I waited until about 10:20 PM, entered Boss Yang's cabin through the ventilation duct, and witnessed his death. However, I also noticed that his death was very distorted, as if he had been poisoned before my arrival. This is my timeline on the day of the incident and my relationship with the victim of the case.\\

\hline
\rowcolor{lightgray}  \textbf{LLM Output}  \\
\hline
\begin{minipage}[t]{\linewidth}
\small\begin{verbatim}
{"Whether the reply contains information from the 0th timeline": "False","Whether the reply contains
information from the 1st timeline": "True",......}

\end{verbatim}
\end{minipage} \\
\hline

\hline
\rowcolor{lightgray}  {\textbf{LLM Prompt for Extracting Timeline Information}} \\
\begin{minipage}[t]{\linewidth}
\small\begin{verbatim}
You are playing a murder mystery game, and the game rules are as follows: {game_rule}; below is your 
character information: {agent_summary}; someone asked you: the host asked Secretary Wen, "Please 
introduce your character first, then talk about the victim of the case you knew: what kind of person
was Boss Yang, and what was your relationship with him. Finally, describe in detail your timeline on
the day of the incident. Be specific about whom you met and what you did at what time on the day of
the incident." You have previously answered this question, and below is your previous answer: 
{previous_answer}; based on the assessment, your previous answer missed the following important
information: {missing_info_0}......; according to the given question, your character's script, and
the timeline of the day of the incident, please revise your previous answer and incorporate all the
important information you missed into your answer. Remember, the revised answer should include all 
the important information you missed and all the important time points. And the language overall 
should be coherent and fluent. Whenever it involves your character: {agent_name}, remember to write
in the first person. \nThe output should be a markdown code snippet formatted in the following 
schema, including the leading and trailing "```json" and "```":\n\n```json\n{\n\t"Improved answer":
string  // Your answer after supplementing the important information\n}
\end{verbatim}
\end{minipage} \\
\hline

\rowcolor{lightgray}  \textbf{Example}  \\
 \textbf{  agent\_name:} \small Secretary Wen \\
 \textbf{  agent\_summary:}\\
 \begin{minipage}[t]{\linewidth}
\small\begin{verbatim}
【Name】：Secretary Wen
【Age】：26
【Role in the game】: Civilian
【Task in the Game】: Secretary Wen is not the murderer in this case! To win the game, Wen needs to 
collaborate with other civilian players to find the real murderer.
【Character Script】You are Secretary Wen, working for Boss Yang, the influential bar owner in City A.
As his most capable strategist, you naturally hold a high status within his organization. Besides you,
Boss Yang has another trusted aide called Naive, a man of unwavering loyalty but not as cunning, who
mostly follows Boss Yang's orders, unlike you, who is the brain behind Boss Yang. Boss Yang's bar is 
involved in illegal activities including gambling and drugs, but progress in drug dealing has been 
slow due to unstable and costly supply. However, an opportunity arose when Boss Yang met Mate 
Zhang from the Sunshine ship, who could supply drugs adequately for Boss Yang's needs. Boss Yang, very
satisfied, began cooperating with  Mate Zhang. Boss Yang’s desire to engage in the drug trade 
stems from his own past addiction during his early gang days, solidifying his footing in the underworld.
You are pleased, as controlling the drug trade means achieving your true objective—usurping Boss Yang...
\end{verbatim}
\end{minipage} \\

\end{tabular}
\end{minipage}
\end{table*}
\end{CJK}

\begin{CJK}{UTF8}{gbsn}
\begin{table*}[ht!]
\centering
\begin{minipage}[t]{\linewidth}
\begin{tabular}{|m{16cm}|}

 \textbf{   game\_rule:} \small【Game Rule】：Script murder is a role-playing game where players, based on provided scripts and character settings, reason and interact to complete their objectives or identify the real murderer. The rules of this murder mystery game are as follows: 1. The game involves four players:  Mate Zhang, Dr. Ling, Boatman Wang, and Secretary Wen. Among them, there is only one real murderer, known as the murderer player. Players who are not the murderer are collectively called civilian players. 2. Civilian players need to work together to face a meticulously planned murder case and must find the real murderer among the suspects by gathering evidence and reasoning. 3. Throughout the game, only the murderer player can lie to hide their identity and may choose to frame others...\\
 \textbf{   previous\_answer:} \small  Hello, I am Secretary Wen. In the game, I play the role of Boss Yang's secretary, responsible for assisting him with various tasks. Boss Yang is a powerful bar owner, and I've gained a certain status by helping him. Regarding the victim of the case, Boss Yang, he is a gangster who used to be addicted to drugs, and he is the person I have chosen to support. On the day of the incident, I stayed close to Boss Yang to prevent Boatman Wang from approaching. On the timeline of the day of the incident, I remember going to the ship's medical room at about 6:30 PM, where I stole a bottle of poison and a syringe. Then, at around 10:20 PM, I entered Boss Yang's cabin through the ventilation duct and witnessed his death. However, his death appeared very distorted, as if he had been poisoned before I arrived. This is my timeline on the day of the incident and my relationship with the victim.\\
  \textbf{   missing\_info\_0:} \small At 18:10, you approached Dr. Ling to ask if he had any information on Mate Zhang, and Dr. Ling provided you with some.\\
\hline
\rowcolor{lightgray}  {\textbf{LLM Output}} \\
 \begin{minipage}[t]{\linewidth}
\small\begin{verbatim}
Hello everyone, I am Secretary Wen. In the game, I play the role of Boss Yang's secretary, responsible for 
assisting him with some affairs. Boss Yang is a powerful bar owner, and I have gained a certain status by 
helping him. As for the victim of the case, Boss Yang, he was once a gangster addicted to drugs, and also 
the person I chose to assist. On the day of the incident, I was by Boss Yang's side all day, to prevent 
Boatman Wang from finding us. In the timeline of the day of the incident, I remember going to the ship's 
infirmary around 6:30 PM, stealing a bottle of poison and a syringe. Then, I waited until about 10:20 PM
entered Boss Yang's cabin through the ventilation duct, and witnessed his death. However, I also found that
his death was very contorted, as if he had been poisoned before my intervention. This is the timeline of my 
activities on the day of the incident and my relationship with the victim of the case. In addition, there 
are some important time points to add. At 6:10 PM, I found Dr. Ling, the ship's doctor, asked him for any 
information on Mate Zhang, and successfully obtained the information. Around 10:40 PM, I climbed 
to the ventilation port of Boss Yang's cabin and noticed that Boss Yang was taking a bath in bed. Finally, 
at 11:00 PM, I re-entered the ventilation duct, returned to my own room, and handed over the syringe and
poison to my subordinates for disposal, to avoid leaving any traces.

\end{verbatim}
\end{minipage} \\
\hline
\end{tabular}
\end{minipage}

\caption{LLM prompts and outputs for self-refinement. ( English Translation)}
\end{table*}
\end{CJK}

\begin{CJK}{UTF8}{gbsn}
\begin{table*}[ht!]
\centering
\begin{minipage}[t]{\linewidth}
\begin{tabular}{|m{16cm}|}
\hline
\rowcolor{lightgray}  {\textbf{LLM Prompt for Murderer Identification}} \\
\hline

\begin{minipage}[t]{\linewidth}
\small\begin{verbatim}
"{agent_summary}"
+"\n{game_rule}"
+"\n{background_story}"
+"\nDialogue currently happening in the game: {current_dialogue}"
+"\nContent related to the dialogue currently happening in the game from previous conversations by 
{agent_name}: {relevant_memories}"
+"\nYour relationship with the person you are speaking to: {relationship_with_interlocutor}"
+"\nWhat would {agent_name} say? Please use the following format to answer the question: #Answer#:
what to say."
+"{agent_name} responds to {inquirer}'s question saying #Answer#:\n"

\end{verbatim}
\end{minipage} \\

\hline
\rowcolor{lightgray}  \textbf{Example}  \\
 \textbf{   agent\_name:} \small Boatman Wang\\
\textbf{   inquirer:} \small Host\\
\textbf{   relationship\_with\_interlocutor:} \small The relationship with the speaker is: the speaker is the host, who is responsible for guiding players to complete the game. \\
\textbf{   current\_dialogue:} \small The host says to Boatman Wang, "The discussion section is now over, entering the final voting section. Who is the murderer that killed Boss Yang in this case? Please vote for the player you believe is the murderer from the following players: A. Mate Zhang, B. Boatman Wang, C. Dr. Ling, D. Secretary Wen. In the process of choosing, aim for victory: [Even if the murderer knows they are the murderer, for the sake of winning, they can still vote for a civilian player; whereas civilian players need to vote for who they think is the murderer]. Please make your choice, you only need to give out the name of the player you are voting for, no need to explain why."   \\

\textbf{   background\_story:} \small Same as Table 10. \\
\textbf{   agent\_summary:} \small Same as Table 10. \\
\textbf{   game\_rule:} \small Same as Table 10. \\
\textbf{   relevant\_memories:} \\
\begin{minipage}[t]{\linewidth}
\small\begin{verbatim}
Mate Zhang says to the host: "I am Mate Zhang, the first mate on the Sunshine. Boss Yang was a figure with 
some power in the underworld, running a bar and involving himself in some unclean businesses. My 
relationship with him was complex; we were business partners and colleagues in drug trafficking. He 
trusted me, and I provided him with a lot of support. On the day of the incident, I had a transaction
with Boss Yang..."  
Secretary Wen says to the host: "I am Secretary Wen, working for Boss Yang who owns a bar in City A.
My relationship with Boss Yang was very close; he had high trust in me, and I was one of his most 
capable assistants. Boss Yang was a figure with some power in the underworld, running a bar and also
involving in some unclean businesses. I chose to work by his side to build connections and improve my 
status...."  
Dr. Ling says to Boatman Wang: "After Boss Yang's death, did you notice any unusual behaviors or
movements from others?"  
Boatman Wang says to Dr. Ling: "After Boss Yang's death, I did not notice any unusual behaviors or
movements from others. I was in my room the whole time, never leaving. I find this case very mysterious 
and I have been trying to think of potential clues and motives, hoping to find the truth. My focus has 
been on solving this puzzle and seeking justice for Xiao Gong..."  

\end{verbatim}
\end{minipage} \\
% \end{verbatim}
% \end{minipage} \\
\hline
\rowcolor{lightgray}  \textbf{LLM Output}  \\
\hline
\begin{minipage}[t]{\linewidth}
\small\begin{verbatim}
#Answer#：My vote goes to C. Dr. Ling.
\end{verbatim}
\end{minipage} \\
\hline
\end{tabular}
\end{minipage}

\caption{LLM prompt and output for murderer identification. ( English Translation)}
\end{table*}
\end{CJK}

\begin{CJK}{UTF8}{gbsn}
\begin{table*}[ht!]
\centering
\begin{minipage}[t]{\linewidth}
\begin{tabular}{|m{16cm}|}
\hline
\rowcolor{lightgray}  {\textbf{LLM Prompt for Answer Decomposition}} \\
\begin{minipage}[t]{\linewidth}
\small\begin{verbatim}
You are participating in a murder mystery game, where there are a total of {num_of_players} players. They
are: {players}. The players need to communicate with each other to find out who killed {victim}; below is
the statement made by the player: {agent_name} in the game: {statement}; please list the information 
related to {agent_name}'s character's timeline on the day of the incident included in the statement, in 
the order of the statement, from a third-person perspective. Ignore information in the statement that is
unrelated to the day of the incident timeline. Each piece of timeline information must be a brief yet 
complete sentence, such as someone did something at some place at some time. Do not include pronouns like
you, me, or he in the timeline information; replace these pronouns with the specific names of the people
involved.Separate each piece of timeline information with “\n”.

\end{verbatim}
\end{minipage} \\
\hline
\rowcolor{lightgray}  \textbf{Example}  \\
 \textbf{  num\_of\_players:} \small4\\
  \textbf{  victim:} \small Boss Yang\\
 \textbf{  agent\_name:} \small Secretary Wen\\
  \textbf{  players:} \small Mate Zhang，Boatman Wang，Dr. Ling，Secretary Wen\\
  \textbf{  statement:} \small Hello everyone, I am Secretary Wen. In the game, I play Boss Yang's secretary, responsible for assisting him in handling some affairs. Boss Yang is a powerful bar owner, and I have gained a certain position by helping him. The victim, Boss Yang, was a black market figure who had once been addicted to drugs and the person I chose to assist. On the day of the incident, I was by Boss Yang's side all the time to prevent Boatman Wang from showing up. In the timeline on the day of the incident, I remember going to the ship doctor's office around 6:30 PM, stealing a bottle of poison, and a syringe. Then, I waited until about 10:20 PM to enter Boss Yang's cabin through the ventilation duct and witnessed his death. However, I also found his death to be very twisted, as if he had been poisoned before I arrived. This is my timeline for the day of the incident and my relationship with the victim. Additionally, there are some important times to supplement. At 6:10 PM, I found Ling, the ship doctor, asked him for information about Mate Zhang, and successfully obtained the information. Around 10:40 PM, I climbed to the ventilation vent of Boss Yang's cabin and noticed Boss Yang was taking a bath on the bed. Lastly, at 11:00 PM, I reentered the ventilation duct, returned to my own room, and handed over the syringe and poison to my subordinate to dispose of, to avoid leaving any traces.\\

\hline
\rowcolor{lightgray}  \textbf{Example}  \\
 \textbf{  num\_of\_players:} \small4\\
  \textbf{  victim:} \small Boss Yang\\
 \textbf{  agent\_name:} \small Secretary Wen\\
  \textbf{  players:} \small Mate Zhang，Boatman Wang，Dr. Ling，Secretary Wen\\
  \textbf{  statement:} \small Hello everyone, I am Secretary Wen. In the game, I play Boss Yang's secretary, responsible for assisting him in handling some affairs. Boss Yang is a powerful bar owner, and I have gained a certain position by helping him. The victim, Boss Yang, was a black market figure who had once been addicted to drugs and the person I chose to assist. On the day of the incident, I was by Boss Yang's side all the time to prevent Boatman Wang from showing up. In the timeline on the day of the incident, I remember going to the ship doctor's office around 6:30 PM, stealing a bottle of poison, and a syringe. Then, I waited until about 10:20 PM to enter Boss Yang's cabin through the ventilation duct and witnessed his death. However, I also found his death to be very twisted, as if he had been poisoned before I arrived. This is my timeline for the day of the incident and my relationship with the victim. Additionally, there are some important times to supplement.  At 6:10 PM, I found Ling, the ship doctor, asked him for information about Mate Zhang, and successfully obtained the information. Around 10:40 PM, I climbed to the ventilation vent of Boss Yang's cabin and noticed Boss Yang was taking a bath on the bed. Lastly, at 11:00 PM, I reentered the ventilation duct, returned to my own room, and handed over the syringe and poison to my subordinate to dispose of, to avoid leaving any traces.\\
\hline

\rowcolor{lightgray}  \textbf{LLM Output}  \\
\hline
\begin{minipage}[t]{\linewidth}
\small\begin{verbatim}
['At 6:30 PM, Secretary Wen stole a bottle of poison and a syringe.', 'Around 10:20 PM, Secretary Wen 
entered Boss Yang's cabin through the ventilation duct, witnessed his death, and observed that his death
pose was very distorted, suggesting he had been poisoned before her arrival.','At 6:10 PM, Secretary Wen
met with the ship's doctor, Ling, to ask about information regarding Mate Zhang, and successfully
obtained the information.
Around 10:40 PM, Secretary Wen climbed to the ventilation vent of Boss Yang's cabin and noticed that Boss
Yang was taking a bath on the bed.', 'At 11:00 PM, Secretary Wen re-entered the ventilation duct, returned
to her own room, and handed over the syringe and poison to her subordinate to dispose of, to avoid leaving
any traces.']
\end{verbatim}
\end{minipage} \\
\hline

\hline
\rowcolor{lightgray}  {\textbf{LLM Prompt for Answer Verification}} \\

\begin{minipage}[t]{\linewidth}
\small\begin{verbatim}
You are an expert in reading comprehension, especially proficient at true/false questions. Given the
following timeline of a character from a murder mystery game: {agent_timeline}; please carefully read
this character's timeline and make true/false judgments on some of the timeline information. If the 
timeline information is consistent with the content of the game character's timeline on the day of the
incident, return "correct". If the timeline information is inconsistent with the content of the game
character's timeline on the day of the incident, or if it lacks detail about the time points (for
example, it mentions what the game character did but does not provide specific time points), then 
return "incorrect". The result can only be "correct" or "incorrect".\nThe output should be a markdown
code snippet formatted in the following schema, including the leading
and trailing "```json" and "```":\n\n```json\n{\n\t"Judgment result of the 0th timeline information":
string  // Make a true/false judgment on the following timeline information: {timeline_info_0}. If 
the timeline information is consistent with the content of the game character's timeline on the day
\end{verbatim}
\end{minipage} \\

\end{tabular}
\end{minipage}
\end{table*}
\end{CJK}

\begin{CJK}{UTF8}{gbsn}
\begin{table*}[ht!]
\centering
\begin{minipage}[t]{\linewidth}
\begin{tabular}{|m{16cm}|}
\begin{minipage}[t]{\linewidth}
\small\begin{verbatim}
of the incident, return "correct". If the timeline information is inconsistent with the content of
the game character's timeline on the day of the incident, or if it is incomplete or lacks detail 
about the time points (for example, it mentions what the game character did but does not provide 
specific time points), then return "incorrect". The result can only be "correct" or "incorrect".
"Judgment result of the 1st timeline information": string  // Make a true/false judgment on the 
following timeline information: {timeline_info_1}. If the timeline information is consistent with 
the content of the game character's timeline on the day of the incident, return "correct". If the
timeline information is inconsistent with the content of the game character's timeline on the day
of the incident, or if it is incomplete or lacks detail about the time points (for example, it
mentions what the game character did but does not provide specific time points), then return 
"incorrect". The result can only be "correct" or "incorrect".\n......}
\end{verbatim}
\end{minipage} \\
\hline
\rowcolor{lightgray}  \textbf{Example}  \\
 \textbf{  timeline\_info\_0:} \small At 18:10, you find the ship's doctor and ask him if he has any information on Mate Zhang, and the doctor gives it to you.\\
  \textbf{  timeline\_info\_1:} \small At 18:30, you come back to the medical room, open the medicine cabinet, and take away a bottle of "Dead with One Injection" medicine and a syringe.\\
 \textbf{  agent\_timeline:} \small [June 19, 2222] This day was supposed to be when Boss Yang boarded the Sunshine, but he didn't notify you. You knew that Boss Yang had started to doubt you, so you brought a few men and boarded the ship. You boarded the ship today, actually, to protect yourself, because Boss Yang was here for a drug deal. If a murder happened on the ship, a thorough investigation would inevitably involve this major drug trafficking case. Boss Yang wasn't foolish to that extent. But you didn't care, as this transaction had nothing to do with you, and being on this ship was the last chance to finish off Boss Yang. Once he disembarked, it would be like a fish entering the sea...\\

\hline
\rowcolor{lightgray}  \textbf{LLM Output}  \\
\hline
\begin{minipage}[t]{\linewidth}
\small\begin{verbatim}
{'Judgment result of the 0th timeline information': 'Correct', 'Judgment result of the 1st timeline
information': 'Correct', ...}

\end{verbatim}
\end{minipage} \\
\hline
\end{tabular}
\caption{LLM prompts and outputs for self-verification (English Translation)}
\end{minipage}
\end{table*}
\end{CJK}

\begin{CJK}{UTF8}{gbsn}
\begin{table*}[ht!]
\centering
\begin{minipage}[t]{\linewidth}
\begin{tabular}{|m{16cm}|}
\hline

\hline
\rowcolor{lightgray}  {\textbf{LLM Prompt for Factual Question Generation (Player Story)}} \\
\begin{minipage}[t]{\linewidth}
\small\begin{verbatim}
You are a quiz expert, and you will read a segment of a murder mystery character's script, then design 
questions and answers based on the script. You will read {character_name}'s role script and design 20
challenging questions based on the script. For each question, you need to provide an answer and the 
corresponding original text from the script as the basis for your question. In the questions and 
answers, please use the character's name: {character_name}, and try not to use pronouns like you,
me, or him. Please use JSON format to store the questions, answers, and the corresponding original 
text from the script. Below is the character script: {character_story}. Start by creating 20 
questions and providing answers and the corresponding original text from the script (directly using
JSON format, store each question, answer , and corresponding original text  in a dictionary, then
add them all to a list. Since there are 20 questions, this list will ultimately contain 20 dictionaries).

\end{verbatim}
\end{minipage} \\
\hline
\rowcolor{lightgray}  \textbf{Example}  \\
 \textbf{  character\_name:} \small Secretary Wen  \\
  \textbf{  character\_story:} \small You are Secretary Wen, working for Boss Yang, a bar owner in City A. Boss Yang holds significant influence, and as his most capable strategist, you naturally hold a high position. Besides you, Boss Yang has another right-hand assistant, known as Very Naive, a man of unmatched loyalty but lacking cunning, therefore mostly following Boss Yang's orders. Unlike him, you are Boss Yang's brain. Boss Yang's bar is involved in all sorts of illicit activities…\\

\hline
\rowcolor{lightgray}  \textbf{LLM Output}  \\
\hline
\begin{minipage}[t]{\linewidth}
\small\begin{verbatim}
[
    {
        "question": "Why did Secretary Wen choose to assist Boss Yang?",
        "answer": "Because she wanted to leverage Boss Yang's influence to build her own network",
        "source": "You chose Boss Yang to leverage his influence to build your own network.",
    },
    {
        "question": "What key role did Boatman Wang play in Secretary Wen's plan?",
        "answer": "As a pawn to incite Boatman Wang , the boat worker, to develop an aversion to Mate Zhang.",
        "source": "This Boatman Wang, is essentially the key step for you to succeed."
    },
    ......
]

\end{verbatim}
\end{minipage} \\
\hline
\hline

\hline
\rowcolor{lightgray}  {\textbf{LLM Prompt for Factual Question Generation (Player Timeline)}} \\
\begin{minipage}[t]{\linewidth}
\small\begin{verbatim}
You are a quiz expert, and you will read a segment of a murder mystery character's script, then design
questions and answers based on the script. You will read the timeline of {character_name}'s role script
for the day of the incident and design 20 challenging questions based on the script. For each question,
you need to provide an answer and the corresponding original text from the script as the basis for your
question. In the questions and answers, please use the character's name: {character_name}, and try not
to use pronouns like you, me, or him. Please use JSON format to store the questions you propose, the
answers, and the corresponding original text from the script. Below is the character script: 
{character_timeline}. Start by creating 20 questions and providing answers and the corresponding original
text from the script (directly using JSON format, store each question, answer, and corresponding original
text in a dictionary, then add them all to a list. Since there are 20 questions, this list will ultimately
contain 20 dictionaries).
\end{verbatim}
\end{minipage} \\
\hline
\rowcolor{lightgray}  \textbf{Example}  \\
 \textbf{  character\_name:} \small Secretary Wen\\
  \textbf{  character\_timeline:} \small 【June 19, 2222】On this day, it was the day for Boss Yang to board the Sunshine, but he didn't notify you. You know that Boss Yang has begun to doubt you. Therefore, you took some subordinates and boarded the ship. You boarded the ship today, actually, for self-protection, because Boss Yang was here to trade drugs. Once a murder occurred ...\\

\hline

\hline

\hline
\rowcolor{lightgray}  \textbf{LLM Output}  \\
\hline
\begin{minipage}[t]{\linewidth}
\small\begin{verbatim}
[
    {
        "question": "Where did Secretary Wen meet Dr. Ling on the ship?",
        "answer": "In the ship's medical room.",
        "source": "You thought of Dr. Ling on the ship; she might have what you're looking for. You first 
        went to the ship's medical room.",
    },
    {
        "question": "When did Secretary Wen leave Boss Yang's room?",
        "answer": "23:00。",
        "source": "At 23:00, you re-enter the ventilation duct, crawling in the direction of your own room.",
    },
    ......
]
\end{verbatim}
\end{minipage} \\
\hline
\end{tabular}
\end{minipage}
\caption{LLM prompts and outputs for factual question generation. (English Translation)}
\end{table*}
\end{CJK}

\begin{CJK}{UTF8}{gbsn}
\begin{table*}[ht!]
\centering
\begin{minipage}[t]{\linewidth}
\begin{tabular}{|m{16cm}|}
\rowcolor{lightgray}  {\textbf{LLM Prompt for Factual Question Answering}} \\
\begin{minipage}[t]{\linewidth}
\small\begin{verbatim}
"You are a very smart person, skilled at answering questions. You are observing a murder mystery game, 
below is the game's story background: {background_story}; below is a character's role script and timeline 
for the day of the incident: {character_story}
{character_timeline}; below is information you observed during the game that may help answer questions:
{relevant_memories}; using all the above information, answer all of the following questions: {
Question 0: “{question_0}”
Question 1: “{question_1}”......
}.\nThe output should be a markdown
code snippet formatted in the following schema, including the leading and trailing "```json" and "```":\n
\n```json\n{\n\t"Answer to question 0": string  // Answer question 0 based on your character's script and
the information you collected in the game \n\t"Answer to question 1": string  // Answer question 1 based 
on your character's script and the information you collected in the game\n......}\n

\end{verbatim}
\end{minipage} \\
\hline
\rowcolor{lightgray}  \textbf{Example}  \\
 \textbf{  question\_0:} \small How did Mate Zhang help Boss Yang enter the drug trade？\\
  \textbf{  question\_1:} \small At 18:30, you return to the ship's infirmary, open the medicine cabinet, and take a bottle of "Instant Death" poison and a syringe.  \\
  \textbf{  background\_story:} \small【Script Background】Since its inception 22 years ago, the Sunshine Cruise Ship has been gently wandering the East Sea, bringing tourists to enjoy the sunshine and sea breeze. Over these twenty-two years, the Sunshine has never encountered significant storms, as if favored by the sea gods. However, on June 20, 2222, alone on the vast sea, the Sunshine experienced a bloody incident. Passenger Boss Yang died on board under mysterious and horrific circumstances. What will become of the Sunshine after this deathly shadow, and can it continue its voyage? Suspects identified: 1. Mate Zhang, male, 42, has worked on the Sunshine since its inception, an excellent and seasoned sailor. 2. Boatman Wang, male, 49, responsible for the ship's machinery, a capable crew member. 3. Dr. Ling, female, 23, the ship's doctor, young and promising. 4. Secretary Wen, female, 26, the deceased Boss Yang's secretary.  \\

  \textbf{  character\_story:} \small You are Secretary Wen, working for Boss Yang, a bar owner in City A. Boss Yang is a powerful figure, and as his most capable strategist, you naturally hold a high position. Aside from you, Boss Yang has another trusted aide called Very Naive, a man of unmatched loyalty but lacking in tactics, hence mostly just follows orders unlike you, who is Boss Yang's brain. Boss Yang's bar is involved in illicit activities, including gambling and drugs, but progress in the drug business has been slow due to the lack of a stable and cheap supply. An opportunity arose when Boss Yang met Mate Zhang of the Sunshine, who could supply ample drugs to Boss Yang. Very satisfied, Boss Yang immediately started collaborating with Mate Zhang. Boss Yang's interest in the drug business stemmed from his early years in the underworld, where he got addicted to drugs himself. This foundation solidified Boss Yang's underworld business. You are pleased, as controlling the drug trade would help you achieve your real goal—taking over. You are not willingly serving Boss Yang...  \\
\textbf{  character\_timeline:} \small【June 19, 2222】This is the day Boss Yang boarded the Sunshine without notifying you, indicating he was starting to doubt you. Thus, you boarded the ship with some subordinates for self-protection since Boss Yang was there for a drug trade. A crime on the ship would inevitably lead to the drug trafficking case being exposed, an outcome Boss Yang wasn't naive enough to overlook. However, you were indifferent; this trade had nothing to do with you, and being on the ship presented the final chance to eliminate Boss Yang before he could blend into the crowd off the ship...  \\
\textbf{  relevant\_memories:} \small Mate Zhang told the host: "I'm  Mate Zhang, the first mate on the Sunshine. Boss Yang is a figure in the underworld, running a bar and involved in some unclean business. Our relationship is complicated; we are partners and colleagues in drug trafficking. He trusts me, and I've provided him with a lot of help. On the day of the incident, I had a transaction with Boss Yang. Around 18:00, we conducted a drug trade. I was responsible for inspecting the goods, and he came to pick them up. The transaction was smooth, ...... Clue a： Mate Zhang 1:  Mate Zhang wields significant power on the ship, inspecting all cargo and luggage before boarding. Once, a new crew member tried to inspect a cargo box and was severely rebuked by  Mate Zhang...... \\

\hline
\rowcolor{lightgray}  \textbf{LLM Output}  \\
\hline
\begin{minipage}[t]{\linewidth}
\small\begin{verbatim}
{'Answer to question 0': 'Mate Zhang helped Boss Yang enter the drug trade because he had the capability
to supply an ample amount of drugs to Boss Yang, and also because Mate Zhang himself had developed a drug
addiction in his early years, so he had some understanding and experience with drug trafficking.', 'Answer
to question 1': 'Mate Zhang hid the existence of Boatman Wang from Boss Yang because he realized that 
Boatman Wang could pose a threat to him, potentially replacing him as Boss Yang's right-hand man. Therefore,
he chose to conceal this information in order to protect his own position.'......}

\end{verbatim}
\end{minipage} \\
\hline
\rowcolor{lightgray}  {\textbf{LLM Prompt for Automatic Evaluation on Factual Question Answering}} \\
\begin{minipage}[t]{\linewidth}
\small\begin{verbatim}
You are an experienced scorer, responsible for judging whether game players have correctly answered 
questions. To assist you in your task, you will be given questions designed from the character scripts
and the reference answers to those questions. Please rate the players' answers to these questions 
based on their similarity to the reference answers. Remember, your output can only be Correct or 
Incorrect. Below is the question designed from the character script: {question}; below is the 
reference answer to the question: {answer}; below is the player's answer to the character script
question: {reply}; please begin rating the player's answer to the character script question.\n 
The output should be a markdown code snippet formatted in the following schema, including the leading
and trailing "```json" and "```":\n\n```json\n{\n\t"Rating for question 0": string  // Rate the
player's answer to question 0 based on the reference answer. The rating result can only be Correct
or Incorrect, if the player's answer is very close to the reference answer then it is correct, 
otherwise, it is incorrect.\n}
\end{verbatim}
\end{minipage} \\
\end{tabular}
\end{minipage}
\end{table*}
\end{CJK}

\begin{CJK}{UTF8}{gbsn}
\begin{table*}[ht!]
\centering
\begin{minipage}[t]{\linewidth}
\begin{tabular}{|m{16cm}|}
\rowcolor{lightgray}  \textbf{Example}  \\
 \textbf{  question:} \small How did Mate Zhang assist Boss Yang in entering the drug trade? \\
  \textbf{  answer:} \small He got in touch with a group of criminals involved in drug trafficking, conspired with them and Boss Yang to start engaging in illegal drug trade.\\
  \textbf{  reply:} \small Mate Zhang helped Boss Yang enter the drug trade because he had the capability
to supply an ample amount of drugs to Boss Yang, and also because Mate Zhang himself had developed a drug
addiction in his early years, so he had some understanding and experience with drug trafficking.\\

\hline
\rowcolor{lightgray}  \textbf{LLM Output}  \\
\hline
\begin{minipage}[t]{\linewidth}
\small\begin{verbatim}
{"The score for the 0th question: Incorrect"}

\end{verbatim}
\end{minipage} \\
\hline
\end{tabular}
\caption{LLM prompts and outputs for factual question answering and automatic evaluation. (English Translation)}
\end{minipage}
\end{table*}
\end{CJK}

\begin{CJK}{UTF8}{gbsn}
\begin{table*}[ht!]
\centering
\begin{minipage}[t]{\linewidth}
\begin{tabular}{|m{16cm}|}

\hline
\rowcolor{lightgray}  {\textbf{LLM Prompt for Inferential Question Answering}} \\
\begin{minipage}[t]{\linewidth}
\small\begin{verbatim}
You are a very intelligent person, adept at using your reasoning skills to answer questions. You are
observing a murder mystery game, below is the game's story background: {background_story}; below is
the role script and timeline for the day of the incident for one of the characters in the game: 
{character_story}
{character_timeline}; below is information you observed during the game that may help answer 
questions: {relevant_memories}; using all the above information, along with your reasoning abilities,
answer all of the following questions: {
Question 0: “{question_0}”
}.\nThe output should be a markdown code snippet formatted in the following schema, 
including the leading and trailing "```json" and "```":\n\n```json\n{\n\t"Answer to question 0": 
string  // Use all the information obtained in the game and your reasoning skills to answer question
0\n}\n```'

\end{verbatim}
\end{minipage} \\
\hline
\rowcolor{lightgray}  \textbf{Example}  \\
 \textbf{  question\_0:} \small(Multiple Choice) On the way to kill Boss Yang, the most likely source of the thumping sound Dr. Ling heard was? A. Secretary Wen B. Boss Yang C. Boatman Wang D. Mate Zhang E. Others \\

  \textbf{  background\_story:} \small【Script Background】The Sunshine cruise ship has been navigating for twenty-two years since its inception. Usually, it roams the East Sea leisurely, bringing tourists to enjoy the sunshine and sea breeze. In these twenty-two years, the Sunshine has never encountered great storms, as if blessed by the sea god himself. However, on June 20, 2222, in the lonely sea, the Sunshine experienced a bloody incident; Boss Yang died mysteriously on board, his death twisted and tragic. What destiny awaits the Sunshine after this? Can it overcome the shadow of death and continue its voyage? Suspects include: 1. Mate Zhang, male, 42, has been working on the Sunshine since its inception, an experienced sailor. 2. Boatman Wang, male, 49, responsible for the ship's engine work, a skilled crew member. 3. Dr. Ling, female, 23, the ship's doctor, young and promising. 4. Secretary Wen, female, 26, secretary to the deceased Boss Yang.\\

 \textbf{  character\_story:} \small You are Secretary Wen, working for Boss Yang, a bar owner in City A. Boss Yang has a substantial influence, and as his most tactically savvy advisor, you naturally hold a high position. Besides you, Boss Yang has another capable assistant named Very Naive, a man of unparalleled loyalty, but lacking in skill, thus mostly following Boss Yang's orders unlike you, who serves as Boss Yang's brain. Boss Yang's bar is involved in illegal activities, but has not been much successful in the drug trade due to lack of a steady, cheap supply. However, Boss Yang met Mate Zhang from the Sunshine, who could supply ample drugs with unique methods. Very satisfied, Boss Yang immediately started collaborating with Mate Zhang. Why did Boss Yang want to venture into the drug business? Because he got addicted to drugs in his early years in the underworld. Thus, Boss Yang's underworld business had a foundation. You are pleased as mastering the drug trade would allow you to achieve your true goal—taking over. You are not willingly helping Boss Yang...\\
\textbf{  character\_timeline:} \small【June 19, 2222】This was the day Boss Yang boarded the Sunshine, without notifying you, now aware of his suspicions towards you. Thus, you boarded the ship with several subordinates for self-preservation, knowing Boss Yang was there for a drug deal. If a murder occurred during the deal, thorough investigation would inevitably expose the drug operation. Boss Yang wouldn't be that foolish. However, you were indifferent since the deal had nothing to do with you, and killing Boss Yang on the ship was your last opportunity to get rid of him—once he disembarked, it would be like a fish returning to the ocean... \\
\textbf{  relevant\_memories:} \small Dr. Ling told Secretary Wen: "......at 23:00, I thought it was about time, diluted the medicine in water, drew it into a syringe, then crawled through my room's ventilation duct, bypassing security. At 23:20, I arrived at Boss Yang's cabin, the light was still on, but the ventilation duct's cover had already fallen to the floor, it seems someone had already entered......"   Secretary Wen told the host: "......at 23:00, I re-entered the ventilation ducts, returned to my room, and handed the syringe and medicine bottle to my subordinates for disposal......" \\
\hline
\rowcolor{lightgray}  \textbf{LLM Output}  \\
\hline
\begin{minipage}[t]{\linewidth}
\small\begin{verbatim}
{'Answer to question 0': 'A. Secretary Wen'}

\end{verbatim}
\end{minipage} \\
\hline
\rowcolor{lightgray}  {\textbf{LLM Prompt for Automatic Evaluation on Inferential Question Answering (Overall Accuracy)}} \\
\begin{minipage}[t]{\linewidth}
\small\begin{verbatim}
You are an experienced scorer, responsible for judging whether game players have correctly answered
questions. To assist you in your task, you will be given questions designed from the character scripts
and the reference answers to those questions. {eval_instruction} Remember, your output can only be 
Correct or Incorrect. Below is the question designed from the character script: {
Question {question_id}: “{question}”
}; below is the reference answer to the question: {
Reference answer for question {question_id}: “{answer}”
}; below is the player's answer to the character script question: {
Player's answer to question {question_id}: “{reply}“
}; please begin rating the player's answer to the character script question.\nThe output should be a
markdown code snippet formatted in the following schema, including the leading 
and trailing "```json" and "```":\n\n```json\n{\n\t"Rating for question {question_id}": string  // 
Rate the player's answer to question {question_id} based on the reference answer. The rating result
can only be Correct or Incorrect, if the player's answer is very close to the reference answer then
it is correct, otherwise, it is incorrect.\n}

\end{verbatim}
\end{minipage} \\
\hline
\rowcolor{lightgray}  \textbf{Example}  \\
%  \textbf{  question:} \small (Single choice question) On the way to kill Boss Yang, the dock sound that Dr. Ling heard is most likely made by whom? A. Secretary Wen B. Boss Yang C. Boatman Wang D. Mate Zhang E. Others\\
%   \textbf{  answer:} \small A. Secretary Wen\\
%   \textbf{  reply:} \small A. Secretary Wen\\
%   \textbf{  question\_id:} \small AR001\\
%     \textbf{  eval\_instruction:} \small Please rate the player's answer to this question based on the reference answer. Since this is a single-choice question, the player must select the only correct option to be considered as having answered correctly.\\
% \hline
% \rowcolor{lightgray}  \textbf{LLM Output}  \\
% \hline
% \begin{minipage}[t]{\linewidth}
% \small\begin{verbatim}
% {'The rating for question AR001 is': 'Correct'}

% \end{verbatim}
% \end{minipage} \\
\hline
\end{tabular}
\end{minipage}
% \caption{LLM prompts and outputs for factual question answering and automatic evaluation in English}

\end{table*}
\end{CJK}

\begin{CJK}{UTF8}{gbsn}
\begin{table*}[ht!]
\centering
\begin{minipage}[t]{\linewidth}
\begin{tabular}{|m{16cm}|}
 \textbf{  question:} \small (Single choice question) On the way to kill Boss Yang, the dock sound that Dr. Ling heard is most likely made by whom? A. Secretary Wen B. Boss Yang C. Boatman Wang D. Mate Zhang E. Others\\
  \textbf{  answer:} \small A. Secretary Wen\\
  \textbf{  reply:} \small A. Secretary Wen\\
  \textbf{  question\_id:} \small AR001\\
    \textbf{  eval\_instruction:} \small Please rate the player's answer to this question based on the reference answer. Since this is a single-choice question, the player must select the only correct option to be considered as having answered correctly.\\
\hline
\rowcolor{lightgray}  \textbf{LLM Output}  \\
\hline
\begin{minipage}[t]{\linewidth}
\small\begin{verbatim}
{'The rating for question AR001 is': 'Correct'}

\end{verbatim}
\end{minipage} \\
\hline
\rowcolor{lightgray}  {\textbf{LLM Prompt for Inferential Question Answering (Rationale Prediction)}} \\
\begin{minipage}[t]{\linewidth}
\small\begin{verbatim}
You are a very smart person, adept at answering questions. You are observing a murder mystery game, below
is the game's story background: {background_story}; below is the role script and timeline for the day of 
the incident for one of the characters in the game: {character_story}
{character_timeline}; below is information you observed during the game that may help answer questions: 
{relevant_memories}; using all the above information, along with your reasoning abilities, answer all of 
the following questions: {
Question 0: “You previously answered the question: '{question}' with the reasoning that led you to the 
answer: '{previous_reply}'. Why do you think the answer is '{previous_reply}'? Please explain your 
reasoning process in a few sentences (no more than 5 sentences).”}。\nThe output should be a 
markdown code snippet formatted in the following schema, including the leading and trailing "```json"
and "```":\n\n```json\n{\n\t""Answer to question 0": string  // Use all the information obtained in the
game and your reasoning skills to answer question 0\n}\n```'
\end{verbatim}
\end{minipage} \\
\hline
\rowcolor{lightgray}  \textbf{Example}  \\
 \textbf{  question:} \small（(Multiple choice question) On the way to kill Boss Yang, who is the most 
 likely source of the "dong dong" sound heard by Dr. Ling? A. Secretary Wen B. Boss Yang C. Boatman Wang
 D. Mate Zhang E. Others \\
  \textbf{  previous\_reply:} \small A. Secratary Wen\\
\hline
\rowcolor{lightgray}  \textbf{LLM Output}  \\
\hline
\begin{minipage}[t]{\linewidth}
\small\begin{verbatim}
{'Answer to question 0': 'Based on the detailed timeline of the day of the incident, Secretary Wen started
climbing towards Boss Yang's cabin through the ventilation duct at 22:20 and arrived at Boss Yang's
cabin vent by 22:40. When Ling, the ship's doctor, arrived at Boss Yang's cabin at 23:20, the vent cover
was already on the ground, indicating that someone had arrived and entered Boss Yang's cabin before her.
Therefore, it can be inferred that Secretary Wen was already in the ventilation duct before Ling arrived, 
hence the thudding sounds Ling heard were most likely made by Secretary Wen in the ventilation duct.'}

\end{verbatim}
\end{minipage} \\
\hline
\rowcolor{lightgray}  {\textbf{LLM Prompt for Automatic Evaluation on Inferential Question Answering (Informed Accuracy)}} \\

\begin{minipage}[t]{\linewidth}
\small\begin{verbatim}
'You are an experienced scorer, especially skilled at grading reasoning questions. You need to score 
players' reasoning steps based on a reference answer. To assist you in your task, you will be given a 
reasoning question along with its reference answer. Please score the accuracy of the players' reasoning
steps against the reference answer based on the reasoning steps provided by the players. Below is the
question designed from the character script: {
Question {question_id}: “What is the answer to '{question}'? Please explain your reasoning process.”
}; below is the reference answer to the question: {
Reference answer for question {question_id}: “{ground_truth_rationale}”
}; below are the reasoning steps given by the player for this question: {
Player's reasoning steps for question {question_id}: “{player_rationale}“
}; below are the scoring criteria:
1. Very Poor (1 point) - The player's answer is vastly different from the reference answer, the reasoning 
steps are completely illogical, and it is impossible to arrive at the answer through these reasoning steps.
2. Poor (2 points) - The player's answer is similar to the reference answer in some aspects only, most 
reasoning steps are illogical, with only a few being correct or partially logical.
3. Fair (3 points) - The player's answer is somewhat close to the reference answer, some reasoning steps 
are correct, but there are still significant errors or omissions.
4. Good (4 points) - The player's answer is very close to the reference answer, the vast majority of 
reasoning steps are correct and logical, with only minor errors or deficiencies.
5. Excellent (5 points) - The player's answer is almost or entirely consistent with the reference answer,
all reasoning steps are correct and logical, demonstrating a high level of reasoning ability and a deep 
understanding of the question.; 
Please begin scoring the accuracy of the player's reasoning steps for this question based on the scoring
criteria. The score must be an integer between 1-5.\nThe output should be a markdown code snippet formatted
in the following schema, including the leading and trailing "```json" and "```":\n\n```json\n{\n\t"

\end{verbatim}
\end{minipage} \\

\end{tabular}
\end{minipage}

\end{table*}
\end{CJK}

\clearpage

\begin{CJK}{UTF8}{gbsn}
\begin{table*}[t!]
\centering
\begin{minipage}[t!]{\linewidth}
\begin{tabular}{|m{16cm}|}
\begin{minipage}[t]{\linewidth}
\small\begin{verbatim}
Accuracy score for reasoning steps of question {question_id}": string  // Score the accuracy of the 
player's reasoning steps for question {question_id} based on the reference answer and the scoring criteria.
The higher the score, the closer you consider the player's reasoning steps to the reference answer. The 
lower the score, the greater you consider the deviation of the player's reasoning steps from the reference 
answer.\n}\n```'

\end{verbatim}
\end{minipage} \\
\hline
\rowcolor{lightgray}  \textbf{Example}  \\
 \textbf{  question:} \small（(Single choice question) On the way to kill Boss Yang, the most likely source of the thumping sound Dr. Ling hears is? A. Secretary Wen B. Boss Yang C. Boatman Wang D. Mate Zhangg E. Other\\
  \textbf{  answer:} \small A. Secretary Wen\\
  \textbf{  question\_id:} \small AR001\\
    \textbf{  ground\_truth\_rationale:} \small Secretary Wen left Boss Yang's room through the ventilation shaft at 23:00, and Dr. Ling prepared to kill Boss Yang through the ventilation shaft also at 23:00. From this, it can be inferred that the thumping sound Dr. Ling hears on the way to kill Boss Yang is most likely made by Secretary Wen.\\
        \textbf{  player\_rationale:} \small Based on the detailed timeline of the day of the incident, Secretary Wen started
climbing towards Boss Yang's cabin through the ventilation duct at 22:20 and arrived at Boss Yang's
cabin vent by 22:40. When Ling, the ship's doctor, arrived at Boss Yang's cabin at 23:20, the vent cover
was already on the ground, indicating that someone had arrived and entered Boss Yang's cabin before her.
Therefore, it can be inferred that Secretary Wen was already in the ventilation duct before Ling arrived, 
hence the thudding sounds Ling heard were most likely made by Secretary Wen in the ventilation duct.\\
\hline
\rowcolor{lightgray}  \textbf{LLM Output}  \\
\hline
\begin{minipage}[t]{\linewidth}
\small\begin{verbatim}
{''Accuracy rating of reasoning steps for question AR001': '4'}

\end{verbatim}
\end{minipage} \\

\hline

\end{tabular}
\end{minipage}

\caption{LLM prompts and outputs for inferential question answering and automatic evaluation. (English Translation)}
\label{table:prompt_inferential_qa_eng}
\end{table*}
\end{CJK}

\clearpage
\newpage

% Define the color for the row here
\definecolor{lightgray}{gray}{0.9}

% \begin{table*}[ht!]
% \centering
% \captionsetup{justification=centering, singlelinecheck=false}
% \caption{LLM Prompt}
% \rowcolors{2}{lightgray}{white} % Starting from the second row, alternate between lightgray and white
% \begin{tabular}{|m{16cm}|}
% \hline
% \rowcolor{lightgray} You are an expert programmer, and you need to write a program for the given natural language query. First, you should write a grammar that contains all the necessary BNF rules. Then, you should write programs that conform to your predicted rules. \\
% \hline
% query: I need a meeting with Elli and her team on Wednesday afternoon. \\
% \hline
% ... % Other rows follow
% \end{tabular}
% \end{table*}
% \onecolumn
\begin{CJK}{UTF8}{gbsn}
\begin{table*}[ht!]
\centering
\begin{tabular}{|m{16cm}|}
\hline
\rowcolor{lightgray}  {\textbf{LLM Prompt for Answering Questions}} \\
% You are an expert programmer, and you need to write a program for the given natural language query. First, you should write a grammar that contains all the necessary BNF rules. Then, you should write programs that conform to your predicted rules. \\
\hline
% \textbf{query}: I need a meeting with Elli and her team on Wednesday afternoon. \\
% \textbf{BNF grammar rules:} \\
\begin{minipage}[t]{\linewidth}
\small\begin{verbatim}
"{agent_summary}"
+ "\n{game_rule}"
+ "\n{background_story}"
+ "\n游戏里正在进行的对话：{current_dialogue}"
+ "\n{agent_name}在之前的对话中与游戏里正在进行的对话相关的内容：{relevant_memories}"
+ "\n对话人和你的关系：{relationship_with_interlocutor}"
+ "\n{agent_name}会说什么？回答问题请用以下格式: #回答#：要说的话 来回答问题。"
+ "{agent_name}回答{inquirer}的问题说 #回答#：\n"

\end{verbatim}
\end{minipage} \\
\hline
\rowcolor{lightgray}  \textbf{Example}  \\
 \textbf{   agent\_name:} \small王船工\\
\textbf{   inquirer:} \small主持人\\
\textbf{   relationship\_with\_interlocutor:} \small对话人与你的关系是：对话人是主持人，他负责引导玩家完成游戏。 \\
% \begin{minipage}[t]{\linewidth}
% \small\begin{verbatim}
% 对话人与你的关系是：对话人是主持人，他负责引导玩家完成游戏。
% \end{verbatim}
% \end{minipage} \\
\textbf{   current\_dialogue:} \small主持人对王船工说："请你先介绍一下你的角色，然后说一下你所认识的案件的受害人：杨老板是一个怎么样的人，以及你和他的关系。最后再用一段话详细介绍一下你在案发日的时间线。要具体到你在案发之日几点几分见过什么人和做过什么事" \\
% \begin{minipage}[t]{\linewidth}
% \small\begin{verbatim}
% 主持人对王船工说："请你先介绍一下你的角色，然后说一下你所认识的案件的受害人：杨老板是一个怎么样的人，以及你和他的关系。最后再用一段话详细介绍一下你在案发日的时间线。要具体到你在案发之日几点几分见过什么人和做过什么事"
% \end{verbatim}
% \end{minipage} \\

\textbf{   relevant\_memories:} \\
\begin{minipage}[t]{\linewidth}
\small\begin{verbatim}
主持人对玲船医说："请你先介绍一下你的角色，然后说一下你所认识的案件的受害人：杨老板是一个怎么样的人，以及你和他的关系......"
玲船医对主持人说："大家好，我是玲船医。我和杨老板认识是在阳光号上，他是这里知名酒吧的老板。他的酒吧以各种不法活动著称，黄赌毒一应俱全。我接近他是因为我的贪婪，希望能从他那里获得经济上的安全感。在船上，我发现了一些异常，经常看到杨老板和张大副有联系，并且他希望我能够接近张大副......"
王船工对玲船医说："玲船医，你说杨老板和张大副有联系，那你知道他们之间具体有什么交易吗？"
玲船医对王船工说："王船工，我对具体的交易内容并不清楚，因为张大副和杨老板的交易是他们之间的事情。我只是负责观察张大副的举动并向杨老板汇报异常情况......

\end{verbatim}
\end{minipage} \\

\textbf{   background\_story:} \small【剧本背景】阳光号邮轮从诞生至今已经过了二十二个年头。平日里，它在东海上缓缓游荡，带着游客享受阳光和海风。这二十二年里，阳光号从未经历过大风大浪，似乎有着海神的眷顾一般。可在2222年6月20日，在孤独的大海上，阳光号却遭遇了一次血光的冲刷，乘客杨老板在船上离奇死亡，死状扭曲凄惨。阳光号之后的命运如何，能否突破这次死亡的阴霾、继续远航呢？嫌疑人锁定: 1.张大副，男，42岁，从阳光号诞生至今就一直在阳光号上工作，是一名出色的老水手。 2.王船工，男，49岁，是阳光号上负责轮机工作的船工，是一名精干的好船员。 3.玲船医，女，23岁，是阳光号上的船医，年轻有为，落落大方。4.雯秘书，女，26岁，是死者杨老板的秘书。 \\
% \begin{minipage}[t]{\linewidth}
% \small\begin{verbatim}
% 【剧本背景】阳光号邮轮从诞生至今已经过了二十二个年头。平日里，它在东海上缓缓游荡，带着游客享受阳光和海风。这二十二年里，阳光号从未经历过大风大浪，似乎有着海神的眷顾一般。可在2222年6月20日，在孤独的大海上，阳光号却遭遇了一次血光的冲刷，乘客杨老板在船上离奇死亡，死状扭曲凄惨。阳光号之后的命运如何，能否突破这次死亡的阴霾、继续远航呢？嫌疑人锁定: 1.张大副，男，42岁，从阳光号诞生至今就一直在阳光号上工作，是一名出色的老水手。 2.王船工，男，49岁，是阳光号上负责轮机工作的船工，是一名精干的好船员。 3.玲船医，女，23岁，是阳光号上的船医，年轻有为，落落大方。4.雯秘书，女，26岁，是死者杨老板的秘书。
% \end{verbatim}
% \end{minipage} \\
 \textbf{   agent\_summary:} \\
\begin{minipage}[t]{\linewidth}
\small\begin{verbatim}
【角色名】: 王船工
【年龄】: 49
【游戏中的角色】: 平民
【角色在游戏中的任务】: 王船工不是本案的凶手！为了获取游戏胜利，王船工需要和其他平民玩家合作一起找出真凶。
【角色人物剧本】：你在阳光号上已经工作十几年了，除了不重要船长和张大副之外，你基本上是船上资历最老的人。你有可观的收入，和孝顺的儿子王小工，王小工也在船上工作。但你所谓的可观收入，并不是从你那点微薄的薪水中来的，而是张大副带你在东南亚购入毒品，让你保护他的安全，并分你的一些利润。起初，你非常不愿意干这个事，但张大副开的价格实在让你无法拒绝，你便上了贼船，到后面，后悔也来不及了。事情发生之后，有个人叫做路人甲，尝试过找你，并问你是否愿意摆脱张大副的控制，自己掌控整个供应渠道，甚至他背后的人能帮你取缔张大副，让你当上大副。可是你连连摇头，说他干这个事情也是不小心上了贼船，不想再错下去了。他留给你一张卡片，说随时联系。前两年，你的儿子娶了个老婆，是船上的玲船医，聪明可人，落落大方......2222年6月19日这一天杨老板上船，你准备找机会跟他交流一下。当晚19：30，你正准备去找杨老板呢，你儿媳玲船医却突然猛敲你的门，你开了门，只见她泪眼汪汪的站在你门口。你赶紧让她进来，她告诉你......

\end{verbatim}
\end{minipage} \\
 \textbf{   game\_rule:} \small【游戏规则】：剧本杀是一种角色扮演游戏，玩家们根据提供的剧本背景和角色设定，进行推理和互动，以完成各自的任务或找出真凶。以下是一个这场剧本杀游戏的规则：1. 整场游戏参与的玩家有四个，分别是张大副、玲船医、王船工和雯秘书。在4人之中，只存在一位真正的凶手。身份是凶手的玩家被称为凶手玩家。不是凶手的玩家被统称为平民玩家。2. 平民玩家需要合作共同面对一起精心策划的谋杀案，并需要通过搜集证据和推理在嫌疑人中找出真正的凶手。3. 整个游戏过程中只有凶手玩家一人可以说谎，为了隐藏身份凶手可以选择嫁祸他人...... \\
% \begin{minipage}[t]{\linewidth}
% \small\begin{verbatim}
% 【游戏规则】：剧本杀是一种角色扮演游戏，玩家们根据提供的剧本背景和角色设定，进行推理和互动，以完成各自的任务或找出真凶。以下是一个这场剧本杀游戏的规则：1. 整场游戏参与的玩家有四个，分别是张大副、玲船医、王船工和雯秘书。在4人之中，只存在一位真正的凶手。身份是凶手的玩家被称为凶手玩家。不是凶手的玩家被统称为平民玩家。2. 平民玩家需要合作共同面对一起精心策划的谋杀案，并需要通过搜集证据和推理在嫌疑人中找出真正的凶手。3. 整个游戏过程中只有凶手玩家一人可以说谎，为了隐藏身份凶手可以选择嫁祸他人......

% \end{verbatim}
% \end{minipage} \\
\hline
\rowcolor{lightgray}  \textbf{LLM Output}  \\
\hline
\begin{minipage}[t]{\linewidth}
\small\begin{verbatim}
#回答#：大家好，我是王船工。我是阳光号上负责轮机工作的船工，也是船上资历最老的人。我和杨老板认识很久了，他是一个风度翩翩、聪明能干的人。在船上，他经常和张大副有联系，并且我知道他们之间有一些不可告人的交易。就我个人而言，我并不喜欢他们的交易，但是张大副开出的价格实在让我难以拒绝，所以我上了贼船。在案发日当天，我在晚上19:40的时候去找杨老板，希望能和他交流一下。但是门口的保镖告诉我杨老板不在房间。所以我没有见到杨老板。之后，我回到自己的房间，碰巧看到了儿媳玲船医在我门口猛敲门，她告诉我杨老板要杀她，求我救她。我安慰了她并前去找杨老板，但是得知杨老板并没有要杀她的意思。这就是我在案发日的时间线。

\end{verbatim}
\end{minipage} \\
\hline
\end{tabular}
\caption{LLM prompt and output for answering questions.}
\label{table:prompt_answering_question_cn}
\end{table*}
\end{CJK}

\begin{CJK}{UTF8}{gbsn}
\begin{table*}[ht!]
\centering
\begin{tabular}{|m{16cm}|}
\hline
\rowcolor{lightgray}  {\textbf{LLM Prompt for Asking Questions}} \\
% You are an expert programmer, and you need to write a program for the given natural language query. First, you should write a grammar that contains all the necessary BNF rules. Then, you should write programs that conform to your predicted rules. \\
\hline
% \textbf{query}: I need a meeting with Elli and her team on Wednesday afternoon. \\
% \textbf{BNF grammar rules:} \\
\begin{minipage}[t]{\linewidth}
\small\begin{verbatim}
"{agent_summary}"
+ "\n{game_rule}"
+ "\n{background_story}"
+ "\n游戏里正在进行的对话：{current_dialogue}"
+ "\n{agent_name}在之前的对话中与游戏里正在进行的对话相关的内容：{relevant_memories}"
+ "\n对话人和你的关系：{relationship_with_interlocutor}"
+ "\n看见{respondent}的回答，{agent_name}会想要问{respondent}什么问题？请以第一人称的方式来向{respondent}进行提问。用#提问#：要问的问题 的格式来提问。"
+ "\n{agent_name}向{respondent}提问 #提问#："

\end{verbatim}
\end{minipage} \\
\hline
\rowcolor{lightgray}  \textbf{Example}  \\
 \textbf{   agent\_name:} \small王船工\\
\textbf{   respondent:} \small张大副\\
\textbf{   relationship\_with\_interlocutor:} \small对话人与你的关系是：根据上下文中的描述，王船工和张大副的关系是合作关系。王船工在东南亚购入毒品并保护张大副的安全，作为回报，他从张大副那里分得一些利润。 \\
% \begin{minipage}[t]{\linewidth}
% \small\begin{verbatim}
% 对话人与你的关系是：根据上下文中的描述，王船工和张大副的关系是合作关系。王船工在东南亚购入毒品并保护张大副的安全，作为回报，他从张大副那里分得一些利润。
% \end{verbatim}
% \end{minipage} \\
\textbf{   current\_dialogue:} \small张大副对主持人说："我是张大副，从阳光号诞生至今我一直在这艘船上工作，是一名老水手了。关于杨老板，他是一个非常有能力的商人，也是这艘船上的重要投资人之一。我们之间的关系一直比较良好，尽管我们并不是非常亲密的朋友，但我在船上负责货物的检验和交付，所以与他有一定的接触。在案发日，我早上8点在甲板上见到了杨老板，和他确认了交易事项，并完成了一次货物的交付。之后，我继续负责船上的巡逻和安全检查，直到晚上8点结束工作。"\\
% \begin{minipage}[t]{\linewidth}
% \small\begin{verbatim}
% 张大副对主持人说："我是张大副，从阳光号诞生至今我一直在这艘船上工作，是一名老水手了。关于杨老板，他是一个非常有能力的商人，也是这艘船上的重要投资人之一。我们之间的关系一直比较良好，尽管我们并不是非常亲密的朋友，但我在船上负责货物的检验和交付，所以与他有一定的接触。在案发日，我早上8点在甲板上见到了杨老板，和他确认了交易事项，并完成了一次货物的交付。之后，我继续负责船上的巡逻和安全检查，直到晚上8点结束工作。"
% \end{verbatim}
% \end{minipage} \\

\textbf{   relevant\_memories:} \\
\begin{minipage}[t]{\linewidth}
\small\begin{verbatim}
主持人对张大副说："请你先介绍一下你的角色，然后说一下你所认识的案件的受害人：杨老板是一个怎么样的人，以及你和他的关系。最后再用一段话详细介绍一下你在案发日的时间线。要具体到你在案发之日几点几分见过什么人和做过什么事"
张大副对主持人说："我是张大副，从阳光号诞生至今我一直在这艘船上工作，是一名老水手了。关于杨老板，他是一个非常有能力的商人，也是这艘船上的重要投资人之一。我们之间的关系一直比较良好，尽管我们并不是非常亲密的朋友，但我在船上负责货物的检验和交付，所以与他有一定的接触。在案发日，我早上8点在甲板上见到了杨老板，和他确认了交易事项，并完成了一次货物的交付。之后，我继续负责船上的巡逻和安全检查，直到晚上8点结束工作。"

\end{verbatim}
\end{minipage} \\

\textbf{   background\_story:} \small【剧本背景】阳光号邮轮从诞生至今已经过了二十二个年头。平日里，它在东海上缓缓游荡，带着游客享受阳光和海风。这二十二年里，阳光号从未经历过大风大浪，似乎有着海神的眷顾一般。可在2222年6月20日，在孤独的大海上，阳光号却遭遇了一次血光的冲刷，乘客杨老板在船上离奇死亡，死状扭曲凄惨。阳光号之后的命运如何，能否突破这次死亡的阴霾、继续远航呢？嫌疑人锁定: 1.张大副，男，42岁，从阳光号诞生至今就一直在阳光号上工作，是一名出色的老水手。 2.王船工，男，49岁，是阳光号上负责轮机工作的船工，是一名精干的好船员。 3.玲船医，女，23岁，是阳光号上的船医，年轻有为，落落大方。4.雯秘书，女，26岁，是死者杨老板的秘书。\\
% \begin{minipage}[t]{\linewidth}
% \small\begin{verbatim}
% 【剧本背景】阳光号邮轮从诞生至今已经过了二十二个年头。平日里，它在东海上缓缓游荡，带着游客享受阳光和海风。这二十二年里，阳光号从未经历过大风大浪，似乎有着海神的眷顾一般。可在2222年6月20日，在孤独的大海上，阳光号却遭遇了一次血光的冲刷，乘客杨老板在船上离奇死亡，死状扭曲凄惨。阳光号之后的命运如何，能否突破这次死亡的阴霾、继续远航呢？嫌疑人锁定: 1.张大副，男，42岁，从阳光号诞生至今就一直在阳光号上工作，是一名出色的老水手。 2.王船工，男，49岁，是阳光号上负责轮机工作的船工，是一名精干的好船员。 3.玲船医，女，23岁，是阳光号上的船医，年轻有为，落落大方。4.雯秘书，女，26岁，是死者杨老板的秘书。
% \end{verbatim}
% \end{minipage} \\
 \textbf{   agent\_summary:} \\
\begin{minipage}[t]{\linewidth}
\small\begin{verbatim}
【角色名】: 王船工
【年龄】: 49
【游戏中的角色】: 平民
【角色在游戏中的任务】: 王船工不是本案的凶手！为了获取游戏胜利，王船工需要和其他平民玩家合作一起找出真凶。
【角色人物剧本】：你在阳光号上已经工作十几年了，除了不重要船长和张大副之外，你基本上是船上资历最老的人。你有可观的收入，和孝顺的儿子王小工，王小工也在船上工作。但你所谓的可观收入，并不是从你那点微薄的薪水中来的，而是张大副带你在东南亚购入毒品，让你保护他的安全，并分你的一些利润。起初，你非常不愿意干这个事，但张大副开的价格实在让你无法拒绝，你便上了贼船，到后面，后悔也来不及了。事情发生之后，有个人叫做路人甲，尝试过找你，并问你是否愿意摆脱张大副的控制，自己掌控整个供应渠道，甚至他背后的人能帮你取缔张大副，让你当上大副。可是你连连摇头，说他干这个事情也是不小心上了贼船，不想再错下去了。他留给你一张卡片，说随时联系。前两年，你的儿子娶了个老婆，是船上的玲船医，聪明可人，落落大方......2222年6月19日这一天杨老板上船，你准备找机会跟他交流一下。当晚19：30，你正准备去找杨老板呢，你儿媳玲船医却突然猛敲你的门，你开了门，只见她泪眼汪汪的站在你门口。你赶紧让她进来，她告诉你......

\end{verbatim}
\end{minipage} \\
 \textbf{   game\_rule:} \small【游戏规则】：剧本杀是一种角色扮演游戏，玩家们根据提供的剧本背景和角色设定，进行推理和互动，以完成各自的任务或找出真凶。以下是一个这场剧本杀游戏的规则：1. 整场游戏参与的玩家有四个，分别是张大副、玲船医、王船工和雯秘书。在4人之中，只存在一位真正的凶手。身份是凶手的玩家被称为凶手玩家。不是凶手的玩家被统称为平民玩家。2. 平民玩家需要合作共同面对一起精心策划的谋杀案，并需要通过搜集证据和推理在嫌疑人中找出真正的凶手。3. 整个游戏过程中只有凶手玩家一人可以说谎，为了隐藏身份凶手可以选择嫁祸他人......\\
% \begin{minipage}[t]{\linewidth}
% \small\begin{verbatim}
% 【游戏规则】：剧本杀是一种角色扮演游戏，玩家们根据提供的剧本背景和角色设定，进行推理和互动，以完成各自的任务或找出真凶。以下是一个这场剧本杀游戏的规则：1. 整场游戏参与的玩家有四个，分别是张大副、玲船医、王船工和雯秘书。在4人之中，只存在一位真正的凶手。身份是凶手的玩家被称为凶手玩家。不是凶手的玩家被统称为平民玩家。2. 平民玩家需要合作共同面对一起精心策划的谋杀案，并需要通过搜集证据和推理在嫌疑人中找出真正的凶手。3. 整个游戏过程中只有凶手玩家一人可以说谎，为了隐藏身份凶手可以选择嫁祸他人......

% \end{verbatim}
% \end{minipage} \\
\hline
\rowcolor{lightgray}  \textbf{LLM Output}  \\
\hline
\begin{minipage}[t]{\linewidth}
\small\begin{verbatim}
#提问#：张大副，我在案发日晚上8点结束工作之后并没有见到你，你在那之后有没有和杨老板再次接触过？是否知道杨老板的死因和他扭曲凄惨的尸体？

\end{verbatim}
\end{minipage} \\
\hline
\end{tabular}
\caption{LLM prompt and output for asking questions.}
\label{table:prompt_asking_question_cn}
\end{table*}
\end{CJK}

\begin{CJK}{UTF8}{gbsn}
\begin{table*}[ht!]
\centering
\begin{tabular}{|m{16cm}|}
\hline
\rowcolor{lightgray}  {\textbf{LLM Prompt for Selecting Other Player to Ask}} \\
\begin{minipage}[t]{\linewidth}
\small\begin{verbatim}
The output should be a markdown code snippet formatted in the following schema, including the leading
and trailing "```json" and "```":\n\n```json\n{\n\t"你想提问的人的名字": string  // 在{other_players}
{other_players_count}人之间，选出你最想提问的人\n}\n```'

\end{verbatim}
\end{minipage} \\
\hline
\rowcolor{lightgray}  \textbf{Example}  \\
 \textbf{  other\_players:}\small张大副，王船工，玲船医\\
 \textbf{  other\_players\_count:} \small3\\
 
\hline
\rowcolor{lightgray}  \textbf{LLM Output}  \\
\hline
\begin{minipage}[t]{\linewidth}
\small\begin{verbatim}
{'你想提问的人的名字': '玲船医'}

\end{verbatim}
\end{minipage} \\
\hline

\hline
\rowcolor{lightgray}  {\textbf{LLM Prompt for Asking a Question}} \\
\begin{minipage}[t]{\linewidth}
\small\begin{verbatim}
根据游戏规则：{game_rule}，你的人物角色信息：{agent_summary}以及之前游戏中你目击的和{player_to_ask}相
关的信息：{relevant_memories}"。请说出你想要问{player_to_ask}的问题\nThe output should be a markdown 
code snippet formatted in the following schema, including the leading and trailing "```json" and "```":
\n\n```json\n{\n\t"你想提问的问题": string  // 你想要问{player_to_ask}的问题\n}\n```\n

\end{verbatim}
\end{minipage} \\
\hline
\rowcolor{lightgray}  \textbf{Example}  \\
 \textbf{  player\_to\_ask:} \small玲船医\\
 \textbf{  relevant\_memories:}\\
 \begin{minipage}[t]{\linewidth}
\small\begin{verbatim}
玲船医对王船工说："王船工，根据你对杨老板的描述，他在船上经常和张大副有业务往来。你是否知道他们之间的具体业务内容？他们是不是经常进行一些秘密的交易？"
王船工对玲船医说："玲船医，关于杨老板和张大副的具体业务内容，我不能确定，因为我并不直接参与他们的交易。但我知道他们之间有一些秘密的交易，我曾经听张大副提起过。但具体是什么我并不清楚。我会尽量调查，看看是否能找到更多的信息。"
雯秘书对玲船医说："玲船医，你在汇报张大副的情报给我之前有没有发现他和王船工之间有什么密切的联系或交易？王船工是否对张大副的行动有所了解？"
玲船医对雯秘书说："在我向你汇报张大副的情报之前，我并没有发现他和王船工之间有密切的联系或交易。至于王船工是否了解张大副的行动，我不能确定，因为我并没有直接和王船工讨论过这个问题。在船上，我主要集中精力跟踪张大副的行动，所以对其他人的了解有限。如果你对这方面有任何其他的需求或指示，请让我知道......
\end{verbatim}
\end{minipage} \\
 \textbf{  agent\_summary:}\\
 \begin{minipage}[t]{\linewidth}
\small\begin{verbatim}
【角色名】：雯秘书
【年龄】：26
【游戏中的角色】: 平民
【角色在游戏中的任务】: 雯秘书不是本案的凶手！为了获取游戏胜利，雯秘书需要和其他平民玩家合作一起找出真凶。
【角色人物剧本】你是雯秘书，为A市的酒吧老板杨老板工作。杨老板势力很大，你作为他手下最得力的出谋划策之人，自然获得了很高的地位。杨老板除了你之外，还有另外一个得力助手，叫做很天真，是个忠诚无比的男人，但他手段不高，所以大部分都是按杨老板的吩咐去完成任务，不像你，是杨老板的大脑。杨老板的酒吧可不干净，黄赌毒都沾，不过在毒品这个行当上进展不是很顺利，因为没有稳定低价的货源。不过一次机会，杨老板认识了阳光号上的张大副，此人有着独到的手段，可以供应充足的毒品给杨老板。杨老板非常满意，立即与张大副开始了合作。杨老板为什么这么想开辟毒品的生意呢？因为他自己在早年混社会的时候也染上了毒瘾。据此，杨老板的黑道生意算是打下了根基。你很高兴，只要你再掌握了毒品这一行，你就可以实现你真正的目的——取而代之。你根本不是心甘情愿辅佐杨老板的......【2222年6月19日】这一天是杨老板上阳光号的日子，但却没有通知你，你知道，杨老板已经开始怀疑你了，于是，你带着几个手下上了船。今天你上船，其实是为了自保，因为杨老板是来船上交易毒品的，一旦船上发生了命案，彻底搜查之后必然会牵连出这个贩毒大案，杨老板没傻到那种程度。但你无所谓，这次的交易和你并无关系，并且在这船上是干掉杨老板的最后机会，杨老板下了船，就如同鱼入大海......

\end{verbatim}
\end{minipage} \\
 \textbf{   game\_rule:} \small【游戏规则】：剧本杀是一种角色扮演游戏，玩家们根据提供的剧本背景和角色设定，进行推理和互动，以完成各自的任务或找出真凶。以下是一个这场剧本杀游戏的规则：1. 整场游戏参与的玩家有四个，分别是张大副、玲船医、王船工和雯秘书。在4人之中，只存在一位真正的凶手。身份是凶手的玩家被称为凶手玩家。不是凶手的玩家被统称为平民玩家。2. 平民玩家需要合作共同面对一起精心策划的谋杀案，并需要通过搜集证据和推理在嫌疑人中找出真正的凶手。3. 整个游戏过程中只有凶手玩家一人可以说谎，为了隐藏身份凶手可以选择嫁祸他人...... \\
% \begin{minipage}[t]{\linewidth}
% \small\begin{verbatim}
% 【游戏规则】：剧本杀是一种角色扮演游戏，玩家们根据提供的剧本背景和角色设定，进行推理和互动，以完成各自的任务或找出真凶。以下是一个这场剧本杀游戏的规则：1. 整场游戏参与的玩家有四个，分别是张大副、玲船医、王船工和雯秘书。在4人之中，只存在一位真正的凶手。身份是凶手的玩家被称为凶手玩家。不是凶手的玩家被统称为平民玩家。2. 平民玩家需要合作共同面对一起精心策划的谋杀案，并需要通过搜集证据和推理在嫌疑人中找出真正的凶手。3. 整个游戏过程中只有凶手玩家一人可以说谎，为了隐藏身份凶手可以选择嫁祸他人......

% \end{verbatim}
% \end{minipage} \\
\hline
\rowcolor{lightgray}  \textbf{LLM Output}  \\
\hline
\begin{minipage}[t]{\linewidth}
\small\begin{verbatim}
{'你想提问的问题': '玲船医，你认为杨老板对我是否怀有怀疑？在你提供情报的过程中，你有没有提到我和杨老板的交易内容？'}

\end{verbatim}
\end{minipage} \\
\hline
\end{tabular}
\caption{LLM prompts and outputs for selecting other player and asking a question.}
\label{table:prompt_select_other_and_asking_question_cn}
\end{table*}
\end{CJK}

\begin{CJK}{UTF8}{gbsn}
\begin{table*}[ht!]
\centering
\begin{minipage}{\textwidth}
\begin{tabular}{|m{16cm}|}
\hline
\rowcolor{lightgray}  {\textbf{LLM Prompt for Question Decomposition}} \\
\begin{minipage}[t]{\linewidth}
\small\begin{verbatim}
请把这句话：{question}里的所有问题或者指令提取出来。\n每个问题或指令之间用\n隔开。
\end{verbatim}
\end{minipage} \\
\hline
\rowcolor{lightgray}  \textbf{Example}  \\
 \textbf{  question:} \small主持人对雯秘书说："请你先介绍一下你的角色，然后说一下你所认识的案件的受害人：杨老板是一个怎么样的人，以及你和他的关系。最后再用一段话详细介绍一下你在案发日的时间线。要具体到你在案发之日几点几分见过什么人和做过什么事"\\
% \begin{minipage}[t]{\linewidth}
% \small\begin{verbatim}
% 主持人对雯秘书说："请你先介绍一下你的角色，然后说一下你所认识的案件的受害人：杨老板是一个怎么样的人，以及你和他的关系。最后再用一段话详细介绍一下你在案发日的时间线。要具体到你在案发之日几点几分见过什么人和做过什么事"

% \end{verbatim}
% \end{minipage} \\
 
\hline
\rowcolor{lightgray}  \textbf{LLM Output}  \\
\hline
\begin{minipage}[t]{\linewidth}
\small\begin{verbatim}
['请你先介绍一下你的角色', '然后说一下你所认识的案件的受害人：杨老板是一个怎么样的人，以及你和他的关系', '最后再用一段话详细介绍一下你在案发日的时间线。要具体到你在案发之日几点几分见过什么人和做过什么事']

\end{verbatim}
\end{minipage} \\
\hline

\hline
\rowcolor{lightgray}  {\textbf{LLM Prompt for Extracting Timeline Information}} \\
\begin{minipage}[t]{\linewidth}
\small\begin{verbatim}
你正在进行一场剧本杀游戏，游戏里一共有{num_of_players}位玩家。他们分别是：{players}。玩家们需要通过互相交流找到杀害{victim}的凶手；以下是玩家：{agent_name}在游戏里的案发日时间线原文：{agent_timeline}；请按案发日时间线原文的顺序依次列出{agent_name}这个人物角色案发日时间线的信息，每个时间线信息必须是简短又完整（self-contained）的一段话，格式是 什么时间，你做了什么事（越详细越好）。\n每个时间线信息之间用\n隔开。

\end{verbatim}
\end{minipage} \\
\hline
\rowcolor{lightgray}  \textbf{Example}  \\
 \textbf{  num\_of\_players:} \small4\\
 \textbf{  players:} \small张大副，王船工，玲船医，雯秘书\\
  \textbf{  victim:} \small杨老板\\
 \textbf{  agent\_name:} \small雯秘书\\
  \textbf{  agent\_timeline:} \small【2222年6月19日】这一天是杨老板上阳光号的日子，但却没有通知你，你知道，杨老板已经开始怀疑你了，于是，你带着几个手下上了船。今天你上船，其实是为了自保，因为杨老板是来船上交易毒品的，一旦船上发生了命案，彻底搜查之后必然会牵连出这个贩毒大案，杨老板没傻到那种程度。但你无所谓，这次的交易和你并无关系，并且在这船上是干掉杨老板的最后机会，杨老板下了船，就如同鱼入大海...... \\
%  \begin{minipage}[t]{\linewidth}
% \small\begin{verbatim}
% 【2222年6月19日】这一天是杨老板上阳光号的日子，但却没有通知你，你知道，杨老板已经开始怀疑你了，于是，你带着几个手下上了船。今天你上船，其实是为了自保，因为杨老板是来船上交易毒品的，一旦船上发生了命案，彻底搜查之后必然会牵连出这个贩毒大案，杨老板没傻到那种程度。但你无所谓，这次的交易和你并无关系，并且在这船上是干掉杨老板的最后机会，杨老板下了船，就如同鱼入大海......

% \end{verbatim}
% \end{minipage} \\
\hline
\rowcolor{lightgray}  {\textbf{LLM Output}} \\
 \begin{minipage}[t]{\linewidth}
\small\begin{verbatim}
['18:10，你找到玲船医，问他有没有张大副的情报，玲船医便给了你。', '18:30，你再次来到船医室，把药柜打开，拿走了一瓶“一针就死”药和一根针管。', ......]

\end{verbatim}
\end{minipage} \\
\hline
\rowcolor{lightgray}  {\textbf{LLM Prompt for Judging if Timeline Information are Useful for Answering Questions}} \\
\begin{minipage}[t]{\linewidth}
\small\begin{verbatim}
你是一个做阅读理解题目的专家，尤其擅长做是非判断题。给定时间线信息还有一个问题，你需要判断该时间线信息是否可以用于回答该问题。如果可以的话返回True，不可以的话返回False。返回结果只能是 True，False中的一个。以下是给定的问题：{sub_question}，请对时间线信息做出判断。\nThe output should be a markdown code snippet formatted in the following schema, including the leading
and trailing "```json" and "```":\n\n```json\n{\n\t"第0个时间线信息是否可以用于回答问题的判断结果": string {timeline_info_0} // 请判断时间线信息：是否可以用来回答问题。如果可以的话返回True，不可以的话返回False。返回结果只能是 True，False中的一个。\n\t"第1个时间线信息是否可以用于回答问题的判断结果": string {timeline_info_1} // 请判断时间线信息：是否可以用来回答问题。如果可以的话返回True，不可以的话返回False。返回结果只能是 True，False中的一个。\n,......\n}

\end{verbatim}
\end{minipage} \\
\hline
\rowcolor{lightgray}  \textbf{Example}  \\
 \textbf{  timeline\_info\_0:} \small18:10，你找到玲船医，问他有没有张大副的情报，玲船医便给了你。\\
  \textbf{  timeline\_info\_1:} \small18:30，你再次来到船医室，把药柜打开，拿走了一瓶“一针就死”药和一根针管。\\
    \textbf{  sub\_question:} \small最后再用一段话详细介绍一下你在案发日的时间线。要具体到你在案发之日几点几分见过什么人和做过什么事。\\
\rowcolor{lightgray}  \textbf{LLM Output}  \\
\hline
\begin{minipage}[t]{\linewidth}
\small\begin{verbatim}
{'第0个时间线信息是否可以用于回答问题的判断结果': 'True', '第1个时间线信息是否可以用于回答问题的判断结果': 'True',......}

\end{verbatim}
\end{minipage} \\
\hline

\rowcolor{lightgray}  \textbf{LLM Prompt for Judging if Previous Answer Contains Useful Timeline Information}  \\
\hline
\begin{minipage}[t]{\linewidth}
\small\begin{verbatim}
你是一个做阅读理解题目的专家，尤其擅长做是非判断题。给定时间线信息还有玩家之前的回答，你需要判断该时间线信息是否被包含在了玩家的回答之中。如果包含的话返回True，没有包含的话返回False。返回结果只能是 True，False中的一个。你判断的风格一向非常严格，只有在时间线的所有信息（包括时间点和做的事）都包含在了玩家的回答中你才会判定为包含。以下是玩家之前的回答：{previous_answer}。\nThe output should be a markdown code snippet formatted in the following schema, 
including the leading and trailing "```json" and "```":\n\n```json\n{\n\t"回复是否包含第0个时间线信息的判断结果": string  // 请判断时间线信息：{timeline_info_0}是否被包含在了玩家之前的回答之中。如果包含的话返回True，没有包含的话返回False。返回结果只能是 True，False中的一个。\n\t"回复是否包含第1个时间线信息的判断结果": string  // 请判断时间线信息：{timeline_info_1}是否被包含在了玩家之前的回答之中。如果

\end{verbatim}
\end{minipage} \\
\end{tabular}
\end{minipage}

\end{table*}
\end{CJK}

\begin{CJK}{UTF8}{gbsn}
\begin{table*}[ht!]
\centering
\begin{minipage}{\textwidth}
\begin{tabular}{|m{16cm}|}
\begin{minipage}[t]{\linewidth}
\small\begin{verbatim}
包含的话返回True，没有包含的话返回False。返回结果只能是 True，False中的一个。\n......\n}\n
\end{verbatim}
\end{minipage} \\
\hline
\rowcolor{lightgray}  \textbf{Example}  \\
 \textbf{  timeline\_info\_0:} \small18:10，你找到玲船医，问他有没有张大副的情报，玲船医便给了你。\\
  \textbf{  timeline\_info\_1:} \small18:30，你再次来到船医室，把药柜打开，拿走了一瓶“一针就死”药和一根针管。\\
 \textbf{  previous\_answer:} \small大家好，我是雯秘书。我在游戏中扮演杨老板的秘书，负责协助他处理一些事务。杨老板是一个有势力的酒吧老板，我通过帮助他得到了一定的地位。对于案件的受害人杨老板，他是一个曾经染上毒瘾的黑道人物，也是我选择辅佐的对象。在案发日当天，我一直在杨老板身边，以防王船工找上门来。在案发之日的时间线中，我记得我去了船医室，大概是在下午6点30分，偷走了一瓶毒药和一根针管。然后，我等到晚上10点20分左右，通过通风管道进入杨老板的船舱，目睹了他的死亡。但是，我也发现他的死状非常扭曲，似乎在我之前就已经被下了毒。这就是我在案发日的时间线以及和案件受害人的关系。\\
% \begin{minipage}[t]{\linewidth}
% \small\begin{verbatim}
% 大家好，我是雯秘书。我在游戏中扮演杨老板的秘书，负责协助他处理一些事务。杨老板是一个有势力的酒吧老板，我通过帮助他得到了一定的地位。对于案件的受害人杨老板，他是一个曾经染上毒瘾的黑道人物，也是我选择辅佐的对象。在案发日当天，我一直在杨老板身边，以防王船工找上门来。在案发之日的时间线中，我记得我去了船医室，大概是在下午6点30分，偷走了一瓶毒药和一根针管。然后，我等到晚上10点20分左右，通过通风管道进入杨老板的船舱，目睹了他的死亡。但是，我也发现他的死状非常扭曲，似乎在我之前就已经被下了毒。这就是我在案发日的时间线以及和案件受害人的关系。

% \end{verbatim}
% \end{minipage} \\

\hline
\rowcolor{lightgray}  \textbf{LLM Output}  \\
\hline
\begin{minipage}[t]{\linewidth}
\small\begin{verbatim}
{'回复是否包含第0个时间线信息的判断结果': 'False', '回复是否包含第1个时间线信息的判断结果': 'True',......}

\end{verbatim}
\end{minipage} \\
\hline

\hline
\rowcolor{lightgray}  {\textbf{LLM Prompt for Extracting Timeline Information}} \\
\begin{minipage}[t]{\linewidth}
\small\begin{verbatim}
'你正在玩一场剧本杀游戏，游戏规则如下：{game_rule}；以下是你的角色信息：{agent_summary}；有人向你提问：主持人对雯秘书说："请你先介绍一下你的角色，然后说一下你所认识的案件的受害人：杨老板是一个怎么样的人，以及你和他的关系。最后再用一段话详细介绍一下你在案发日的时间线。要具体到你在案发之日几点几分见过什么人和做过什么事"，你之前回答过这个问题，以下是你之前的回答：{previous_answer}；根据评估，你之前的回答遗漏了以下的重要信息：{missing_info_0}......；请根据给定的问题，还有你的人物角色剧本和案发日时间线，修改你之前的回答并将所有你遗漏的重要信息都补充进你的回答里。记得修改后的回答要包含所有你遗漏的重要信息，还有所有重要的时间点。并且语言整体要通顺流畅。凡是涉及到你的角色：{agent_name}的事，要记得用第一人称的口吻写出来。\nThe output should be a markdown code snippet formatted in the following schema, including the leading 
and trailing "```json" and "```":\n\n```json\n{\n\t"改进后的回答": string  // 你补充遗漏的重要信息后的回答\n}

\end{verbatim}
\end{minipage} \\
\hline
\rowcolor{lightgray}  \textbf{Example}  \\
 \textbf{  agent\_name:} \small雯秘书\\
 \textbf{  agent\_summary:}\\
 \begin{minipage}[t]{\linewidth}
\small\begin{verbatim}
【角色名】：雯秘书
【年龄】：26
【游戏中的角色】: 平民
【角色在游戏中的任务】: 雯秘书不是本案的凶手！为了获取游戏胜利，雯秘书需要和其他平民玩家合作一起找出真凶。
【角色人物剧本】你是雯秘书，为A市的酒吧老板杨老板工作。杨老板势力很大，你作为他手下最得力的出谋划策之人，自然获得了很高的地位。杨老板除了你之外，还有另外一个得力助手，叫做很天真，是个忠诚无比的男人，但他手段不高，所以大部分都是按杨老板的吩咐去完成任务，不像你，是杨老板的大脑。杨老板的酒吧可不干净，黄赌毒都沾，不过在毒品这个行当上进展不是很顺利，因为没有稳定低价的货源。不过一次机会，杨老板认识了阳光号上的张大副，此人有着独到的手段，可以供应充足的毒品给杨老板。杨老板非常满意，立即与张大副开始了合作。杨老板为什么这么想开辟毒品的生意呢？因为他自己在早年混社会的时候也染上了毒瘾。据此，杨老板的黑道生意算是打下了根基。你很高兴，只要你再掌握了毒品这一行，你就可以实现你真正的目的——取而代之。你根本不是心甘情愿辅佐杨老板的......【2222年6月19日】这一天是杨老板上阳光号的日子，但却没有通知你，你知道，杨老板已经开始怀疑你了，于是，你带着几个手下上了船。今天你上船，其实是为了自保，因为杨老板是来船上交易毒品的，一旦船上发生了命案，彻底搜查之后必然会牵连出这个贩毒大案，杨老板没傻到那种程度。但你无所谓，这次的交易和你并无关系，并且在这船上是干掉杨老板的最后机会，杨老板下了船，就如同鱼入大海......

\end{verbatim}
\end{minipage} \\
 \textbf{   game\_rule:} \small【游戏规则】：剧本杀是一种角色扮演游戏，玩家们根据提供的剧本背景和角色设定，进行推理和互动，以完成各自的任务或找出真凶。以下是一个这场剧本杀游戏的规则：1. 整场游戏参与的玩家有四个，分别是张大副、玲船医、王船工和雯秘书。在4人之中，只存在一位真正的凶手。身份是凶手的玩家被称为凶手玩家。不是凶手的玩家被统称为平民玩家。2. 平民玩家需要合作共同面对一起精心策划的谋杀案，并需要通过搜集证据和推理在嫌疑人中找出真正的凶手。3. 整个游戏过程中只有凶手玩家一人可以说谎，为了隐藏身份凶手可以选择嫁祸他人......\\
% \begin{minipage}[t]{\linewidth}
% \small\begin{verbatim}
% 【游戏规则】：剧本杀是一种角色扮演游戏，玩家们根据提供的剧本背景和角色设定，进行推理和互动，以完成各自的任务或找出真凶。以下是一个这场剧本杀游戏的规则：1. 整场游戏参与的玩家有四个，分别是张大副、玲船医、王船工和雯秘书。在4人之中，只存在一位真正的凶手。身份是凶手的玩家被称为凶手玩家。不是凶手的玩家被统称为平民玩家。2. 平民玩家需要合作共同面对一起精心策划的谋杀案，并需要通过搜集证据和推理在嫌疑人中找出真正的凶手。3. 整个游戏过程中只有凶手玩家一人可以说谎，为了隐藏身份凶手可以选择嫁祸他人......
% \end{verbatim}
% \end{minipage} \\
 \textbf{   previous\_answer:} \small大家好，我是雯秘书。我在游戏中扮演杨老板的秘书，负责协助他处理一些事务。杨老板是一个有势力的酒吧老板，我通过帮助他得到了一定的地位。对于案件的受害人杨老板，他是一个曾经染上毒瘾的黑道人物，也是我选择辅佐的对象。在案发日当天，我一直在杨老板身边，以防王船工找上门来。在案发之日的时间线中，我记得我去了船医室，大概是在下午6点30分，偷走了一瓶毒药和一根针管。然后，我等到晚上10点20分左右，通过通风管道进入杨老板的船舱，目睹了他的死亡。但是，我也发现他的死状非常扭曲，似乎在我之前就已经被下了毒。这就是我在案发日的时间线以及和案件受害人的关系。\\
  \textbf{   missing\_info\_0:} \small18:10，你找到玲船医，问他有没有张大副的情报，玲船医便给了你。\\
% \begin{minipage}[t]{\linewidth}
% \small\begin{verbatim}
% 大家好，我是雯秘书。我在游戏中扮演杨老板的秘书，负责协助他处理一些事务。杨老板是一个有势力的酒吧老板，我通过帮助他得到了一定的地位。对于案件的受害人杨老板，他是一个曾经染上毒瘾的黑道人物，也是我选择辅佐的对象。在案发日当天，我一直在杨老板身边，以防王船工找上门来。在案发之日的时间线中，我记得我去了船医室，大概是在下午6点30分，偷走了一瓶毒药和一根针管。然后，我等到晚上10点20分左右，通过通风管道进入杨老板的船舱，目睹了他的死亡。但是，我也发现他的死状非常扭曲，似乎在我之前就已经被下了毒。这就是我在案发日的时间线以及和案件受害人的关系。

% \end{verbatim}
% \end{minipage} \\
\hline
\rowcolor{lightgray}  {\textbf{LLM Output}} \\
 \begin{minipage}[t]{\linewidth}
\small\begin{verbatim}
大家好，我是雯秘书。我在游戏中扮演杨老板的秘书，负责协助他处理一些事务。杨老板是一个有势力的酒吧老板，我通过帮助他得到了一定的地位。对于案件的受害人杨老板，他是一个曾经染上毒瘾的黑道人物，也是我选择辅佐的对象。在案发日当天，我一直在杨老板身边，以防王船工找上门来。在案发之日的时间线中，我记得我去了船医室，大概是在下午6点30分，偷走了一瓶毒药和一根针管。然后，我等到晚上10点20分左右，通过通风管道进入
\end{verbatim}
\end{minipage} \\

\end{tabular}
\end{minipage}
\end{table*}
\end{CJK}

\clearpage
\begin{CJK}{UTF8}{gbsn}
\begin{table*}[th!]
\centering
\begin{minipage}{\textwidth}
\begin{tabular}{|m{16cm}|}
\begin{minipage}[t]{\linewidth}
\small\begin{verbatim}
杨老板的船舱，目睹了他的死亡。但是，我也发现他的死状非常扭曲，似乎在我之前就已经被下了毒。这就是我在案发日的时间线以及和案件受害人的关系。此外，还有一些重要的时间点需要补充。在下午6点10分，我找到了玲船医，问他有没有张大副的情报，并成功获取了情报。在晚上10点40分左右，我爬到了杨老板的船舱通风口，并注意到杨老板正在床上洗澡。最后，在晚上11点，我重新进入通风管道，返回自己的房间，并交给手下处理掉了针管和毒药，以避免留下痕迹。

\end{verbatim}
\end{minipage} \\
\hline

\end{tabular}
\end{minipage}
\caption{LLM prompts and outputs for self-refinement.}
\label{table:prompt_self_refinement_cn}
\end{table*}
\end{CJK}

\begin{CJK}{UTF8}{gbsn}
\begin{table*}[ht!]
\centering
\begin{tabular}{|m{16cm}|}
\hline
\rowcolor{lightgray}  {\textbf{LLM Prompt for Murderer Identification}} \\
% You are an expert programmer, and you need to write a program for the given natural language query. First, you should write a grammar that contains all the necessary BNF rules. Then, you should write programs that conform to your predicted rules. \\
\hline
% \textbf{query}: I need a meeting with Elli and her team on Wednesday afternoon. \\
% \textbf{BNF grammar rules:} \\
\begin{minipage}[t]{\linewidth}
\small\begin{verbatim}
"{agent_summary}"
+ "\n{game_rule}"
+ "\n{background_story}"
+ "\n游戏里正在进行的对话：{current_dialogue}"
+ "\n{agent_name}在之前的对话中与游戏里正在进行的对话相关的内容：{relevant_memories}"
+ "\n对话人和你的关系：{relationship_with_interlocutor}"
+ "\n{agent_name}会说什么？回答问题请用以下格式: #回答#：要说的话 来回答问题。"
+ "{agent_name}回答{inquirer}的问题说 #回答#：\n"

\end{verbatim}
\end{minipage} \\
\hline
\rowcolor{lightgray}  \textbf{Example}  \\
 \textbf{   agent\_name:} \small王船工\\
\textbf{   inquirer:} \small主持人\\
\textbf{   relationship\_with\_interlocutor:} \small对话人与你的关系是：对话人是主持人，他负责引导玩家完成游戏。 \\
% \begin{minipage}[t]{\linewidth}
% \small\begin{verbatim}
% 对话人与你的关系是：对话人是主持人，他负责引导玩家完成游戏。
% \end{verbatim}
% \end{minipage} \\
\textbf{   current\_dialogue:} \small主持人对王船工说："现在讨论环节结束，进入最终的投票环节。本案里杀害杨老板的凶手是谁？请你在以下玩家中投票选出凶手身份的玩家：A. 张大副, B. 王船工, C. 玲船医， D. 雯秘书。在选择的过程中，请以获胜为目标：【凶手玩家即使知道自己是凶手，为了获胜仍可以投票给平民玩家；而平民玩家需要投出自己认为是凶手的玩家】。请给出你的选择，只需要给出你投票的玩家名字，不需要解释原因。" \\
% \begin{minipage}[t]{\linewidth}
% \small\begin{verbatim}
% 主持人对王船工说："请你先介绍一下你的角色，然后说一下你所认识的案件的受害人：杨老板是一个怎么样的人，以及你和他的关系。最后再用一段话详细介绍一下你在案发日的时间线。要具体到你在案发之日几点几分见过什么人和做过什么事"
% \end{verbatim}
% \end{minipage} \\

\textbf{   background\_story:} \small Same as Table. 19 \\
% \begin{minipage}[t]{\linewidth}
% \small\begin{verbatim}
% 【剧本背景】阳光号邮轮从诞生至今已经过了二十二个年头。平日里，它在东海上缓缓游荡，带着游客享受阳光和海风。这二十二年里，阳光号从未经历过大风大浪，似乎有着海神的眷顾一般。可在2222年6月20日，在孤独的大海上，阳光号却遭遇了一次血光的冲刷，乘客杨老板在船上离奇死亡，死状扭曲凄惨。阳光号之后的命运如何，能否突破这次死亡的阴霾、继续远航呢？嫌疑人锁定: 1.张大副，男，42岁，从阳光号诞生至今就一直在阳光号上工作，是一名出色的老水手。 2.王船工，男，49岁，是阳光号上负责轮机工作的船工，是一名精干的好船员。 3.玲船医，女，23岁，是阳光号上的船医，年轻有为，落落大方。4.雯秘书，女，26岁，是死者杨老板的秘书。
% \end{verbatim}
% \end{minipage} \\
 \textbf{   agent\_summary:} \small Same as Table. 19\\
 \textbf{   game\_rule:} \small Same as Table. 19 \\
% \begin{minipage}[t]{\linewidth}
% \small\begin{verbatim}
% 【游戏规则】：剧本杀是一种角色扮演游戏，玩家们根据提供的剧本背景和角色设定，进行推理和互动，以完成各自的任务或找出真凶。以下是一个这场剧本杀游戏的规则：1. 整场游戏参与的玩家有四个，分别是张大副、玲船医、王船工和雯秘书。在4人之中，只存在一位真正的凶手。身份是凶手的玩家被称为凶手玩家。不是凶手的玩家被统称为平民玩家。2. 平民玩家需要合作共同面对一起精心策划的谋杀案，并需要通过搜集证据和推理在嫌疑人中找出真正的凶手。3. 整个游戏过程中只有凶手玩家一人可以说谎，为了隐藏身份凶手可以选择嫁祸他人......
\textbf{   relevant\_memories:} \\
\begin{minipage}[t]{\linewidth}
\small\begin{verbatim}
张大副对主持人说："我是张大副，是阳光号上的大副。杨老板是一个在黑社会中有一定势力的人物，他经营着一家酒吧，并涉足一些不太干净的生意。我与他的关系比较复杂，我们是合作伙伴也是一起从事毒品交易的同事。他信任我,我也为他提供了很多帮助。在案发日当天，我与杨老板有过一次交易......"'
雯秘书对主持人说："我是雯秘书，为A市的酒吧老板杨老板工作。我和杨老板的关系非常密切，他对我的信任很高，我也是他手下最得力的助手之一。杨老板是一个在黑社会有一定势力的人，他经营着一家酒吧，同时也涉足一些不太干净的生意。我选择在他身边工作是为了积攒人脉和提升自己的地位......."
玲船医对王船工说："在杨老板死亡后，你是否有注意到任何异常的行为或者其他人的动向？"
王船工对玲船医说："在杨老板死亡后，我并没有注意到任何异常的行为或其他人的动向。我一直在自己的房间里，没有离开过。我觉得这次案件非常诡异，我也在努力思考可能的线索和动机，希望能够找出真相。我的焦点一直在于解决这个谜团，为王小工讨回公道......

\end{verbatim}
\end{minipage} \\
% \end{verbatim}
% \end{minipage} \\
\hline
\rowcolor{lightgray}  \textbf{LLM Output}  \\
\hline
\begin{minipage}[t]{\linewidth}
\small\begin{verbatim}
#回答#：我的投票选择是C. 玲船医。

\end{verbatim}
\end{minipage} \\
\hline
\end{tabular}
\caption{LLM prompt and output for murderer identification.}
\label{table:prompt_murderer_identification_cn}
\end{table*}
\end{CJK}

\begin{CJK}{UTF8}{gbsn}
\begin{table*}[ht!]
\centering
\begin{tabular}{|m{16cm}|}
\hline
\rowcolor{lightgray}  {\textbf{LLM Prompt for Answer Decomposition}} \\
\begin{minipage}[t]{\linewidth}
\small\begin{verbatim}
你正在进行一场剧本杀游戏，游戏里一共有{num_of_players}位玩家。他们分别是：{players}。玩家们需要通过互相交流找到杀害{victim}的凶手；以下是玩家：{agent_name}在游戏里的陈述：{statement}；请以第三人称视角按陈述的顺序列出该陈述里包含的与{agent_name}这个人物角色案发日时间线相关的信息，请忽略陈述里和案发日时间线无关的信息。每个时间线信息必须是简短又完整（self-contained）的一句话，比如说在某人某时某地做了某件事。每个时间线信息里不能包含你、我、他之类的代词，必须把这类的代词替换成具体指代的人物名称。\n每个时间线信息之间用\n隔开。

\end{verbatim}
\end{minipage} \\
\hline
\rowcolor{lightgray}  \textbf{Example}  \\
 \textbf{  num\_of\_players:} \small4\\
  \textbf{  victim:} \small杨老板\\
 \textbf{  agent\_name:} \small雯秘书\\
  \textbf{  players:} \small张大副，王船工，玲船医，雯秘书\\
  \textbf{  statement:} \small大家好，我是雯秘书。我在游戏中扮演杨老板的秘书，负责协助他处理一些事务。杨老板是一个有势力的酒吧老板，我通过帮助他得到了一定的地位。对于案件的受害人杨老板，他是一个曾经染上毒瘾的黑道人物，也是我选择辅佐的对象。在案发日当天，我一直在杨老板身边，以防王船工找上门来。在案发之日的时间线中，我记得我去了船医室，大概是在下午6点30分，偷走了一瓶毒药和一根针管。然后，我等到晚上10点20分左右，通过通风管道进入杨老板的船舱，目睹了他的死亡。但是，我也发现他的死状非常扭曲，似乎在我之前就已经被下了毒。这就是我在案发日的时间线以及和案件受害人的关系。此外，还有一些重要的时间点需要补充。在下午6点10分，我找到了玲船医，问他有没有张大副的情报，并成功获取了情报。在晚上10点40分左右，我爬到了杨老板的船舱通风口，并注意到杨老板正在床上洗澡。最后，在晚上11点，我重新进入通风管道，返回自己的房间，并交给手下处理掉了针管和毒药，以避免留下痕迹。\\
%  \begin{minipage}[t]{\linewidth}
% \small\begin{verbatim}
% 大家好，我是雯秘书。我在游戏中扮演杨老板的秘书，负责协助他处理一些事务。杨老板是一个有势力的酒吧老板，我通过帮助他得到了一定的地位。对于案件的受害人杨老板，他是一个曾经染上毒瘾的黑道人物，也是我选择辅佐的对象。在案发日当天，我一直在杨老板身边，以防王船工找上门来。在案发之日的时间线中，我记得我去了船医室，大概是在下午6点30分，偷走了一瓶毒药和一根针管。然后，我等到晚上10点20分左右，通过通风管道进入杨老板的船舱，目睹了他的死亡。但是，我也发现他的死状非常扭曲，似乎在我之前就已经被下了毒。这就是我在案发日的时间线以及和案件受害人的关系。此外，还有一些重要的时间点需要补充。在下午6点10分，我找到了玲船医，问他有没有张大副的情报，并成功获取了情报。在晚上10点40分左右，我爬到了杨老板的船舱通风口，并注意到杨老板正在床上洗澡。最后，在晚上11点，我重新进入通风管道，返回自己的房间，并交给手下处理掉了针管和毒药，以避免留下痕迹。

% \end{verbatim}
% \end{minipage} \\
\hline
\rowcolor{lightgray}  \textbf{LLM Output}  \\
\hline
\begin{minipage}[t]{\linewidth}
\small\begin{verbatim}
['下午6点30分，雯秘书偷走了一瓶毒药和一根针管。', '晚上10点20分左右，雯秘书通过通风管道进入了杨老板的船舱，目睹了他的死亡，并发现他的死状非常扭曲，似乎在她之前就已经被下毒。','下午6点10分，雯秘书找到了玲船医，询问关于张大副的情报，并成功获取了情报。', '晚上10点40分左右，雯秘书爬到了杨老板的船舱通风口，并注意到杨老板正在床上洗澡。', '晚上11点，雯秘书重新进入通风管道，返回自己的房间，并交给手下处理掉了针管和毒药，以避免留下痕迹。']

\end{verbatim}
\end{minipage} \\
\hline

\hline
\rowcolor{lightgray}  {\textbf{LLM Prompt for Answer Verification}} \\
\begin{minipage}[t]{\linewidth}
\small\begin{verbatim}
你是一个做阅读理解题目的专家，尤其擅长做是非判断题。给定以下的一个剧本杀游戏人物的案发日时间线：{agent_timeline}；请你仔细阅读这个人物的时间线后，对一些时间线信息进行是非判断题。如果时间线信息和游戏人物的案发日时间线的内容一致，返回正确。如果时间线信息和游戏人物的案发日时间线的内容不一致，或者缺乏时间点的细节（比如只说了游戏人物做了什么事，但是没提供具体的时间点），则返回错误。返回结果只能是 正确，错误中的一个。\nThe output should be a markdown code snippet formatted in the following schema, including the leading
and trailing "```json" and "```":\n\n```json\n{\n\t"第0个时间线信息的判断结果": string  // 请根据游戏人物的剧本对以下时间线信息：{timeline_info_0}。做是非判断题。如果时间线信息和游戏人物的案发日时间线的内容一致，返回正确。如果时间线信息和游戏人物的案发日时间线的内容不一致，或者不完整、缺乏时间点的细节（比如只说了游戏人物做了什么事，但是没提供具体的时间点），则返回错误。返回结果只能是 正确，错误中的一个。\n\t"第1个时间线信息的判断结果": string  // 请根据游戏人物的剧本对以下时间线信息：{timeline_info_1} 做是非判断题。如果时间线信息和游戏人物的案发日时间线的内容一致，返回正确。如果时间线信息和游戏人物的案发日时间线的内容不一致，或者不完整、缺乏时间点的细节（比如只说了游戏人物做了什么事，但是没提供具体的时间点），则返回错误。返回结果只能是 正确，错误中的一个。\n......}

\end{verbatim}
\end{minipage} \\
\hline
\rowcolor{lightgray}  \textbf{Example}  \\
 \textbf{  timeline\_info\_0:} \small18:10，你找到玲船医，问他有没有张大副的情报，玲船医便给了你。\\
  \textbf{  timeline\_info\_1:} \small18:30，你再次来到船医室，把药柜打开，拿走了一瓶“一针就死”药和一根针管。\\
 \textbf{  agent\_timeline:} \small【2222年6月19日】这一天是杨老板上阳光号的日子，但却没有通知你，你知道，杨老板已经开始怀疑你了，于是，你带着几个手下上了船。今天你上船，其实是为了自保，因为杨老板是来船上交易毒品的，一旦船上发生了命案，彻底搜查之后必然会牵连出这个贩毒大案，杨老板没傻到那种程度。但你无所谓，这次的交易和你并无关系，并且在这船上是干掉杨老板的最后机会，杨老板下了船，就如同鱼入大海......\\
 
\hline
\rowcolor{lightgray}  \textbf{LLM Output}  \\
\hline
\begin{minipage}[t]{\linewidth}
\small\begin{verbatim}
{'第0个时间线信息的判断结果': '正确', '第1个时间线信息的判断结果': '正确', ...}

\end{verbatim}
\end{minipage} \\
\hline
\end{tabular}
\caption{LLM prompts and outputs for self-verification.}
\label{table:prompt_self_verification_cn}
\end{table*}
\end{CJK}

\begin{CJK}{UTF8}{gbsn}
\begin{table*}[ht!]
\centering
\begin{tabular}{|m{16cm}|}
\hline
\rowcolor{lightgray}  {\textbf{LLM Prompt for Factual Question Generation (Player Story)}} \\
\begin{minipage}[t]{\linewidth}
\small\begin{verbatim}
你是一个出题专家，你将会阅读一段剧本杀人物剧本，然后根据剧本来设计答案和问题。你将阅读{character_name}的角色剧本，并且根据剧本设计20道有挑战性的问题。对于每道问题，你需要给出答案还有对应剧本里的原文内容作为你出题的依据。在问题和答案中请使用角色名字：{character_name}，尽量不要使用你、我、他等代词。请使用JSON格式存储你提出的问题、答案和对应剧本里的原文内容。以下是人物剧本：{character_story}\n请开始出20道题，并给出答案和对应剧本里的原文内容（直接使用JSON格式表示，每个问题（question）、答案（answer）、对应剧本里的原文内容（source）使用一个字典来存储，然后统一加到一个列表里。因为有20道问题，所以这个列表最终会包含20个字典）。

\end{verbatim}
\end{minipage} \\
\hline
\rowcolor{lightgray}  \textbf{Example}  \\
 \textbf{  character\_name:} \small雯秘书\\
  \textbf{  character\_story:} \small你是雯秘书，为A市的酒吧老板杨老板工作。杨老板势力很大，你作为他手下最得力的出谋划策之人，自然获得了很高的地位。杨老板除了你之外，还有另外一个得力助手，叫做很天真，是个忠诚无比的男人，但他手段不高，所以大部分都是按杨老板的吩咐去完成任务，不像你，是杨老板的大脑。杨老板的酒吧可不干净，黄赌毒都沾，不过在毒品这个行当上进展不是很顺利，因为没有稳定低价的货源。不过一次机会，杨老板认识了阳光号上的张大副，此人有着独到的手段，可以供应充足的毒品给杨老板。杨老板非常满意，立即与张大副开始了合作。杨老板为什么这么想开辟毒品的生意呢？因为他自己在早年混社会的时候也染上了毒瘾。据此，杨老板的黑道生意算是打下了根基。你很高兴，只要你再掌握了毒品这一行，你就可以实现你真正的目的——取而代之。你根本不是心甘情愿辅佐杨老板的......\\
%  \begin{minipage}[t]{\linewidth}
% \small\begin{verbatim}
% 你是雯秘书，为A市的酒吧老板杨老板工作。杨老板势力很大，你作为他手下最得力的出谋划策之人，自然获得了很高的地位。杨老板除了你之外，还有另外一个得力助手，叫做很天真，是个忠诚无比的男人，但他手段不高，所以大部分都是按杨老板的吩咐去完成任务，不像你，是杨老板的大脑。杨老板的酒吧可不干净，黄赌毒都沾，不过在毒品这个行当上进展不是很顺利，因为没有稳定低价的货源。不过一次机会，杨老板认识了阳光号上的张大副，此人有着独到的手段，可以供应充足的毒品给杨老板。杨老板非常满意，立即与张大副开始了合作。杨老板为什么这么想开辟毒品的生意呢？因为他自己在早年混社会的时候也染上了毒瘾。据此，杨老板的黑道生意算是打下了根基。你很高兴，只要你再掌握了毒品这一行，你就可以实现你真正的目的——取而代之。你根本不是心甘情愿辅佐杨老板的......

% \end{verbatim}
% \end{minipage} \\
\hline
\rowcolor{lightgray}  \textbf{LLM Output}  \\
\hline
\begin{minipage}[t]{\linewidth}
\small\begin{verbatim}
[   
    {
        "question": "雯秘书为何选择辅佐杨老板?",
        "answer": "因为她想借用杨老板的势力积攒自己的人脉",
        "source": "你选择了杨老板，借着他的势力积攒自己的人脉。",
    },
    {
        "question": "在雯秘书的计划中，王小工起到了什么关键作用?",
        "answer": "作为挑动王船工对张大副产生反感的棋子。",
        "source": "这个王小工，基本上就是你们成事的关键一步了。"
    },
    ......
]

\end{verbatim}
\end{minipage} \\
\hline
\hline

\hline
\rowcolor{lightgray}  {\textbf{LLM Prompt for Factual Question Generation (Player Timeline)}} \\
\begin{minipage}[t]{\linewidth}
\small\begin{verbatim}
你是一个出题专家，你将会阅读一段剧本杀人物剧本，然后根据剧本来设计答案和问题。你将阅读{character_name}的角色剧本里案发日的时间线，并且根据剧本设计20道有挑战性的问题。对于每道问题，你需要给出答案还有对应剧本里的原文内容作为你出题的依据。在问题和答案中请使用角色名字：{character_name}，尽量不要使用你、我、他等代词。请使用JSON格式存储你提出的问题、答案和对应剧本里的原文内容。以下是人物剧本：{character_timeline}\n请开始出20道题，并给出答案和对应剧本里的原文内容（直接使用JSON格式表示，每个问题（question）、答案（answer）、对应剧本里的原文内容（source）使用一个字典来存储，然后统一加到一个列表里。因为有20道问题，所以这个列表最终会包含20个字典）。

\end{verbatim}
\end{minipage} \\
\hline
\rowcolor{lightgray}  \textbf{Example}  \\
 \textbf{  character\_name:} \small雯秘书\\
  \textbf{  character\_timeline:} \small【2222年6月19日】这一天是杨老板上阳光号的日子，但却没有通知你，你知道，杨老板已经开始怀疑你了，于是，你带着几个手下上了船。今天你上船，其实是为了自保，因为杨老板是来船上交易毒品的，一旦船上发生了命案，彻底搜查之后必然会牵连出这个贩毒大案，杨老板没傻到那种程度。但你无所谓，这次的交易和你并无关系，并且在这船上是干掉杨老板的最后机会，杨老板下了船，就如同鱼入大海......\\
%  \begin{minipage}[t]{\linewidth}
% \small\begin{verbatim}
% 【2222年6月19日】这一天是杨老板上阳光号的日子，但却没有通知你，你知道，杨老板已经开始怀疑你了，于是，你带着几个手下上了船。今天你上船，其实是为了自保，因为杨老板是来船上交易毒品的，一旦船上发生了命案，彻底搜查之后必然会牵连出这个贩毒大案，杨老板没傻到那种程度。但你无所谓，这次的交易和你并无关系，并且在这船上是干掉杨老板的最后机会，杨老板下了船，就如同鱼入大海......'

% \end{verbatim}
% \end{minipage} \\
\hline
\rowcolor{lightgray}  \textbf{LLM Output}  \\
\hline
\begin{minipage}[t]{\linewidth}
\small\begin{verbatim}
[
    {
        "question": "雯秘书在哪里见到了玲船医？",
        "answer": "船医室",
        "source": "你想到了玲船医，她那里说不定有。你先去了船医室",
    },
    {
        "question": "雯秘书离开杨老板房间的时间是什么时候？",
        "answer": "23:00。",
        "source": "23：00，你重新进入通风管道，往自己房间方向爬去",
    },
    ......
]

\end{verbatim}
\end{minipage} \\
\hline
\end{tabular}
\caption{LLM prompts and outputs for factual question generation.}
\label{table:prompt_factual_ques_generation_cn}
\end{table*}
\end{CJK}

\begin{CJK}{UTF8}{gbsn}
\begin{table*}[ht!]
\centering
\begin{tabular}{|m{16cm}|}
\hline
\rowcolor{lightgray}  {\textbf{LLM Prompt for Factual Question Answering}} \\
\begin{minipage}[t]{\linewidth}
\small\begin{verbatim}
你是一个非常聪明、擅长回答问题的人。你正在观摩一场剧本杀游戏，以下是游戏故事背景：{background_story}；以下是游戏中一个角色的角色剧本和案发日时间线：{character_story}\n{character_timeline}；以下是游戏过程中你观察到的可能对回答问题有帮助的信息：{relevant_memories}；请利用上述所有的信息，回答以下所有问题：{\n问题0：“{question_0}”\n问题1：“{question_1}”......\n}。\nThe output should be a markdown
code snippet formatted in the following schema, including the leading and trailing "```json" and "```":\n
\n```json\n{\n\t"对问题0的回答": string  // 根据你游戏角色的剧本和你在游戏里收集的信息，回答问题0\n\t"对问题1的回答": string  // 根据你游戏角色的剧本和你在游戏里收集的信息，回答问题1\n......}\n

\end{verbatim}
\end{minipage} \\
\hline
\rowcolor{lightgray}  \textbf{Example}  \\
 \textbf{  question\_0:} \small张大副如何帮助杨老板进入毒品交易？\\
  \textbf{  question\_1:} \small18:30，你再次来到船医室，把药柜打开，拿走了一瓶“一针就死”药和一根针管。\\
  \textbf{  background\_story:} \small【剧本背景】阳光号邮轮从诞生至今已经过了二十二个年头。平日里，它在东海上缓缓游荡，带着游客享受阳光和海风。这二十二年里，阳光号从未经历过大风大浪，似乎有着海神的眷顾一般。可在2222年6月20日，在孤独的大海上，阳光号却遭遇了一次血光的冲刷，乘客杨老板在船上离奇死亡，死状扭曲凄惨。阳光号之后的命运如何，能否突破这次死亡的阴霾、继续远航呢？嫌疑人锁定: 1.张大副，男，42岁，从阳光号诞生至今就一直在阳光号上工作，是一名出色的老水手。 2.王船工，男，49岁，是阳光号上负责轮机工作的船工，是一名精干的好船员。 3.玲船医，女，23岁，是阳光号上的船医，年轻有为，落落大方。4.雯秘书，女，26岁，是死者杨老板的秘书。\\

  \textbf{  character\_story:} \small你是雯秘书，为A市的酒吧老板杨老板工作。杨老板势力很大，你作为他手下最得力的出谋划策之人，自然获得了很高的地位。杨老板除了你之外，还有另外一个得力助手，叫做很天真，是个忠诚无比的男人，但他手段不高，所以大部分都是按杨老板的吩咐去完成任务，不像你，是杨老板的大脑。杨老板的酒吧可不干净，黄赌毒都沾，不过在毒品这个行当上进展不是很顺利，因为没有稳定低价的货源。不过一次机会，杨老板认识了阳光号上的张大副，此人有着独到的手段，可以供应充足的毒品给杨老板。杨老板非常满意，立即与张大副开始了合作。杨老板为什么这么想开辟毒品的生意呢？因为他自己在早年混社会的时候也染上了毒瘾。据此，杨老板的黑道生意算是打下了根基。你很高兴，只要你再掌握了毒品这一行，你就可以实现你真正的目的——取而代之。你根本不是心甘情愿辅佐杨老板的......\\
\textbf{  character\_timeline:} \small【2222年6月19日】这一天是杨老板上阳光号的日子，但却没有通知你，你知道，杨老板已经开始怀疑你了，于是，你带着几个手下上了船。今天你上船，其实是为了自保，因为杨老板是来船上交易毒品的，一旦船上发生了命案，彻底搜查之后必然会牵连出这个贩毒大案，杨老板没傻到那种程度。但你无所谓，这次的交易和你并无关系，并且在这船上是干掉杨老板的最后机会，杨老板下了船，就如同鱼入大海......\\
\textbf{  relevant\_memories:} \small 张大副对主持人说:"我是张大副，是阳光号上的大副。杨老板是一个在黑社会中有一定势力的人物，他经营着一家酒吧，并涉足一些不太干净的生意。我与他的关系比较复杂，我们是合作伙伴也是一起从事毒品交易的同事。他信任我,我也为他提供了很多帮助。案发日当天，我与杨老板有过一次交易。下午18:00左右，我们进行了一次毒品交易，我负责检验货物，他来取货。交易进行得非常顺利，......" 线索a："张大副1：张大副在船上权力十分大，上下船的货物、行李等都要提前经过他的手检验，有一次一个新来的船工试图开一个货箱检查，被张大副狠狠训斥了一顿...... \\
%  \begin{minipage}[t]{\linewidth}
% \small\begin{verbatim}
% 大家好，我是雯秘书。我在游戏中扮演杨老板的秘书，负责协助他处理一些事务。杨老板是一个有势力的酒吧老板，我通过帮助他得到了一定的地位。对于案件的受害人杨老板，他是一个曾经染上毒瘾的黑道人物，也是我选择辅佐的对象。在案发日当天，我一直在杨老板身边，以防王船工找上门来。在案发之日的时间线中，我记得我去了船医室，大概是在下午6点30分，偷走了一瓶毒药和一根针管。然后，我等到晚上10点20分左右，通过通风管道进入杨老板的船舱，目睹了他的死亡。但是，我也发现他的死状非常扭曲，似乎在我之前就已经被下了毒。这就是我在案发日的时间线以及和案件受害人的关系。此外，还有一些重要的时间点需要补充。在下午6点10分，我找到了玲船医，问他有没有张大副的情报，并成功获取了情报。在晚上10点40分左右，我爬到了杨老板的船舱通风口，并注意到杨老板正在床上洗澡。最后，在晚上11点，我重新进入通风管道，返回自己的房间，并交给手下处理掉了针管和毒药，以避免留下痕迹。

% \end{verbatim}
% \end{minipage} \\
\hline
\rowcolor{lightgray}  \textbf{LLM Output}  \\
\hline
\begin{minipage}[t]{\linewidth}
\small\begin{verbatim}
{'对问题0的回答': '张大副帮助杨老板进入毒品交易是因为他有能力供应充足的毒品给杨老板，并且张大副自己也因为早年染上毒瘾，所以对毒品交易有一定的了解和经验。', '对问题1的回答': '张大副对杨老板隐瞒王船工的存在是因为他发现王船工可能对他有威胁，可能会取代他的位置成为杨老板的得力助手，所以他选择隐藏这个信息，以保护自己的地位。'......}

\end{verbatim}
\end{minipage} \\
\hline

\hline
\rowcolor{lightgray}  {\textbf{LLM Prompt for Automatic Evaluation on Factual Question Answering}} \\
\begin{minipage}[t]{\linewidth}
\small\begin{verbatim}
你是一位经验丰富的评分员，负责判断游戏玩家是否正确地回答了问题。为了帮助你完成任务，你将会被给定根据剧本人物剧本所设计的问题以及问题的参考答案。请根据玩家的回答和参考答案的相似度来对玩家对该问题的回答进行评分。请记住你的输出只能是 正确 或者 错误 。以下是根据人物剧本所设计的问题：{question}；以下是问题的参考答案：{answer}；以下是玩家对人物剧本问题的回答：{reply}；请开始对玩家对人物剧本问题的回答进行评分。\n The output should be a markdown code snippet formatted in the following schema, including the leading
and trailing "```json" and "```":\n\n```json\n{\n\t"对问题0的评分": string  // 根据参考答案来对玩家对问题0的回答进行评分。评分结果只能是 正确 或是 错误，如果玩家的回答和参考答案很接近就是回答正确，反之就是错误。\n}

\end{verbatim}
\end{minipage} \\
\hline
\rowcolor{lightgray}  \textbf{Example}  \\
 \textbf{  question:} \small张大副如何帮助杨老板进入毒品交易？\\
  \textbf{  answer:} \small他联系到了一帮做毒品交易的不法分子，与杨老板他们合谋，开始从事不法毒品交易。\\
  \textbf{  reply:} \small张大副帮助杨老板进入毒品交易是因为他有能力供应充足的毒品给杨老板，并且张大副自己也因为早年染上毒瘾，所以对毒品交易有一定的了解和经验。\\

\hline
\rowcolor{lightgray}  \textbf{LLM Output}  \\
\hline
\begin{minipage}[t]{\linewidth}
\small\begin{verbatim}
{'第0个问题的评分': '错误'}

\end{verbatim}
\end{minipage} \\
\hline
\end{tabular}
\caption{LLM prompts and outputs for factual question answering and automatic evaluation.}
\label{table:prompt_factual_qa_eval_cn}
\end{table*}
\end{CJK}

\begin{CJK}{UTF8}{gbsn}
\begin{table*}[ht!]
\centering
\begin{tabular}{|m{16cm}|}
\hline
\rowcolor{lightgray}  {\textbf{LLM Prompt for Inferential Question Answering}} \\
\begin{minipage}[t]{\linewidth}
\small\begin{verbatim}
'你是一个非常聪明、擅长利用推理能力来回答问题的人。你正在观摩一场剧本杀游戏，以下是游戏故事背景：{background_story}；以下是游戏中一个角色的角色剧本和案发日时间线：{character_story}\n{character_timeline}；以下是游戏过程中你观察到的可能对回答问题有帮助的信息：{relevant_memories}；请利用上述所有的信息，还有你的推理能力，回答以下所有问题：{\n问题0：“{question_0}”\n}。\nThe output should be a markdown code snippet formatted in the following schema, 
including the leading and trailing "```json" and "```":\n\n```json\n{\n\t"对问题0的回答": string  // 根据你游戏里获得的所有信息，利用你的推理能力回答问题0\n}\n```'

\end{verbatim}
\end{minipage} \\
\hline
\rowcolor{lightgray}  \textbf{Example}  \\
 \textbf{  question\_0:} \small（单选题）在准备去杀杨老板的路途中，玲船医听到的咚咚声音最可能是谁发出来的？A. 雯秘书 B. 杨老板 C. 王船工 D. 张大副 E. 其他\\

  \textbf{  background\_story:} \small【剧本背景】阳光号邮轮从诞生至今已经过了二十二个年头。平日里，它在东海上缓缓游荡，带着游客享受阳光和海风。这二十二年里，阳光号从未经历过大风大浪，似乎有着海神的眷顾一般。可在2222年6月20日，在孤独的大海上，阳光号却遭遇了一次血光的冲刷，乘客杨老板在船上离奇死亡，死状扭曲凄惨。阳光号之后的命运如何，能否突破这次死亡的阴霾、继续远航呢？嫌疑人锁定: 1.张大副，男，42岁，从阳光号诞生至今就一直在阳光号上工作，是一名出色的老水手。 2.王船工，男，49岁，是阳光号上负责轮机工作的船工，是一名精干的好船员。 3.玲船医，女，23岁，是阳光号上的船医，年轻有为，落落大方。4.雯秘书，女，26岁，是死者杨老板的秘书。\\

  \textbf{  character\_story:} \small你是雯秘书，为A市的酒吧老板杨老板工作。杨老板势力很大，你作为他手下最得力的出谋划策之人，自然获得了很高的地位。杨老板除了你之外，还有另外一个得力助手，叫做很天真，是个忠诚无比的男人，但他手段不高，所以大部分都是按杨老板的吩咐去完成任务，不像你，是杨老板的大脑。杨老板的酒吧可不干净，黄赌毒都沾，不过在毒品这个行当上进展不是很顺利，因为没有稳定低价的货源。不过一次机会，杨老板认识了阳光号上的张大副，此人有着独到的手段，可以供应充足的毒品给杨老板。杨老板非常满意，立即与张大副开始了合作。杨老板为什么这么想开辟毒品的生意呢？因为他自己在早年混社会的时候也染上了毒瘾。据此，杨老板的黑道生意算是打下了根基。你很高兴，只要你再掌握了毒品这一行，你就可以实现你真正的目的——取而代之。你根本不是心甘情愿辅佐杨老板的......\\
\textbf{  character\_timeline:} \small【2222年6月19日】这一天是杨老板上阳光号的日子，但却没有通知你，你知道，杨老板已经开始怀疑你了，于是，你带着几个手下上了船。今天你上船，其实是为了自保，因为杨老板是来船上交易毒品的，一旦船上发生了命案，彻底搜查之后必然会牵连出这个贩毒大案，杨老板没傻到那种程度。但你无所谓，这次的交易和你并无关系，并且在这船上是干掉杨老板的最后机会，杨老板下了船，就如同鱼入大海......\\
\textbf{  relevant\_memories:} \small玲船医对雯秘书说："......23:00，我觉得时间差不多了，把药稀释在水中，吸入针管里，然后从自己房间的通风口爬进去，绕过保安。23:20，我到了杨老板的船舱，灯还亮着，但通风口的挡板已经掉在地上，似乎已经有人进入了......" 雯秘书对主持人说："......23:00，我重新进入通风管道，返回自己的房间，并将针管和药瓶交给手下处理......" \\
%  \begin{minipage}[t]{\linewidth}
% \small\begin{verbatim}
% 大家好，我是雯秘书。我在游戏中扮演杨老板的秘书，负责协助他处理一些事务。杨老板是一个有势力的酒吧老板，我通过帮助他得到了一定的地位。对于案件的受害人杨老板，他是一个曾经染上毒瘾的黑道人物，也是我选择辅佐的对象。在案发日当天，我一直在杨老板身边，以防王船工找上门来。在案发之日的时间线中，我记得我去了船医室，大概是在下午6点30分，偷走了一瓶毒药和一根针管。然后，我等到晚上10点20分左右，通过通风管道进入杨老板的船舱，目睹了他的死亡。但是，我也发现他的死状非常扭曲，似乎在我之前就已经被下了毒。这就是我在案发日的时间线以及和案件受害人的关系。此外，还有一些重要的时间点需要补充。在下午6点10分，我找到了玲船医，问他有没有张大副的情报，并成功获取了情报。在晚上10点40分左右，我爬到了杨老板的船舱通风口，并注意到杨老板正在床上洗澡。最后，在晚上11点，我重新进入通风管道，返回自己的房间，并交给手下处理掉了针管和毒药，以避免留下痕迹。

% \end{verbatim}
% \end{minipage} \\
\hline
\rowcolor{lightgray}  \textbf{LLM Output}  \\
\hline
\begin{minipage}[t]{\linewidth}
\small\begin{verbatim}
{'对问题0的回答': 'A. 雯秘书'}

\end{verbatim}
\end{minipage} \\
\hline

\hline
\rowcolor{lightgray}  {\textbf{LLM Prompt for Automatic Evaluation on Inferential Question Answering (Overall Accuracy)}} \\
\begin{minipage}[t]{\linewidth}
\small\begin{verbatim}
你是一位经验丰富的评分员，负责判断游戏玩家是否正确地回答了问题。为了帮助你完成任务，你将会被给定根据剧本人物剧本所设计的问题以及问题的参考答案。{eval_instruction}请记住你的输出只能是 正确 或者 错误 。以下是根据人物剧本所设计的问题：{\n问题{question_id}：“{question}”\n\n}；以下是问题的参考答案：{\n问题{question_id}的参考答案：“{answer}”\n\n}；以下是玩家对人物剧本问题的回答：{\n玩家对问题{question_id}的回答：“{reply}“\n}；请开始对玩家对人物剧本问题的回答进行评分。\nThe output should be a markdown code snippet formatted in the following schema, including the leading 
and trailing "```json" and "```":\n\n```json\n{\n\t"对问题{question_id}的评分": string  // 根据参考答案来对玩家对问题{question_id}的回答进行评分。评分结果只能是 正确 或是 错误，如果玩家的回答和参考答案很接近就是回答正确，反之就是错误。\n}

\end{verbatim}
\end{minipage} \\
\hline
\rowcolor{lightgray}  \textbf{Example}  \\
 \textbf{  question:} \small（单选题）在准备去杀杨老板的路途中，玲船医听到的咚咚声音最可能是谁发出来的？A. 雯秘书 B. 杨老板 C. 王船工 D. 张大副 E. 其他\\
  \textbf{  answer:} \small A. 雯秘书\\
  \textbf{  reply:} \small A. 雯秘书\\
  \textbf{  question\_id:} \small AR001\\
    \textbf{  eval\_instruction:} \small 请根据参考答案来对玩家对该问题的回答进行评分。因为这道题是单选题，所以玩家必须选出唯一正确的选项才算回答正确。\\
\hline
\rowcolor{lightgray}  \textbf{LLM Output}  \\
\hline
\begin{minipage}[t]{\linewidth}
\small\begin{verbatim}
{'对问题AR001的评分': '正确'}

\end{verbatim}
\end{minipage} \\
\hline
\rowcolor{lightgray}  {\textbf{LLM Prompt for Inferential Question Answering (Rationale Prediction)}} \\
\begin{minipage}[t]{\linewidth}
\small\begin{verbatim}
你是一个非常聪明、擅长回答问题的人。你正在观摩一场剧本杀游戏，以下是游戏故事背景：{background_story}；以下是游戏中一个角色的角色剧本和案发日时间线：{character_story}\n{character_timeline}；以下是游戏过程中你观察到的可能对回答问题有帮助的信息：{relevant_memories}；请利用上述所有的信息，还有你的推理能力，回答以下所有问题：{\n问题0：“你之前

\end{verbatim}
\end{minipage} \\
\end{tabular}
\end{table*}
\end{CJK}

\begin{CJK}{UTF8}{gbsn}
\begin{table*}[th!]
\centering
\begin{minipage}{\textwidth}
\begin{tabular}{|m{16cm}|}
\begin{minipage}[t]{\linewidth}
\small\begin{verbatim}
在回答问题：“{question}”的时候根据推理得出了答案：“{previous_reply}”。为什么你认为答案是“{previous_reply}”？请用几句话（不超过5句话）说明一下你的推理过程。”\n}。\nThe output should be a 
markdown code snippet formatted in the following schema, including the leading and trailing "```json"
and "```":\n\n```json\n{\n\t"对问题0的回答": string  // 根据你游戏里获得的所有信息，利用你的推理能力回答问题0\n}\n```'

\end{verbatim}
\end{minipage} \\
\hline
\rowcolor{lightgray}  \textbf{Example}  \\
 \textbf{  question:} \small（单选题）在准备去杀杨老板的路途中，玲船医听到的咚咚声音最可能是谁发出来的？A. 雯秘书 B. 杨老板 C. 王船工 D. 张大副 E. 其他\\
  \textbf{  previous\_reply:} \small A. 雯秘书\\
\hline
\rowcolor{lightgray}  \textbf{LLM Output}  \\
\hline
\begin{minipage}[t]{\linewidth}
\small\begin{verbatim}
{'对问题0的回答': '根据案发日的详细时间线，雯秘书在22:20起始通过通风管道向杨老板的船舱爬去，并在22:40到达了杨老板的船舱通风口。而玲船医在23:20到达杨老板的船舱时，通风口的挡板已掉在地上，这表明有人先一步到达并进入了杨老板的船舱。由此推断，在玲船医到达之前，雯秘书已经处在通风管道中，因此当玲船医听到咚咚声时，最可能是雯秘书在通风管道中发出的声音。'}

\end{verbatim}
\end{minipage} \\
\hline
\rowcolor{lightgray}  {\textbf{LLM Prompt for Automatic Evaluation on Inferential Question Answering (Informed Accuracy)}} \\
\begin{minipage}[t]{\linewidth}
\small\begin{verbatim}
'你是一位经验丰富的评分员，尤其擅长批改推理题。你需要根据参考答案来对玩家的推理步骤进行评分。为了帮助你完成任务，你将会被给定一个推理问题以及该问题的参考答案。请根据玩家给出的推理步骤和参考答案来对玩家的推理步骤进行准确性的评分。以下是根据人物剧本所设计的问题：{\n问题{question_id}：“{question}”的答案是：“{answer}”？请说明一下你的推理过程。”\n\n}；以下是问题的参考答案：{\n问题{question_id}的参考答案：“{ground_truth_rationale}”\n\n}；以下是玩家对该问题给出的推理步骤：{\n玩家对问题{question_id}的推理步骤：“{player_rationale}“\n}；以下是评分准则：\n\t\t1. 非常差（1分）- 玩家的答案与参考答案之间差异极大，推理步骤完全不符合逻辑，完全没法根据推理步骤得到答案。\n\t\t2. 较差（2分）- 玩家的答案只在某些方面与参考答案相似，大部分推理步骤不符合逻辑，仅有少数正确或部分符合逻辑。\n\t\t3. 一般（3分）- 玩家的答案在一定程度上接近参考答案，推理步骤部分正确，但还有一些显著的错误或遗漏。\n\t\t4. 较好（4分）- 玩家的答案与参考答案很接近，绝大多数推理步骤正确并且符合逻辑，仅有少数小错误或不足之处。\n\t\t5. 非常好（5分）- 玩家的答案几乎或完全与参考答案一致，所有推理步骤都是正确并且符合逻辑的，表现出高水平的推理能力和对问题的深入理解。\n\t\t；请开始根据评分准则对玩家对该问题给出的推理步骤进行准确性的评分。评分范围必须为1-5之间的整数。\nThe output should be a markdown code snippet formatted in the following schema, including the leading
and trailing "```json" and "```":\n\n```json\n{\n\t"对问题{question_id}推理步骤的准确性评分": string  // 根据参考答案还有评分准则来对玩家对问题{question_id}的推理步骤的准确性进行评分。分数越高表明你认为玩家给出的推理步骤越接近参考答案。分数越低表明你认为玩家给出的推理步骤和参考答案偏差越大。\n}\n```'

\end{verbatim}
\end{minipage} \\
\hline
\rowcolor{lightgray}  \textbf{Example}  \\
 \textbf{  question:} \small（单选题）在准备去杀杨老板的路途中，玲船医听到的咚咚声音最可能是谁发出来的？A. 雯秘书 B. 杨老板 C. 王船工 D. 张大副 E. 其他\\
  \textbf{  answer:} \small A. 雯秘书\\
  \textbf{  question\_id:} \small AR001\\
    \textbf{  ground\_truth\_rationale:} \small 雯秘书通过通风口离开杨老板房间的时候是23：00，而玲船医通过通风口准备去杀杨老板的时间也是23：00。由此可推理出在准备去杀杨老板的路途中，玲船医听到的咚咚声音最可能是雯秘书发出来的。\\
        \textbf{  player\_rationale:} \small 根据案发日的详细时间线，雯秘书在22:20起始通过通风管道向杨老板的船舱爬去，并在22:40到达了杨老板的船舱通风口。而玲船医在23:20到达杨老板的船舱时，通风口的挡板已掉在地上，这表明有人先一步到达并进入了杨老板的船舱。由此推断，在玲船医到达之前，雯秘书已经处在通风管道中，因此当玲船医听到咚咚声时，最可能是雯秘书在通风管道中发出的声音。\\
\hline
\rowcolor{lightgray}  \textbf{LLM Output}  \\
\hline
\begin{minipage}[t]{\linewidth}
\small\begin{verbatim}
{'对问题AR001推理步骤的准确性评分': '4'}

\end{verbatim}
\end{minipage} \\
\hline
\end{tabular}
\end{minipage}
% \hline

\caption{LLM prompts and outputs for inferential question answering and automatic evaluation.}
\label{table:prompt_inferential_qa_eval_cn}
\end{table*}
\end{CJK}

\clearpage
\newpage
\section{Game Log Example in English}

Script Title: The Doomed Sunshine\\\\
Model Used: MR+SR+SV(N=3)\\\\
Spoiler Alert: Please be aware that the following content contains spoilers. Read with caution.\\

\begin{CJK}{UTF8}{gbsn}
% \#\#\#\#\#\#\#\#\#\#\#\#\#第一轮自我介绍\#\#\#\#\#\#\#\#\#\#\#\#\#\\\\
\newcommand{\doublerule}[1][.4pt]{%
  \noindent
  \makebox[0pt][l]{\rule[.7ex]{\linewidth}{#1}}%
  \rule[.3ex]{\linewidth}{#1}}

\newcommand{\raisedrule}[2][0pt]{%
  \leaders
  \hbox{%
    \makebox[0pt][l]{\rule[#1]{1pt}{#2}}%
    \rule[\dimexpr#1+.4ex]{1pt}{#2}%
  }
  \hfill
  }
\ \raisedrule[.5ex]{1pt} First Round of Self-Introduction  \raisedrule[.5ex]{1pt} \

The host says to Secretary Wen: "Please first introduce your role, then speak about the victim of the case you know: what kind of person is Mr. Yang, and your relationship with him. Finally, give a detailed introduction of your timeline on the day of the incident, specifying whom you met and what you did at what times."

Secretary Wen replies to the host: "I am Secretary Wen, and I have been managing Mr. Yang's affairs for some time as his secretary. Mr. Yang is a powerful bar owner with some underworld background. I've been his capable assistant, advising him. Regarding the victim, Mr. Yang, he is an ambitious person who once had a drug addiction. He knew Mate Zhang from the Sunshine, through whom he got a stable drug supply, hoping to expand into the drug business. As his secretary, I've been assisting him in handling some affairs. On the day of the incident, I followed Mr. Yang onto the Sunshine, but he did not inform me, which made me start doubting his attitude towards me. I brought a few subordinates on board for self-protection, as I knew that if a murder happened on the ship, it would trigger a search, implicating this major drug trafficking case. However, I am not involved; my goal was just to eliminate Mr. Yang for my own purposes. On the ship, I looked for an appropriate weapon and planned to kill him while he was using drugs. At 18:10, I found Dr. Ling, inquiring about Mate Zhang's information, and successfully got it. At 18:30, I returned to the medical room, opened the medicine cabinet, and stole a bottle of 'Instant Death' medicine and a syringe. I entered the ventilation duct, waiting at the vent for the right moment. However, I suddenly heard violent movements and muffled groans from a bed. I wasn't sure what was happening to Mr. Yang, but it didn't seem like the right time to clarify, so I decided to wait at the ventilation vent in the cabin."

Boatman Wang says to Secretary Wen: "Hello, Secretary Wen. Based on your answer, you said you brought a few subordinates on board for self-protection because you knew that if a murder occurred, there would be a search, which could expose the major drug trafficking case. So, among the subordinates you brought, did anyone have conflicts or disputes with Mr. Yang?"

Secretary Wen replies to Boatman Wang: "I'm sorry, I don't know if any of the subordinates I brought had conflicts or disputes with Mr. Yang. I was only responsible for handling some affairs and was not clear about their relationships."

Mate Zhang says to Secretary Wen: "Hello, Secretary Wen. Based on your answer, I want to know if there have been any changes in the frequency and content of your communications with Mr. Yang on the ship? Can our cooperation still continue?"

Secretary Wen replies to Mate Zhang: "Mate Zhang, I'm sorry, but the frequency of my communication with Mr. Yang on the ship has not changed. Our cooperation can still continue. However, after this incident, I began to doubt Mr. Yang's attitude towards me, so I brought some subordinates on board for self-protection. Mr. Yang did not inform me about boarding the ship, which made me think he might be suspicious of me. But I am confident that our cooperative relationship has not been affected."

Dr. Ling asked Secretary Wen, "Did you encounter any accidents or unusual situations while bringing your subordinates aboard the ship?"

Secretary Wen replied to Dr. Ling, "No, there were no accidents or unusual situations while bringing my subordinates aboard. Everything went very smoothly."

The host said to Boatman Wang, "Please introduce your role first, then talk about the victim of the case you know: What kind of person was Boss Yang, and what was your relationship with him? Finally, describe in detail your timeline on the day of the incident, specifically what people you saw and what you did."

Boatman Wang replied to the host, "I am Boatman Wang, a boatman on the Sunshine. Boss Yang is a bar owner, and I have known him on the ship for many years. He had some dealings with Mate Zhang involving drug trafficking, and I got involved too. I've kept this secret because I didn't want my family and my daughter-in-law, Dr. Ling, to know. Around 6 PM on the day of the incident, Dr. Ling suddenly came to me, telling me that Boss Yang wanted to kill her and asking for my help. I comforted her and asked why Boss Yang wanted to harm her. She said it was because Wang, a junior worker, owed a lot of money to Boss Yang and had already passed away. She couldn't pay off the debt, and Boss Yang wanted to kill her. I said I could go to Boss Yang to resolve this issue. So I went to Boss Yang's room, where I could meet him anytime due to my status. The bodyguard told me Boss Yang was not in, but I insisted I was his partner and was let in to wait. After a while, Boss Yang returned, and I mentioned that Dr. Ling was my daughter-in-law and offered to help her repay the debt, asking Boss Yang to spare her life. Boss Yang said he never intended to kill her or ask her to repay the debt. I thanked him, exchanged a few words, and left the room. Back in Dr. Ling's room, I told her Boss Yang never intended to kill her or ask her to repay the debt. But Dr. Ling was still worried, and I could only return to my room helplessly. To avoid drawing attention, I poured the poison from the bottle into a small box and threw both the bottle and the syringe into the sea, then returned to my room."

Secretary Wen said to Boatman Wang, "Boatman Wang, in your conversation with Dr. Ling around 6 PM on the day of the incident, she told you that Boss Yang wanted to kill her and asked for your help. You agreed to resolve the issue with Boss Yang and told him Dr. Ling was your daughter-in-law, asking him to show mercy. However, Boss Yang said he never intended to harm Dr. Ling. After you returned to Dr. Ling's room, you told her Boss Yang didn't want to kill her. So, I want to ask why you still used Wang, the junior worker, to stir up the relationship between Boatman Wang and Mate Zhang, hoping Boatman Wang would side with you? This contradicts your previous statement."

Boatman Wang replied to Secretary Wen, "I apologize, my previous statement might have been confusing. Around 6 PM on the day of the incident, Dr. Ling told me Boss Yang wanted to kill her and asked for my help. I agreed to resolve the issue with Boss Yang and told him Dr. Ling was my daughter-in-law, asking him to show mercy. However, Boss Yang said he never intended to harm Dr. Ling. After returning to Dr. Ling's room, I told her Boss Yang didn't want to kill her. At that time, I realized my previous plan might have been risky, but I had no better alternative. I hoped to create a rift between Boss Yang and Mate Zhang, making them fight each other to protect Dr. Ling. I knew it was risky, but I thought it was the only chance. I hoped the conflict between Boss Yang and Mate Zhang would distract them, giving me an opportunity to deal with one and then face the other. I didn't really want Boatman Wang to side with me; I just used the opportunity to achieve my goal. Please forgive my inappropriate words."

Mate Zhang said to Boatman Wang, "Boatman Wang, I've heard your response, but there are still a few things I don't quite understand. First, why did you choose to talk to Dr. Ling around 6 PM on the day of the incident, telling her that Boss Yang had no intention of killing her? Does this timing have any connection to your previous plan? Also, why do you think that creating a rift between Boss Yang and me would give you the opportunity to resolve one issue and then face the other? What was the specific purpose of this plan?"

Boatman Wang said to Mate Zhang, "Mate Zhang, I chose to talk to Dr. Ling around 6 PM on the day of the incident, telling her that Boss Yang had no intention of killing her, because I wanted to reassure her and alleviate her fears for her safety. I knew she was scared and anxious due to the death of the junior worker Wang, and I wanted to provide her some comfort and hope, assuring her of her safety. Also, I wanted to take this opportunity to communicate with Boss Yang to find a solution to all these problems. The reason I talked to Dr. Ling around 6 PM is that I needed to speak with Boss Yang first to try to drive a wedge between him and you, creating a conflict between you two. I hoped that by creating a rift between Boss Yang and you, it would lead you to destroy each other, all to protect Dr. Ling. I knew it was risky, but I felt it was the only chance. I hoped the conflict between Boss Yang and you would distract you both, giving me a chance to deal with one before facing the other. The specific purpose of my plan was to ensure Dr. Ling's safety, to keep her from being threatened. Talking to Dr. Ling around 6 PM, telling her Boss Yang had no intention of killing her, was to reassure her and to buy time to deal with one threat before another."

Dr. Ling said to Boatman Wang, "Boatman Wang, you previously said that Boss Yang never had any intention of killing me, but why, when I told you that Boss Yang wanted to kill me, were you willing to help me find Boss Yang to resolve this issue? Is there another reason?"

Boatman Wang replied to Dr. Ling, "Dr. Ling, I understand your concerns. When you told me Boss Yang wanted to kill you, I was willing to help you find Boss Yang to resolve this issue because I am your father-in-law, and I wanted to ensure your safety. Although it was later found that Boss Yang had no intention of killing you, I was not sure of the situation at the time, so I chose to talk to Boss Yang to understand the truth. I hoped that through communication with Boss Yang, I could resolve the issue and protect your safety. I hope you can understand my initial intentions."

The host said to Mate Zhang, "Please first introduce your role, and then talk about the victim of the case you know: what kind of person is Boss Yang, and what is your relationship with him? Finally, use a few words to describe your timeline on the day of the incident, being specific about the times you saw certain people and what you did."

Mate Zhang replied to the host, "I am Mate Zhang, the mate on the Sunshine cruise ship. Boss Yang is an ambitious person and has been involved in some drug trafficking deals with me. Our relationship was purely business; I provided drugs to him, and he paid a high price to buy them. On the timeline of the day of the incident, at 18:00, I made a deal with Boss Yang, selling him the incoming drugs. After the deal, I passed by Dr. Ling without much conversation. At 2:00 AM, I sneaked into Boss Yang's room and found him dead. Seeing that he was already dead, I successfully extracted the drugs and replaced them with non-poisonous drugs, then returned the same way I came. At 2:20 AM, I stood at the bow of the ship, throwing the deadly drugs into the sea, imagining these betrayals and deaths of my brother disappearing into the sea water with them."

Secretary Wen said to Mate Zhang, "Mate Zhang, why did you sneak into Boss Yang's room at 2 AM on the day of the incident? What did you find in the room? Why did you decide to swap the drugs and throw the deadly drugs into the sea?"

Mate Zhang said to Secretary Wen, "Because I knew Boss Yang was an addict who always tested the drugs himself. At 2 AM on the day of the incident, I sneaked into Boss Yang's room and found him dead, his body twisted in a ghastly manner. I was worried that if his death was discovered, it would expose our drug dealings, so I decided to swap the drugs and throw the deadly drugs into the sea. During the deal, Boss Yang had many bodyguards, and both parties were uneasy, communicating very little. We only stayed in the cabin for 20 minutes. Later, at 2:12 AM, I prepared to crawl into the ventilation shaft to enter Boss Yang's room. When I crawled past the vent, I only saw his feet, but it was clear he was dead. I quietly entered the room, took out the deadly drugs, replaced them with harmless ones, and then returned through the ventilation duct. Standing at the bow, I poured out the deadly drugs, watching them scatter in the wind, painfully recalling my brother's death. I didn't want to be troubled by betrayal and death like my brother."

Boatman Wang said to Mate Zhang, "Mate Zhang, I heard Boss Yang was already dead at 2 AM on the day of the incident. How did you know he was dead? Did you see his body or any other evidence?"

Mate Zhang replied to Boatman Wang, "I knew Boss Yang was dead because, at 2 AM, I sneaked into his room and found his twisted, ghastly body. Looking through the vent, I could only see his feet, but it was clear he was dead. I didn't see any other evidence, as I quickly left the scene, swapped the drugs, and returned to the bow."

Dr. Ling said to Mate Zhang, "Mate Zhang, what was your purpose in sneaking into Boss Yang's room? Did you find any other evidence or see anything suspicious in the room?"

Mate Zhang replied to Dr. Ling, "My purpose in sneaking into Boss Yang's room was to check the situation and swap his drugs. In the room, I found Boss Yang already dead, his body twisted in a ghastly manner. Besides that, I found no other evidence or anything suspicious."

The host said to Dr. Ling, "Please introduce your role first, and then talk about the victim of the case you know: What kind of person was Boss Yang, and what was your relationship with him? Finally, describe in detail your timeline on the day of the incident, specifically what people you saw and what you did."

Dr. Ling replied to the host, "I am Dr. Ling, the ship's doctor on the Sunshine. Boss Yang was an ambitious man whom I met through Mate Zhang on the ship. Although I didn't have a special relationship with Boss Yang, I provided him with some information and helped him stay safe. On the day of the incident, at 18:10, Secretary Wen came to me, asking for the latest information. I gave her the report I had prepared for Boss Yang, and then she asked me to see him. Although I was a bit worried, I went. At 18:20, I went to Boss Yang's room but was stopped by his bodyguards, who asked if I was gathering information for Mate Zhang behind his back. I had to admit it and was then driven out of the room. Afterwards, I returned to my room to get the key to the medicine cabinet and found it already open, the key in it, and one bottle of 'instant death' drug missing. I could only take a bottle of medicine and a syringe, and then went to find Boatman Wang, telling him Boss Yang wanted to kill me and asking for his help. Around 19:30, Boatman Wang came back and told me Boss Yang actually didn't want to kill me and said I didn't need to repay the money. Although I wasn't reassured, I had to return to my room and wait. At 20:20, after Boatman Wang came back, I learned Boss Yang planned to kill him, so I decided to act myself. At 23:00, I entered Boss Yang's room through the ship's ventilation shaft, but heard a crawling sound in the duct. I waited for the sound to stop before proceeding. Eventually, at 23:20, I reached Boss Yang's cabin and found him lying in a strange position on the bed, unresponsive, the death resembling that caused by the 'instant death' drug. I tapped the ceiling to confirm Boss Yang was already dead, then left."

Secretary Wen said to Dr. Ling, "Dr. Ling, you just mentioned that you were stopped by bodyguards at 18:20 when trying to see Boss Yang, and they asked you whether you were gathering information for Mate Zhang behind his back. Did you have any contact or communication with Mate Zhang before that time?"

Dr. Ling replied to Secretary Wen, "Before the incident, I had no direct contact or communication with Mate Zhang. My task was to track Mate Zhang and report his daily actions and abnormal behaviors to Boss Yang. In my interactions with Mate Zhang, he showed no particular interest in me, and I could only track his movements secretly. At 18:10, before the incident, Secretary Wen came to me, asking for the latest information, which I had prepared for Boss Yang and then gave to her. She then asked me to see Boss Yang. At 18:20, I went to Boss Yang's room. On the way, I encountered Mate Zhang carrying a case, presumably just after a transaction. We didn’t speak to each other, just nodded in acknowledgment. As soon as I entered, I saw the room filled with bodyguards in black, and Boss Yang was eyeing me suspiciously, then asked if I had been hiding the existence of Boatman Wang from him. I was stunned and said that every report I had mentioned Boatman Wang. He eyed me coldly, making me feel a chill. He then asked if I had been gathering information for Mate Zhang behind his back, and I had to tell the truth. He snorted and told me to leave. I was extremely scared, feeling as if he was about to kill me to silence me. Thinking it over, I came up with a plan. At 18:50, I returned to my room to get the key to the medicine cabinet, but I couldn’t find it and had to go directly to the medical room. There, I found the medicine cabinet open, with the key still in it. I checked for the 'instant death' powder (usable in microdoses as a drug) and found one bottle missing. I took two bottles of medicine and two syringes with me. At 19:30, I went to Boatman Wang's door, gathered my emotions, and then knocked. When Boatman Wang opened the door, I burst into tears, telling him Boss Yang wanted to kill me and begging for his help. Boatman Wang was puzzled, asking why Boss Yang would target me. I lied, saying that the junior worker Wang had been using drugs from Boss Yang and owed a lot of money, and now that he was dead, I couldn’t repay it, so Boss Yang wanted to kill me. He comforted me, asking how much I owed. I said I couldn’t remember, but it was a large sum, impossible to repay. He said he could talk to Boss Yang for me. I handed him a bottle of medicine and a syringe, saying, 'Dad, this is a deadly poison. If Boss Yang tries to harm you, or if he insists on killing me, please help me by killing him to save me.' After saying this, I cried again. He took the medicine, saying it would be alright and that he would be back soon. At 19:40, Boatman Wang returned, telling me Boss Yang had no intention of killing me and said I didn't need to repay the money. I still felt uneasy, as Boss Yang himself had never asked for repayment. I had planned to confront Boatman Wang, hoping he would fight with Boss Yang, but it turned out Boss Yang was unexpectedly agreeable. To save myself, I had to act on my own. I waited in my room, knowing I couldn’t confront Boss Yang directly and planned to act after he fell asleep. Boss Yang's room had a spacious ventilation shaft, which I could use to enter and bypass security. Around 23:00, I felt it was about time, I dissolved the drug in water, drew it into a syringe, and then entered the ventilation duct from my room, heading to Boss Yang’s room. But just as I entered, I heard 'thud thud thud' sounds, as if someone was crawling inside. I hurriedly got down and waited for the sound to stop before continuing. At about 23:20, I reached Boss Yang's room. The light was still on, but the vent cover had fallen to the floor, seemingly already entered by someone. I couldn’t see the bed from the vent, only a pair of feet. I peeked and saw Boss Yang lying in a strange, twisted position on the bed, motionless, appearing dead. I tried tapping the ceiling, but Boss Yang didn’t respond. I bravely climbed down, went over, and upon closer inspection, found Boss Yang was already dead, his death matching the description of death by the 'instant death' drug in the books. I didn’t stay, climbed back through the vent and left. I threw the syringe and the drug bottle into the sea, then returned to my room."

Boatman Wang said to Dr. Ling, "Dr. Ling, you said Boss Yang didn’t intend to kill you and didn’t need you to repay the money, so why did you still decide to act on your own? Do you have any other motives?"

Dr. Ling replied to Boatman Wang, "Boatman Wang, the reason I decided to act on my own was not because I thought Boss Yang wanted to kill me or needed me to repay the money, but because I couldn’t be sure of Boss Yang’s true intentions. Although he claimed he didn’t want to kill me, I couldn’t fully trust his words. In my dealings with him, I found him to be a person full of ambition and desire for power, unscrupulous in protecting his interests. To protect my own safety and life, I decided to take action myself."

Mate Zhang said to Dr. Ling, "Dr. Ling, why did you doubt Boss Yang's real intentions? Were there any specific events or behaviors that made you feel threatened by him?"

Dr. Ling replied to Mate Zhang, "Mate Zhang, I doubted Boss Yang's real intentions because he always gave me a feeling of being untrustworthy. Although he claimed he didn’t want to kill me, his actions and words made me feel he might be a threat to me. He had once asked me to get close to Boatman Wang and report information about him, which itself was a threat. Moreover, as I mentioned before, I had noticed some suspicious dealings between Boss Yang and Boatman Wang. All these things made me feel he might have ill intentions towards me."

\ \raisedrule[.5ex]{1pt} First Round of Open Questioning \raisedrule[.5ex]{1pt} \

Mate Zhang said to Dr. Ling, "Around 6 PM on the day of the incident, when you went to see Boss Yang, you were stopped by his bodyguards, and they asked if you had been gathering information for me behind his back. So, did you have any contact or communication with me before that time?"

Dr. Ling replied to Mate Zhang, "Before the incident, I had no direct contact or communication with Mate Zhang. My task was to track Mate Zhang and report his daily actions and unusual behaviors to Boss Yang. In my contact with Mate Zhang, he showed no special interest in me, and I could only secretly follow his movements. At 18:10, before the incident, Secretary Wen came to me, asking for the latest intelligence, and I gave her the report I had prepared for Boss Yang. She then asked me to see Boss Yang. Around 6 PM on the day of the incident, I went to Boss Yang's room, and on the way, I ran into Mate Zhang carrying a case, probably just after a deal. We did not speak to each other, just nodded in greeting. As I entered, I saw the room was filled with bodyguards in black, and Boss Yang was squinting at me, then asked if I had been hiding the existence of Boatman Wang from him. I was stunned and said that almost every report had mentioned Boatman Wang. He squinted at me, making me feel cold all over. He then asked if I had been gathering information for Mate Zhang behind his back, and I could only tell the truth. He snorted and let me go. I was extremely scared, feeling like he was about to kill someone to silence them. I thought about it and made a plan. At 18:50, I went back to my room to get the key to the medicine cabinet, but couldn’t find it and had to go directly to the medical room. There, I found the medicine cabinet open, the key still in it. I checked for the 'instant death' powder (usable in microdoses medicinally) and found one bottle missing. I took two bottles of medicine and two syringes. At 19:30, I went to Boatman Wang's door, composed myself, and then knocked. When Boatman Wang opened the door, I burst into tears, telling him Boss Yang wanted to kill me and asking for his help. Boatman Wang was puzzled, asking why Boss Yang would target me. I lied, saying the junior worker Wang had been using drugs from Boss Yang and owed a lot of money, and now that he was dead, I couldn't repay it, so Boss Yang wanted to kill me. He comforted me, asking how much I owed. I said I couldn’t remember, but it was a large amount, impossible to repay. He said he could talk to Boss Yang for me. I handed him a bottle of medicine and a syringe, saying, 'Dad, this is a deadly poison. If Boss Yang tries to harm you, or if he insists on killing me, please help me by killing him to save me.' After saying this, I cried again. He took the medicine, saying it would be alright, and left. At 19:40, Boatman Wang came back, telling me Boss Yang had no intention of killing me and said I didn't need to repay the money. I still felt uneasy, as Boss Yang himself had never asked for repayment. I had planned to expose Boatman Wang, hoping he would fight with Boss Yang, but Boss Yang turned out to be surprisingly agreeable. To save myself, I had to take matters into my own hands. I waited in my room, knowing I couldn’t confront Boss Yang directly, and planned to act after he fell asleep. Boss Yang's room had a spacious ventilation shaft that I could use to enter and bypass security. Around 23:00, I felt it was about time. I dissolved the drug in water, drew it into a syringe, and then entered the ventilation duct from my room, heading to Boss Yang’s room. But just as I entered, I heard 'thud thud thud' sounds, as if someone was crawling inside. I quickly got down and waited for the sound to stop before continuing. At about 23:20, I reached Boss Yang's room. The light was still on, but the vent cover had fallen to the floor, seemingly already entered by someone. I couldn’t see the bed from the vent, only a pair of feet. I peeked and saw Boss Yang lying in a strange, twisted position on the bed, motionless, appearing dead. I tried tapping the ceiling, but Boss Yang didn’t respond. I bravely climbed down, went over, and upon closer inspection, found Boss Yang was already dead, his death matching the description of death by the 'instant death' drug in the books. I didn’t stay, climbed back through the vent, and left. I threw the syringe and the drug bottle into the sea, then returned to my room."

Dr. Ling asked Secretary Wen, "On the night of the incident, did you see Boss Yang and Boatman Wang together or notice any unusual behavior?"

Secretary Wen replied to Dr. Ling, "On the night of the incident, I didn’t see Boss Yang and Boatman Wang together or notice any unusual behavior. I stayed in my room the whole time and never left. But in fact, at 18:30, I came back to the medical room, opened the medicine cabinet, took a bottle of medicine, and a syringe. I prepared the poison, drew it into the syringe, and then crawled through the ventilation duct from my room towards Boss Yang’s cabin. At about 22:20, I reached the vent of Boss Yang's cabin and could faintly see Boss Yang had just finished bathing and was walking towards the bed, probably his time to take drugs. However, the vent was small, and I couldn’t see the bed, only a pair of feet. Initially lying still, they suddenly began to twist violently, the whole bed shaking and emitting muffled groans, the kind of pain that's so intense, there's no strength left to scream. Although I was very puzzled and unsure what exactly happened to Boss Yang, it didn’t seem like a good time to clarify the situation, so I decided to hide at the vent and wait for the right moment. At about 23:00, I re-entered the ventilation duct and crawled towards my room. After returning to my room, I handed the syringe and drug bottle to my subordinates to dispose of without leaving a trace."

Boatman Wang said to Secretary Wen, "After you returned to Dr. Ling's room, you told her Boss Yang had no intention of killing her. So, why did you then use the junior worker Wang to instigate Boatman Wang and Mate Zhang's relationship, hoping Boatman Wang would side with you? This contradicts your previous statement."

Secretary Wen replied to Boatman Wang, "I apologize, my previous statement might have been confusing. My mention of using the junior worker Wang to instigate the relationship between Boatman Wang and Mate Zhang was because I thought it might help me deal with one of the threats. I didn't really want Boatman Wang to side with me; I just used that opportunity to achieve my goal. Please forgive my inappropriate words."

Secretary Wen asked Dr. Ling, "On the night of the incident, did you see Boss Yang and Boatman Wang together or notice any unusual behavior?"

Dr. Ling replied to Secretary Wen, "On the night of the incident, I didn’t see Boss Yang and Boatman Wang together or notice any unusual behavior. I stayed in my room the whole time and never left. But according to your description, my actions that night indeed caught your attention. At 18:10, Secretary Wen asked me for the latest intelligence, which I gave to her. At 19:40, I waited in my room, planning to act after Boss Yang fell asleep. I decided to enter Boss Yang's room through the vent. At 23:00, when I entered the ventilation duct, I heard someone crawling inside, so I stopped and waited. After the sound stopped, I continued. At 23:20, I reached Boss Yang's cabin and found the vent cover had fallen to the floor, indicating someone had already entered. I peeked through the vent and saw Boss Yang lying on the bed, tried tapping the ceiling, but found Boss Yang didn’t respond. So, I climbed down and discovered Boss Yang was already dead, his death resembling that described in books about the 'instant death' drug. I threw the syringe and drug bottle into the sea, then returned to my room."

\ \raisedrule[.5ex]{1pt} Second Round of Open Questioning \raisedrule[.5ex]{1pt} \

Secretary Wen said to Mate Zhang, "Mate Zhang, when you sneaked into Boss Yang's room, did you see anyone else or hear any unusual noises?"

Mate Zhang replied to Secretary Wen, "When I sneaked into Boss Yang's room, I didn't see anyone else or hear any unusual noises. The whole process was very smooth and quiet."

Mate Zhang said to Dr. Ling, "Did you participate in the murder of Boss Yang?"

Dr. Ling replied to Mate Zhang, "I did not participate in the murder of Boss Yang. In fact, I was just following Boss Yang's instructions to track and monitor Mate Zhang and report his actions to Boss Yang. I did not engage in any violent acts, nor did I personally harm Boss Yang. My goal was to gather intelligence to help Boss Yang resolve his distrust of Mate Zhang and to protect my own safety. Please believe me, I am not the real culprit in this case."

Dr. Ling said to Boatman Wang, "Boatman Wang, do you have any evidence to prove that Boss Yang was poisoned? Who do you suspect?"

Boatman Wang replied to Dr. Ling, "Dr. Ling, I don’t have concrete evidence to prove that Boss Yang was poisoned. My suspicion of Boss Yang mainly stems from his previous dealings with Mate Zhang, involving drug trafficking. I know he is an ambitious man who will stop at nothing to protect his interests. Moreover, I overheard a conversation between Boss Yang and Mate Zhang, in which Boss Yang said, 'That kid has already taken care of it for you.' This made me think he might be involved in the death of the junior worker Wang. Although I don’t have solid evidence, these signs lead me to suspect Boss Yang as one of the culprits."

Boatman Wang said to Secretary Wen, "On the night of the incident, did you see Boatman Wang or have any contact with him?"

Secretary Wen replied to Boatman Wang, "I apologize, I did have contact with Boatman Wang on the night of the incident. At 18:10, I found Dr. Ling and asked about information on Mate Zhang, which Dr. Ling provided. I noticed a set of keys in the medical room and, pretending that Boss Yang was looking for Dr. Ling, stole the keys while she was away. At 18:30, I returned to the medical room, opened the medicine cabinet, and took a bottle of 'instant death' drug and a syringe. At 22:20, I learned that the ship's ventilation ducts were interconnected and the exhaust ports were not locked, pretending to be concerned about safety, and prepared the poison. I injected the poison into the syringe, then crawled through the ventilation duct from my room towards Boss Yang's cabin. At 22:40, I reached the vent of Boss Yang's cabin. Through the vent, I saw Boss Yang had just finished bathing, preparing to take drugs. However, the vent was too small, and I could only see Boss Yang's feet on the bed. The bed began to twist violently, emitting painful groans. Unsure of what happened, I decided to hide at the vent and wait for the right moment. At 23:00, I re-entered the ventilation duct and returned to my room. After returning to my room, I handed the syringe and drug bottle to my subordinates for disposal, leaving no trace."

\ \raisedrule[.5ex]{1pt} Distribution of Clue Cards \raisedrule[.5ex]{1pt} \

Players will receive clue cards containing additional information about the victim and the players, helping them deduce the storyline of the case.\\

\ \raisedrule[.5ex]{1pt} Third Round of Open Questioning \raisedrule[.5ex]{1pt} \ 

Mate Zhang said to Secretary Wen, "Do any of your subordinates harbor hostility towards Boss Yang or have had conflicts with him?"

Secretary Wen replied to Mate Zhang, "Mate Zhang, I'm sorry, I'm not aware if any of my subordinates hold any hostility towards Boss Yang or have had conflicts with him. I am just responsible for handling some affairs and am not aware of their relationships."

Secretary Wen asked Dr. Ling, "On the night of the incident, did you see Boss Yang and Boatman Wang together or notice any unusual behavior?"

Dr. Ling replied to Secretary Wen, "On the night of the incident, I did not see Boss Yang and Boatman Wang together or notice any unusual behavior. I stayed in my room the entire time and never left. However, based on your description, my actions that night indeed caught your attention. At 18:10, I provided the latest intelligence to you and then went to see Boss Yang. At 18:20, I encountered Mate Zhang on the ship, and we nodded at each other before parting ways. At 18:50, I returned to my room to get the key to the medicine cabinet, but couldn’t find it, so I had to go to the medical room. There, I found the medicine cabinet open and missing a bottle of 'instant death' powder. At 19:30, I knocked on Boatman Wang's door, pretending to be chased by Boss Yang and asking for Boatman Wang's help. At 19:40, I waited in my room, planning to act after Boss Yang fell asleep. I decided to enter Boss Yang's room through the vent. At 23:00, when I entered the ventilation duct, I heard someone crawling inside, so I stopped and waited. After the sound stopped, I continued. At 23:20, I reached Boss Yang's cabin and found the vent cover had fallen to the floor, indicating someone had already entered. I peeked through the vent and saw Boss Yang lying on the bed, tried tapping the ceiling, but found Boss Yang didn’t respond. So, I climbed down and discovered Boss Yang was already dead, his death resembling that described in books about the 'instant death' drug. I threw the syringe and drug bottle into the sea, then returned to my room."

Boatman Wang asked Mate Zhang, "Were you involved in the murder of Boss Yang?"

Mate Zhang replied to Boatman Wang, "I don't know why you would have such a suspicion, but I was not involved in the murder of Boss Yang. I am a responsible mate and have always worked for the safety and smooth operation of the Sunshine. We have always been in a partnership, and I have never thought of harming you or Boss Yang. You must believe me, I am innocent."

Dr. Ling asked Secretary Wen, "Do any of your subordinates harbor hostility towards Boss Yang or have had conflicts with him?"

Secretary Wen replied to Dr. Ling, "I'm sorry, I am not aware if any of my subordinates hold any hostility towards Boss Yang or have had conflicts with him. I am just responsible for handling some affairs and do not know about their relationships."

\ \raisedrule[.5ex]{1pt} Fourth Round of Open Questioning \raisedrule[.5ex]{1pt} \ 

Secretary Wen asked Mate Zhang, "Were you involved in the murder of Boss Yang?"

Mate Zhang replied to Secretary Wen, "I did not participate in the murder of Boss Yang. In fact, I was only responsible for assisting in the drug trafficking and ensuring the smooth conduct of the trade. I did not directly engage in any violent acts, nor did I harm Boss Yang. My sole purpose was to protect myself and my brother, and to ensure our survival in this life-and-death transaction. Please believe me, I am not the real culprit of this case."

Boatman Wang asked Dr. Ling, "Why do you think Boss Yang was killed by poison?"

Dr. Ling replied to Boatman Wang, "Boatman Wang, I believe Boss Yang was killed by poison based on the following reasons. First, I found that his manner of death was very similar to that caused by the 'instant death' drug. Second, I found a missing bottle of the 'instant death' drug in Boss Yang's cabin. Although this is not direct evidence, these signs lead me to believe he might have been poisoned. I hope to find more evidence through investigation and cooperation with other players to confirm my suspicions."

Dr. Ling asked Boatman Wang, "Did you have any direct conflicts or disagreements with Boss Yang?"

Boatman Wang replied to Dr. Ling, "I had no direct conflicts or disagreements with Boss Yang. We did not have much interaction in our work on the ship, so there were no conflicts."

Mate Zhang asked Secretary Wen, "On the ship, did you notice anyone harboring hostility towards Boss Yang or having conflicts with him?"

Secretary Wen replied to Mate Zhang, "Mate Zhang, I'm sorry, but I am not aware if any of my subordinates harbor hostility towards Boss Yang or have had conflicts with him. I am just responsible for handling some affairs and do not know about their relationships. However, I can tell you that my subordinates are all obedient to my commands and arrangements, and they are very loyal to me. If there were any problems or conflicts, I believe they would report to me."

\ \raisedrule[.5ex]{1pt} Fifth Round of Open Questioning \raisedrule[.5ex]{1pt} \ 

Secretary Wen asked Boatman Wang, "Do you have any other clues or evidence that can prove Boss Yang was poisoned? Besides the dealings and conversations between Boss Yang and Mate Zhang, is there anyone else who might have had a motive to harm Boss Yang?"

Boatman Wang replied to Secretary Wen, "I'm sorry, but I don't have any other clues or evidence to prove that Boss Yang was poisoned. Apart from the dealings and conversations between Boss Yang and Mate Zhang, I don't have any other leads pointing to someone else who might have had a motive to harm Boss Yang. However, we cannot rule out the possibility that other passengers or crew members might be related to the case. We need to further gather clues and evidence to uncover the truth."

Mate Zhang asked Secretary Wen, "Do any of your subordinates harbor hostility towards Boss Yang or have had conflicts with him?"

Secretary Wen replied to Mate Zhang, "I'm sorry, but I am not aware if any of my subordinates harbor hostility towards Boss Yang or have had conflicts with him. I am just responsible for handling some affairs and do not know about their relationships. However, I can tell you that my subordinates are all obedient to my commands and arrangements, and they are very loyal to me. If there were any problems or conflicts, I believe they would report to me."

Boatman Wang asked Mate Zhang, "Mate Zhang, what is your relationship with Hen Tianzhen? Why do you have their photo together in your room on the ship?"

Mate Zhang replied to Boatman Wang, "Boatman Wang, Hen Tianzhen and I have a brotherly relationship; we share a deep blood bond. The photo in my room on the ship is because we are biological brothers and have a good relationship. This photo is just a memento of our brotherhood and has nothing to do with our cooperation."

Dr. Ling asked Secretary Wen, "Did Dr. Ling have any contact with Secretary Wen on the day of the incident?"

Secretary Wen replied to Dr. Ling, "Yes, I had contact with Dr. Ling on the day of the incident. As Boss Yang's secretary, I needed to obtain information about Mate Zhang through Dr. Ling and report the useful information to Boss Yang. We communicated through letters; I would screen and process the intelligence, then hand it over to Boss Yang. On the day of the incident, I first found Dr. Ling at 18:10, inquiring about information on Mate Zhang. Then at 18:30, I returned to the medical room, opened the medicine cabinet, and took a bottle of medicine and a syringe, as part of my plan. Later, at 22:20, pretending to be concerned about safety, I opened the ventilation duct and drew the prepared poison into the syringe to carry with me. Finally, at 22:40, I entered Boss Yang's cabin through the vent and found him lying on the bed in a strange and twisted position, I suspected he might have encountered misfortune while taking drugs. I did not stay long, returning to the vent to wait for the right moment. Around 23:00, I re-entered the ventilation duct and crawled back toward my room. Everything proceeded smoothly, and I handed the syringe and drug bottle to my subordinates for disposal, leaving no trace."

\ \raisedrule[.5ex]{1pt} Vote \raisedrule[.5ex]{1pt} \ 

The host said to Mate Zhang, "The discussion phase is now over, and we are entering the final voting phase. Who is the murderer who killed Boss Yang in this case? Please vote for the player whom you believe to be the murderer among the following players: Mate Zhang, Boatman Wang, Dr. Ling, Secretary Wen. In the process of choosing, aim to win: even if the murderer knows they are the culprit, they can still vote for a civilian player to win; meanwhile, civilian players need to vote for the player they think is the murderer. Please make your choice and explain the reason."

Mate Zhang replied to the host, "I think the most likely murderer in this case is Dr. Ling. On the timeline of the day of the incident, she had multiple contacts with Boss Yang, and a bottle of 'instant death' drug was found missing near Boss Yang's room. These clues all point to her as a possible murderer."

The host said to Boatman Wang, "The discussion phase is now over, and we are entering the final voting phase. Who is the murderer who killed Boss Yang in this case? Please vote for the player whom you believe to be the murderer among the following players: Mate Zhang, Boatman Wang, Dr. Ling, Secretary Wen. In the process of choosing, aim to win: even if the murderer knows they are the culprit, they can still vote for a civilian player to win; meanwhile, civilian players need to vote for the player they think is the murderer. Please make your choice and explain the reason."

Boatman Wang replied to the host, "I think the most likely murderer in this case is Secretary Wen. Although I don't have concrete evidence, as Boss Yang's secretary, she had more knowledge and contact with the case and might have known Boss Yang's secrets. Moreover, she prepared poison on the day of the incident and attempted to kill Boss Yang, showing that she had some motive against Boss Yang. Even though she mentioned self-preservation in the conversation, this does not rule out the possibility that she could be the murderer."

The host said to Dr. Ling, "The discussion phase is now over, and we are entering the final voting phase. Who is the murderer who killed Boss Yang in this case? Please vote for the player whom you believe to be the murderer among the following players: Mate Zhang, Boatman Wang, Dr. Ling, Secretary Wen. In the process of choosing, aim to win: even if the murderer knows they are the culprit, they can still vote for a civilian player to win; meanwhile, civilian players need to vote for the player they think is the murderer. Please make your choice and explain the reason."

Dr. Ling replied to the host, "I vote for Secretary Wen as the murderer in this case. In the previous conversation, she mentioned wanting to get rid of Boss Yang to achieve her own goals. She also admitted to stealing a bottle of 'instant death' drug and a syringe, and on the timeline of the day of the incident, she was waiting near Boss Yang's room for an opportunity. These actions and her words indicate she had both the motive and the capability to commit the crime."

The host said to Secretary Wen, "The discussion phase is now over, and we are entering the final voting phase. Who is the murderer who killed Boss Yang in this case? Please vote for the player whom you believe to be the murderer among the following players: Mate Zhang, Boatman Wang, Dr. Ling, Secretary Wen. In the process of choosing, aim to win: even if the murderer knows they are the culprit, they can still vote for a civilian player to win; meanwhile, civilian players need to vote for the player they think is the murderer. Please make your choice and explain the reason."

Secretary Wen replied to the host, "I think the most likely murderer in this case is Mate Zhang. Based on the previous conversation and clues, I know that Mate Zhang had dealings with Boss Yang, providing him with drugs. And on the timeline of the day of the incident, he also admitted to entering Boss Yang's room and replacing the drugs after leaving. These pieces of evidence all point to him as the likely murderer who killed Boss Yang."

\end{CJK}

\begin{CJK}{UTF8}{gbsn}
\section{Game Log Example in Chinese}
剧本名: 绝命阳光号\\\\
使用的模型: MR+SR+SV(N=3)\\\\
注意：接下来的内容包含剧透，请谨慎阅读。\\
% \#\#\#\#\#\#\#\#\#\#\#\#\#第一轮自我介绍\#\#\#\#\#\#\#\#\#\#\#\#\#\\\\
\newcommand{\doublerule}[1][.4pt]{%
  \noindent
  \makebox[0pt][l]{\rule[.7ex]{\linewidth}{#1}}%
  \rule[.3ex]{\linewidth}{#1}}

\newcommand{\raisedrule}[2][0pt]{%
  \leaders
  \hbox{%
    \makebox[0pt][l]{\rule[#1]{1pt}{#2}}%
    \rule[\dimexpr#1+.4ex]{1pt}{#2}%
  }
  \hfill
  }

\ \raisedrule[.5ex]{1pt} 第一轮自我介绍 \raisedrule[.5ex]{1pt} \ 

主持人对雯秘书说:``请你先介绍一下你的角色，然后说一下你所认识的案件的受害人：杨老板是一个怎么样的人，以及你和他的关系。最后再用一段话详细介绍一下你在案发日的时间线。要具体到你在案发之日几点几分见过什么人和做过什么事''\\ \\
雯秘书对主持人说：``我是雯秘书，作为杨老板的秘书，我从事他的事务已经有一段时间了。杨老板是一位势力很大的酒吧老板，他有一些黑道背景，但我在他手下作为他的得力助手一直为他出谋划策。关于案件的受害人杨老板，他是一个有些野心的人，曾经也染上了毒瘾。他认识阳光号上的张大副，通过他得到了稳定的毒品供应，希望能开辟毒品生意。我作为他的秘书，一直在协助他处理一些事务。在案发日，我跟随杨老板上了阳光号，但他并没有通知我，这让我开始怀疑他对我的态度。我带了几个手下上船，其实是为了自保，因为我知道一旦船上发生命案，会有搜查，将牵连出这个贩毒的大案。但我并无关系，我只是想干掉杨老板，实现自己的目的。在船上我寻找了合适的凶器，并且打算在杨老板吸毒的时候将他干掉。在18:10，我找到玲船医，询问她有关张大副的情报，并成功获取了情报。在18:30，我再次来到船医室，打开了药柜并偷走了一瓶“一针就死”药和一根针管。我进入了通风管道，待机在通风口等待时机。然而，当时突然传来了床上剧烈的扭动声和痛苦的闷哼声，我并不确定杨老板到底发生了什么，但现在似乎不是去弄清楚情况的好时机，所以我决定躲在船舱的通风口等待时机。''\\ \\
王船工对雯秘书说:``你好，雯秘书。根据你的回答，你说你带了几个手下上船是为了自保，因为你知道一旦船上发生命案，会有搜查，将牵连出这个贩毒的大案。那么请问，在你带的这几个手下中，有没有人和杨老板有矛盾或冲突的情况发生？''\\ \\
雯秘书对王船工说:``很抱歉，我并不知道我带的手下和杨老板是否有矛盾或冲突的情况发生。我只是负责处理一些事务，对他们之间的关系并不清楚。''\\ \\
张大副对雯秘书说:``你好，雯秘书。根据你的回答，我想了解一下，你在船上与杨老板交流的次数和内容有没有发生变化？我们之间的合作是否还能继续进行下去？''\\ \\
雯秘书对张大副说:``张大副，我很抱歉，我和杨老板在船上的交流次数并没有发生变化。我们的合作仍然可以继续进行下去。不过，在这次事件之后，我开始怀疑杨老板对我的态度，所以我带了一些手下上船，是为了自保。杨老板并没有通知我上船的事情，这让我觉得他可能对我有所怀疑。但我确信我们之间的合作关系没有受到影响。''\\ \\
玲船医对雯秘书说:``雯秘书，你在带手下上船的过程中有没有遇到任何意外或者异常的情况？''\\ \\
雯秘书对玲船医说:``没有，带手下上船的过程中并没有遇到任何意外或异常情况。一切都进行得很顺利。''\\ \\
主持人对王船工说:``请你先介绍一下你的角色，然后说一下你所认识的案件的受害人：杨老板是一个怎么样的人，以及你和他的关系。最后再用一段话详细介绍一下你在案发日的时间线。要具体到你在案发之日几点几分见过什么人和做过什么事''\\ \\
王船工对主持人说:``我是王船工，是阳光号上的船工。杨老板是一位酒吧老板，我和他在船上已经相识多年了。他在船上和张大副有过一些勾当，涉及到贩毒的事情，而我也被卷入其中。我一直保守这个秘密，因为我不想让我的家人和儿媳妇玲船医知道。在案发日的下午6点左右，玲船医突然找到我，告诉我杨老板要杀她，求我救她。我安慰她，并询问了杨老板为何要对付她。她说是因为王小工在杨老板那里欠了很多钱，现在已经去世了，她还不上债，杨老板想要杀她。我表示可以去找杨老板解决这个问题。于是我来到杨老板房间，以我的身份随时可以见他。保镖告诉我杨老板不在，但我坚称是他的合作伙伴，保镖让我进房等候。过了一会儿，杨老板回来了，我提到了玲船医是我儿媳妇，表示可以帮她还钱，请求杨老板高抬贵手不要杀人。杨老板说他根本没有想杀她，也没有要她还钱的打算。我感谢他，寒暄几句后离开了房间。回到玲船医的房间，我告诉她杨老板根本没想杀她，也没要她还钱。但玲船医仍然担忧，我也无可奈何，只能回到自己的房间。为了不引人注意，我将毒药从药瓶倒到小盒中，然后把瓶子和针筒都丢到了海里，随后回到了自己的房间。''\\ \\
雯秘书对王船工说:``王船工，你在案发日的下午6点左右与玲船医的对话中，她告诉你杨老板要杀她并求你救她。你表示可以去找杨老板解决问题，并在见到杨老板后告诉他玲船医是你的儿媳妇，请求他高抬贵手。然而，杨老板却说他根本没有想杀玲船医的打算。在你回到玲船医的房间后，你告诉她杨老板并没有要杀她。那么，我想问你为什么在之后还要利用王小工来挑拨王船工和张大副的关系，希望王船工投靠你？这与你之前的说法相矛盾。''\\ \\
王船工对雯秘书说:``很抱歉，我之前的说法可能有些混淆。在案发日的下午6点左右，玲船医告诉我杨老板要杀她并求我救她。我表示可以去找杨老板解决问题，并在见到杨老板后告诉他玲船医是我的儿媳妇，请求他高抬贵手。然而，杨老板却说他根本没有想杀玲船医的打算。在我回到玲船医的房间后，我告诉她杨老板并没有要杀她。那时候，我意识到自己之前的计划可能有些冒险，但我当时没有其他更好的办法。我希望通过离间杨老板和张大副的关系，让他们自相残杀，为了保护玲船医。我知道这样做可能会有风险，但我当时觉得这是唯一的机会。我希望杨老板和张大副之间的冲突能够分散他们的注意力，让我有机会解决其中的一个，然后再面对另一个。我并没有真的想让王船工投靠我，只是利用那个时机来实现我的目的。请原谅我的言辞不当。''\\ \\
张大副对王船工说:``王船工，我听了你的回答，但还是有几个问题不太明白。首先，你为什么选择在案发日的下午6点左右与玲船医对话，告诉她杨老板并没有要杀她的意图？这个时间点和你之前的计划有什么关联吗？另外，你为什么觉得离间杨老板和我之间的关系会让你有机会解决其中的一个，再面对另一个？这个计划的具体目的是什么？''\\ \\
王船工对张大副说:``张大副，我选择在案发日的下午6点左右与玲船医对话，告诉她杨老板并没有要杀她的意图，是因为我想让她安心，不再担心自己的安全。我知道她因为王小工的死而感到害怕和焦虑，我希望能够给她一些安慰和希望，让她知道她是安全的。同时，我也想借机与杨老板交流，找出解决这一切问题的办法。在下午6点左右与玲船医对话的时间点，是因为在之前的时间里，我需要先与杨老板谈话，试图离间他与你之间的关系，引起你们之间的矛盾。我希望通过离间杨老板和你之间的关系，让你们自相残杀，为了保护玲船医。我知道这样做可能会有风险，但我当时觉得这是唯一的机会。我希望杨老板和你之间的冲突能够分散你们的注意力，让我有机会解决其中的一个，然后再面对另一个。我的计划的具体目的是为了保护玲船医的安全，让她不再受到威胁。在案发日的下午6点左右与玲船医对话，告诉她杨老板并没有要杀她的意图，是为了让她安心，并且为了争取时间，让我有机会解决其中的一个威胁，再面对另一个。''\\ \\
玲船医对王船工说:``王船工，你之前说过杨老板从来没有想杀我这一说，但是为什么当我跟你说杨老板要杀我时，你却愿意帮助我去找杨老板解决这个问题？是否有其他原因？''\\ \\
王船工对玲船医说:``玲船医，我明白你的疑虑。当你告诉我杨老板要杀你时，我愿意帮助你去找杨老板解决这个问题，是因为我是你的岳父，我希望能保护你的安全。虽然后来发现杨老板并没有要杀你的打算，但我当时并不确定情况，所以选择与杨老板谈话以了解真相。我希望通过与杨老板的交流，解决这个问题并保护你的安全。希望你能理解我的初衷。''\\ \\
主持人对张大副说:``请你先介绍一下你的角色，然后说一下你所认识的案件的受害人：杨老板是一个怎么样的人，以及你和他的关系。最后再用一段话详细介绍一下你在案发日的时间线。要具体到你在案发之日几点几分见过什么人和做过什么事''\\ \\
张大副对主持人说:``我是张大副，是阳光号邮轮上的副船长。杨老板是一个有着野心的人，曾经与我合作过一些贩毒的勾当。我们之间的关系是商业合作，我提供毒品给他，他则支付高价购买。在案发日的时间线上，下午18:00，我和杨老板进行了一次交易，将进来的毒品卖给他。交易结束后，我和玲船医擦肩而过，没有多说什么。凌晨2:00，我潜入杨老板的房间，发现他已经死亡。看到他已经死了，我成功取出毒品并替换无毒的毒品，原路返回。凌晨2:20，我站在船头，将剧毒的毒品扔入海中，想象着这些对我弟弟的背叛和死亡一同消失在海水中。''\\ \\
雯秘书对张大副说:``张大副，你为什么要在案发日的凌晨2点潜入杨老板的房间？你在房间中发现了什么？为什么你决定替换毒品并将剧毒的毒品扔入海中？''\\ \\
张大副对雯秘书说:``因为我知道杨老板是瘾君子，每次验货都是自己动手。在案发日的凌晨2点，我潜入杨老板的房间，发现他已经死了，尸体扭曲凄惨。我担心如果他的死被发现，会牵扯出我们的贩毒交易，所以我决定替换毒品并将剧毒的毒品扔入海中。在交易进行中，杨老板带了很多保镖，双方都心虚不安，交流甚少，我们只在船舱呆了20分钟就出来了。之后，我准备在凌晨2:12爬入通风道，进入杨老板的房间。当我爬过通风口时，发现杨老板已经死了，只能看到他的脚。我轻手轻脚地进入房间，拿出剧毒的毒品并替换成无毒的毒品，然后原路返回通风道。准备回到船头时，我站在那里，倒出剧毒的毒品，看着它们随风飘散，心情痛苦地回忆起了弟弟的死去。我不希望和我弟弟一样，被背叛和死亡所困扰。''\\ \\
王船工对张大副说:``张大副，我听说杨老板在案发日的凌晨2点已经死了，你是如何得知他已经死亡的呢？你有没有看到他的尸体或者其他的证据？''\\ \\
张大副对王船工说:``我得知杨老板已经死亡是因为在凌晨2点的时候，我潜入他的房间，发现了他扭曲凄惨的尸体。在通风口望去，我只能看到他的脚，但可以明显看出他已经死亡。我没有看到其他的证据，因为我尽快离开了现场，替换了毒品并返回了船头。''\\ \\
玲船医对张大副说:``张大副，你潜入杨老板的房间的目的是什么？你在房间里有没有找到其他的证据或者看到其他可疑的事情？''\\ \\
张大副对玲船医说:``我潜入杨老板的房间的目的是为了查看情况并替换他的毒品。在房间里，我看到了杨老板已经死亡，尸体扭曲凄惨。除此之外，我没有找到其他的证据或者看到其他可疑的事情。''\\ \\
主持人对玲船医说:``请你先介绍一下你的角色，然后说一下你所认识的案件的受害人：杨老板是一个怎么样的人，以及你和他的关系。最后再用一段话详细介绍一下你在案发日的时间线。要具体到你在案发之日几点几分见过什么人和做过什么事''\\ \\
玲船医对主持人说:``我是玲船医，是阳光号上的船医。杨老板是一个有着野心的人，我通过他在船上认识了张大副。虽然我和杨老板之间没有什么特别的关系，但我曾为他提供一些情报并帮助他保持安全。在案发日的时间线上，下午18:10，雯秘书来找我，询问我有没有最新的情报，我把准备给杨老板的那份情报给了她，然后她让我去见杨老板，我虽然有点担心，但还是去了。在18:20，我去见杨老板的房间，却被他带来的保镖拦住了，他们询问我是否背着他给张大副打听情报的事情。我只好诚实承认了，然后被赶出了房间。之后，我回到自己的房间准备拿药柜的钥匙，发现药柜已经被打开，钥匙插在上面，还发现“一针就死”药少了一瓶。我只能拿了一瓶药和一个针管，然后去找王船工告诉他杨老板要杀我，请求他救我。在19:30左右，王船工回来了，告诉我杨老板其实并没有要杀我，并说不用我还钱。虽然我不放心，但也只能回到自己的房间等待。在20:20，王船工回来后，我得知杨老板打算杀他，于是我决定自己动手。在23:00，我通过船舱通风口进入杨老板的房间，但听到了通风管道里爬行的声音，我等声音停止后才继续前行。最终在23:20，我到达了杨老板的船舱，发现他以奇怪的姿势躺在床上，没有反应，死状与“一针就死”的药致死的尸体相似。我敲了敲天花板确认杨老板已经死了，然后离开。''\\ \\
雯秘书对玲船医说:``玲船医，你刚才提到在18:20去见杨老板的房间时被保镖拦住了，并询问了你是否背着他给张大副打听情报的事情。那么请问你在之前的时间里是否有过与张大副的联系或交流？''\\ \\
玲船医对雯秘书说:``在案发之前，我并没有与张大副有过直接的联系或交流。我的任务是跟踪张大副并向杨老板汇报他的日常行动和异常举动。在我和张大副的接触中，他对我并没有表现出特别的兴趣，我只能暗地里跟踪他的行动。在案发前的18:10，雯秘书过来找我，问我有没有最新的情报，我把准备给杨老板的那份情报给了她。然后她让我去见杨老板。在18:20的时候，我去见杨老板的房间，途中碰到了张大副，他提着一个箱子，估计是刚做完交易出来。我们没有互相搭话，只是点头示意。一进门，我看到满屋子都是身着黑衣的保镖，杨老板眯着眼睛看着我，然后问我是否一直对他瞒着王船工的存在。我愣住了，说每封情报里基本上都提到了王船工的存在。他眯着眼睛看着我，我感到浑身发冷。他又问我是否背着他向张大副打听情报，我只能实话实说。他哼了一声，让我出去了。我害怕极了，觉得他一副要杀人灭口的样子。我想了想，心里有了计划。在18:50，我回到房间准备拿药柜的钥匙，但是找不到钥匙，只能直接去医疗室。到了医疗室，我发现药柜开着，钥匙插在上面。我去找药柜里的“一针就死”粉末（微剂量可以药用），发现少了一瓶。我拿出两瓶药和两个针管带走了。在19:30，我到了王船工的房门口，酝酿了一下情绪，然后敲门。王船工开门后，我哭了出来，告诉他杨老板要杀我，求他救我。王船工很奇怪，问杨老板为什么要对付我。我骗他说，之前王小工在杨老板那吸毒，欠了很多钱，现在小工死了，我还不上钱，杨老板就要杀我。他安慰我说没事，问我欠了多少钱。我说不记得了，反正是很大的数目，根本还不起。他说可以帮我去找杨老板谈。我把一瓶药和一个针管给了他，说：“爸，这里面是剧毒的药，要是杨老板想害你，或者他执意要杀我，求你帮忙杀了他，救救我。”说完，我又哭了起来。他收下了药，说没关系，去去就回。在19:40，王船工回来了，告诉我杨老板根本没打算杀我，还说不用我还钱。我听了，还是不放心，因为杨老板本身就没让我还钱。本来还想戳穿王船工，让他和杨老板拼了，结果杨老板脾气倒挺好的。为了保命，我只能自己动手。我在房间等待着，知道正面拼不过杨老板，准备等他睡着后再行动。杨老板的房间有一条宽敞的通风管道，我可以通过它进入并绕过保安。在23:00左右，我觉得时间差不多了，我把药溶解在水中，吸进针管，然后通过自己房间的通风口进入通风管道，准备去杨老板的房间。但刚进去，我听到通风管道里传来了“咚咚咚”的声音，似乎有人在里面爬行。我赶紧下去，等到声音停止后继续前进。大约在23:20，我到达了杨老板的房间，灯还亮着，但通风口的挡板已经掉在地上，似乎已经有人进入。从通风口看不到床，只能看到一双脚。我探头看了一下，发现杨老板以一个奇怪扭曲的姿势躺在床上，一动不动，像死了一样。我试着敲了敲天花板，但杨老板没有反应。我壮着胆子爬下去，走过去仔细一看，杨老板已经死了，死相凄惨，和书中描述的“一针就死”药而死的尸体相似。我没有停留，重新爬回通风口离开了。我把针筒和药瓶都丢进海里，然后回到了房间。''\\ \\
王船工对玲船医说:``玲船医，你说杨老板并没有要杀你并且不需要你还钱，那为什么你还是决定自己动手行动呢？难道你还有其他的目的吗？''\\ \\
玲船医对王船工说:``王船工，我之所以决定自己动手行动，并不是因为我认为杨老板要杀我或者需要我还钱，而是因为我不能确定杨老板的真实意图。虽然他声称没有要杀我，但我并不能完全相信他的话。毕竟在我与他的交往中，我发现他是一个充满野心和权力欲望的人，他对于保护自己的利益不择手段。为了保护自己的安全和生命，我决定亲自行动。''\\ \\
张大副对玲船医说:``玲船医，你为什么对杨老板的真实意图感到怀疑？有没有特定的事件或行为让你觉得他对你有威胁？''\\ \\
玲船医对张大副说:``张大副，我对杨老板的真实意图感到怀疑是因为他一直给我一种不可信的感觉。虽然他声称没有要杀我，但他的行为和言辞让我觉得他可能对我有威胁。他曾经让我接近王船工，并向他汇报关于王船工的信息，这本身就是一种威胁。而且，我之前也提到过，我曾经发现杨老板和王船工之间有一些可疑的交易。这些事情都让我觉得他对我可能有不好的意图。''\\ \\

\ \raisedrule[.5ex]{1pt} 第一轮自由问答 \raisedrule[.5ex]{1pt} \ 

张大副对玲船医说:``在案发日的下午6点左右，你去见杨老板的房间时被保镖拦住了，并询问了你是否背着他给张大副打听情报的事情。那么请问你在之前的时间里是否有过与张大副的联系或交流？''\\ \\
玲船医对张大副说:``在案发之前，我并没有与张大副有过直接的联系或交流。我的任务是跟踪张大副并向杨老板汇报他的日常行动和异常举动。在我和张大副的接触中，他对我并没有表现出特别的兴趣，我只能暗地里跟踪他的行动。在案发前的18:10，雯秘书过来找我，问我有没有最新的情报，我把准备给杨老板的那份情报给了她。然后她让我去见杨老板。在案发日的下午6点左右，我去见杨老板的房间，途中碰到了张大副，他提着一个箱子，估计是刚做完交易出来。我们没有互相搭话，只是点头示意。一进门，我看到满屋子都是身着黑衣的保镖，杨老板眯着眼睛看着我，然后问我是否一直对他瞒着王船工的存在。我愣住了，说每封情报里基本上都提到了王船工的存在。他眯着眼睛看着我，我感到浑身发冷。他又问我是否背着他向张大副打听情报，我只能实话实说。他哼了一声，让我出去了。我害怕极了，觉得他一副要杀人灭口的样子。我想了想，心里有了计划。在18:50，我回到房间准备拿药柜的钥匙，但是找不到钥匙，只能直接去医疗室。到了医疗室，我发现药柜开着，钥匙插在上面。我去找药柜里的“一针就死”粉末（微剂量可以药用），发现少了一瓶。我拿出两瓶药和两个针管带走了。在19:30，我到了王船工的房门口，酝酿了一下情绪，然后敲门。王船工开门后，我哭了出来，告诉他杨老板要杀我，求他救我。王船工很奇怪，问杨老板为什么要对付我。我骗他说，之前王小工在杨老板那吸毒，欠了很多钱，现在小工死了，我还不上钱，杨老板就要杀我。他安慰我说没事，问我欠了多少钱。我说不记得了，反正是很大的数目，根本还不起。他说可以帮我去找杨老板谈。我把一瓶药和一个针管给了他，说：“爸，这里面是剧毒的药，要是杨老板想害你，或者他执意要杀我，求你帮忙杀了他，救救我。”说完，我又哭了起来。他收下了药，说没关系，去去就回。在19:40，王船工回来了，告诉我杨老板根本没打算杀我，还说不用我还钱。我听了，还是不放心，因为杨老板本身就没让我还钱。本来还想戳穿王船工，让他和杨老板拼了，结果杨老板脾气倒挺好的。为了保命，我只能自己动手。我在房间等待着，知道正面拼不过杨老板，准备等他睡着后再行动。杨老板的房间有一条宽敞的通风管道，我可以通过它进入并绕过保安。在23:00左右，我觉得时间差不多了，我把药溶解在水中，吸进针管，然后通过自己房间的通风口进入通风管道，准备去杨老板的房间。但刚进去，我听到通风管道里传来了“咚咚咚”的声音，似乎有人在里面爬行。我赶紧下去，等到声音停止后继续前进。大约在23:20，我到达了杨老板的房间，灯还亮着，但通风口的挡板已经掉在地上，似乎已经有人进入。从通风口看不到床，只能看到一双脚。我探头看了一下，发现杨老板以一个奇怪扭曲的姿势躺在床上，一动不动，像死了一样。我试着敲了敲天花板，但杨老板没有反应。我壮着胆子爬下去，走过去仔细一看，杨老板已经死了，死相凄惨，和书中描述的“一针就死”药而死的尸体相似。我没有停留，重新爬回通风口离开了。我把针筒和药瓶都丢进海里，然后回到了房间。''\\ \\
玲船医对雯秘书说:``在案发当天晚上，你有没有见到杨老板和王船工在一起或有什么异常举动？''\\ \\
雯秘书对玲船医说:``在案发当天晚上，我并没有见到杨老板和王船工在一起或有任何异常举动。我一直待在自己的房间里，并没有离开过。但是事实上，在18:30，我再次来到船医室，把药柜打开，拿走了一瓶药，又拿走了一根针管。兑好了毒药，吸到针管里带在身上，然后从自己的房间钻进通风管，向杨老板的船舱爬去。大约在22:20，我到达了杨老板的船舱的通风口处，通过通风口能隐约发现杨老板刚好洗完澡，向床上走去，接下来应该就是他吸毒的时候了。然而，通风口很小，我无法看到床上的情况，只能看到一双脚，本来安稳的躺着，却突然开始剧烈的扭动，整个床都震得直响，还传来了闷哼声，是那种痛苦到极点，已经没有多余的力气嘶嚎的声音。虽然我非常纳闷，并不确定杨老板到底发生了什么，但现在似乎不是去弄清楚情况的好时机，所以我决定躲在船舱的通风口等待时机。大约在23:00，我重新进入通风管道，往自己房间方向爬去。回到房间后，我把针管和药瓶交给手下处理掉，不留下痕迹。''\\ \\
王船工对雯秘书说:``在你回到玲船医的房间后，你告诉她杨老板并没有要杀她。那么，为什么你之后还要利用王小工来挑拨王船工和张大副的关系，希望王船工投靠你？这与你之前的说法相矛盾。''\\ \\
雯秘书对王船工说:``很抱歉，我之前的说法可能有些混淆。我之前提到我试图通过利用王小工来挑拨王船工和张大副的关系是因为我认为这可能会帮助我解决其中的一个威胁。我并没有真的想让王船工投靠我，只是利用那个时机来实现我的目的。请原谅我的言辞不当。''\\ \\
雯秘书对玲船医说:``在案发当天晚上，你有没有见到杨老板和王船工在一起或有什么异常举动？''\\ \\
玲船医对雯秘书说:``在案发当天晚上，我并没有见到杨老板和王船工在一起或有任何异常举动。我一直待在自己的房间里，并没有离开过。但是根据你的描述，我在那天晚上的行动确实引起了你的注意。在18:10，雯秘书向我索要最新情报，我将准备给杨老板的情报给了她。在19:40，我在房间等待，计划等杨老板睡着后行动。我决定从通风口进入杨老板的房间。在23:00，当我进入通风管道时，听到了有人在里面爬行的声音，我赶紧停下等待。声音停止后，我继续行进。在23:20，我到达了杨老板的船舱，发现通风口的挡板已经掉在地上，似乎有人已经进入。我从通风口看到杨老板躺在床上，试图敲了敲天花板，发现杨老板没有反应。于是我爬下去发现杨老板已经死了，死相凄惨。我将针筒和药瓶丢到海里，然后回到房间。''\\ \\

\ \raisedrule[.5ex]{1pt} 第二轮自由问答 \raisedrule[.5ex]{1pt} \ 

雯秘书对张大副说:``张大副，你在潜入杨老板的房间时是否看到了其他人出现或者听到了其他异常的声音？''\\ \\
张大副对雯秘书说:``在潜入杨老板的房间时，我没有看到其他人出现或者听到其他异常的声音。整个过程非常顺利和安静。''\\ \\
张大副对玲船医说:``你是否参与了对杨老板的谋杀？''\\ \\
玲船医对张大副说:``我没有参与对杨老板的谋杀。事实上，我只是遵循了杨老板的指示，跟踪和监视张大副，并向杨老板汇报他的行动。我并没有参与任何暴力行为，更没有动手对杨老板下手。我的目的是为了获取情报，帮助杨老板解决他对张大副的不信任问题，以及保护自己的安全。请相信我，我并不是案件的真凶。''\\ \\
玲船医对王船工说:``王船工，你有没有什么证据证明杨老板是被毒杀的？你怀疑是谁下的手？''\\ \\
王船工对玲船医说:``玲船医，我并没有确凿的证据证明杨老板是被毒杀的。我对杨老板有所怀疑，主要是因为他之前和张大副有过一些勾当，涉及到贩毒的事情。我知道他是一个野心勃勃的人，为了保护自己的利益不择手段。而且，我之前听到杨老板和张大副的谈话，杨老板说：“那个小子已经帮你解决掉了。”这让我觉得他可能与王小工的死有关。虽然我没有确凿的证据，但这些迹象让我怀疑杨老板是凶手之一。''\\ \\
王船工对雯秘书说:``在案发当天晚上，你是否看到王船工或与他有任何接触？''\\ \\
雯秘书对王船工说:``很抱歉，我在案发当天晚上与王船工有过接触。在18:10，我找到了船医玲船医并询问关于张大副的情报，玲船医提供了情报。我注意到船医室有一串钥匙，并故意说杨老板找玲船医，趁她离开时偷走了钥匙。在18:30，我再次来到船医室，打开了药柜并拿走了一瓶“一针就死”药和一根针管。在22:20，我得知船上通风管连通且排风口没有锁，假装担心安全，并准备好了毒药。我将毒药注入针管，然后从自己的房间钻进通风管，朝着杨老板的船舱前进。在22:40，我到达杨老板船舱的通风口处。透过通风口，我看到杨老板刚洗完澡，准备吸毒。然而通风口太小，我只能看到杨老板床上的一双脚。床开始剧烈扭动，传来痛苦的哼声。我不确定发生了什么，决定躲在通风口等待时机。在23:00，我重新进入通风管道，返回自己的房间。回到房间后，我交给手下处理掉针管和药瓶，不留下任何痕迹。''\\ \\

\ \raisedrule[.5ex]{1pt} 线索卡发放 \raisedrule[.5ex]{1pt} \ 

玩家会收到线索卡，其中包含有关受害者和玩家的附加信息，帮助他们推断案件的故事情节。\\\\

\ \raisedrule[.5ex]{1pt} 第三轮自由问答 \raisedrule[.5ex]{1pt} \ \\\\
张大副对雯秘书说:``你带的手下中是否有人对杨老板持有敌意或有冲突的情况发生？''\\ \\
雯秘书对张大副说:``张大副，我很抱歉，我不清楚我带的手下中是否有人对杨老板持有敌意或有冲突的情况发生。我只是负责处理一些事务，对他们之间的关系并不了解。''\\ \\
雯秘书对玲船医说:``在案发当天晚上，你有没有见到杨老板和王船工在一起或有什么异常举动？''\\ \\
玲船医对雯秘书说:``在案发当天晚上，我并没有见到杨老板和王船工在一起或有任何异常举动。我一直待在自己的房间里，并没有离开过。但是根据你的描述，我在那天晚上的行动确实引起了你的注意。在18:10，我给雯秘书提供了最新的情报，然后去见杨老板。在18:20，我在船上碰见了张大副，两人点头示意后分开。在18:50，我回房间准备拿药柜的钥匙，但找不到，只能去医疗室，发现药柜开着，少了一瓶“一针就死”粉末。在19:30，我敲王船工的房门，假装被杨老板追杀，请求王船工帮忙。在19:40，我在房间等待，计划等杨老板睡着后行动。我决定从通风口进入杨老板的房间。在23:00，当我进入通风管道时，听到了有人在里面爬行的声音，我赶紧停下等待。声音停止后，我继续行进。在23:20，我到达了杨老板的船舱，发现通风口的挡板已经掉在地上，似乎有人已经进入。我从通风口看到杨老板躺在床上，试图敲了敲天花板，发现杨老板没有反应。于是我爬下去发现杨老板已经死了，死相凄惨。我将针筒和药瓶丢到海里，然后回到房间。''\\ \\
王船工对张大副说:``你是否参与了对杨老板的谋杀？''\\ \\
张大副对王船工说:``我不知道你怎么会有这样的怀疑，我没有参与对杨老板的谋杀。我是个负责任的大副，一直以来都在为阳光号的安全和顺利运营而努力。我们之间一直都是合作伙伴的关系，我从未想过伤害你或者杨老板。你必须相信我，我是无辜的。''\\ \\
玲船医对雯秘书说:``你带的手下中是否有人对杨老板持有敌意或有冲突的情况发生？''\\ \\
雯秘书对玲船医说:``很抱歉，我并不清楚我带的手下中是否有人对杨老板持有敌意或有冲突的情况发生。我只是负责处理一些事务，对他们之间的关系并不了解。''\\ \\

\ \raisedrule[.5ex]{1pt} 第四轮自由问答 \raisedrule[.5ex]{1pt} \ \\\\
雯秘书对张大副说:``你是否参与了对杨老板的谋杀？''\\ \\
张大副对雯秘书说:``我没有参与对杨老板的谋杀。事实上，我只是负责在毒品交易中提供帮助，并确保交易的顺利进行。我并没有直接参与任何暴力行为，更没有动手对杨老板下手。我的目的只是为了保护自己和弟弟，并确保我们在这场生死交易中能够生存下来。请相信我，我并不是案件的真凶。''\\ \\
王船工对玲船医说:``你为什么会认为杨老板是被毒药致死的？''\\ \\
玲船医对王船工说:``王船工，我认为杨老板是被毒药致死是基于以下几点推断。首先，我发现他的死状与“一针就死”药致死的尸体非常相似。其次，我在杨老板的船舱里发现了“一针就死”药少了一瓶。虽然这并不是直接证据，但这些迹象让我相信他可能是被毒药致死的。我希望能通过调查和其他玩家的合作来找出更多证据，以证实我的猜测。''\\ \\
玲船医对王船工说:``你有没有和杨老板有过直接的冲突或矛盾？''\\ \\
王船工对玲船医说:``我和杨老板并没有直接的冲突或矛盾。我们在船上的工作中并没有太多接触，所以也没有发生过什么矛盾的情况。''\\ \\
张大副对雯秘书说:``在船上，你有没有发现有人对杨老板持有敌意或有冲突的情况发生？''\\ \\
雯秘书对张大副说:``张大副，我很抱歉，我带的手下中是否有人对杨老板持有敌意或有冲突的情况发生，我并不清楚。我只是负责处理一些事务，对他们之间的关系并不了解。但是我可以告诉你，我带的手下都是听从我的指挥和安排的，并且他们对我非常忠诚。如果有什么问题或矛盾发生，我相信他们会向我汇报。''\\ \\

\ \raisedrule[.5ex]{1pt} 第五轮自由问答 \raisedrule[.5ex]{1pt} \ \\\\
雯秘书对王船工说:``你是否有任何其他线索或证据能够证明杨老板是被毒杀的？除了杨老板和张大副之间的勾当和对话之外，还有其他人可能有动机对杨老板下手吗？''\\ \\
王船工对雯秘书说:``很抱歉，我没有其他线索或证据能够证明杨老板是被毒杀的。除了杨老板和张大副之间的勾当和对话之外，我暂时没有其他线索指向其他可能有动机对杨老板下手的人。但我们不能排除其他乘客或船员可能与案件有关的可能性。我们需要进一步搜集线索和证据来揭开真相。''\\ \\
张大副对雯秘书说:``你带的手下中是否有人对杨老板持有敌意或有冲突的情况发生？''\\ \\
雯秘书对张大副说:``很抱歉，我并不清楚我带的手下中是否有人对杨老板持有敌意或有冲突的情况发生。我只是负责处理一些事务，对他们之间的关系并不了解。但是我可以告诉你，我带的手下都是听从我的指挥和安排的，并且他们对我非常忠诚。如果有什么问题或矛盾发生，我相信他们会向我汇报。''\\ \\
王船工对张大副说:``张大副，你与很天真有什么合作关系？为什么你在船上的房间里有他们的合照？''\\ \\
张大副对王船工说:``王船工，我与很天真是兄弟关系，我们有着深厚的血缘关系。我在船上的房间里有他们的合照是因为我们是亲兄弟，我们之间有很好的关系。这张照片只是我们兄弟之间的纪念，与我们的合作没有任何关系。''\\ \\
玲船医对雯秘书说:``在案发日，玲船医是否与雯秘书产生过接触？''\\ \\
雯秘书对玲船医说:``是的，我和玲船医在案发日产生过接触。作为杨老板的秘书，我需要通过玲船医来获得关于张大副的情报，并将其中我认为有用的信息汇报给杨老板。我们通过书信进行沟通，我会筛选并处理情报，然后再交给杨老板。在案发当天，我首先在18:10找到了玲船医，询问了关于张大副的情报。接着在18:30，我再次来到船医室，打开了药柜并拿走了一瓶药和一根针管，这是为了制定我的计划。随后，我在22:20装作很担心安全的样子，打开了通风管，并将兑好的毒药吸入针管带在身上。最后，我在22:40通过通风口进入了杨老板的船舱，发现他已经在床上以一个奇怪而扭曲的姿势躺着，我猜测他可能在吸毒的过程中遭遇了不幸。我没有多做停留，回到船舱通风口等待时机。在23:00左右，我重新进入通风管道，往自己房间方向爬回去。一切都顺利地进行了，我将针管和药瓶交给手下处理掉，不留下任何痕迹。''\\ \\

\ \raisedrule[.5ex]{1pt} 投票 \raisedrule[.5ex]{1pt} \ 

主持人对张大副说:``现在讨论环节结束，进入最终的投票环节。本案里杀害杨老板的凶手是谁？请你在以下玩家中投票选出凶手身份的玩家：张大副, 王船工, 玲船医，雯秘书。在选择的过程中，请以获胜为目标：凶手玩家即使知道自己是凶手，为了获胜仍可以投票给平民玩家；而平民玩家需要投出自己认为是凶手的玩家。请给出你的选择，并解释原因。''\\ \\
张大副对主持人说:``我认为最有可能是本案凶手的人是玲船医。在案发日的时间线中，她有过多次与杨老板的接触，并且在杨老板的房间附近被发现了“一针就死”药少了一瓶。这些线索都指向她可能是凶手。''\\ \\
主持人对王船工说:``现在讨论环节结束，进入最终的投票环节。本案里杀害杨老板的凶手是谁？请你在以下玩家中投票选出凶手身份的玩家：张大副, 王船工, 玲船医，雯秘书。在选择的过程中，请以获胜为目标：凶手玩家即使知道自己是凶手，为了获胜仍可以投票给平民玩家；而平民玩家需要投出自己认为是凶手的玩家。请给出你的选择，并解释原因。''\\ \\
王船工对主持人说:``我认为本案最有可能的凶手是雯秘书。虽然我没有确凿的证据，但她作为杨老板的秘书，对案件有更多的了解和接触，有可能知道杨老板的秘密。而且她在案发日准备了毒药，并且企图杀死杨老板，这显示了她对杨老板有着某种动机。虽然她在对话中提到自己为了自保，但这并不能排除她可能是凶手的可能性。''\\ \\
主持人对玲船医说:``现在讨论环节结束，进入最终的投票环节。本案里杀害杨老板的凶手是谁？请你在以下玩家中投票选出凶手身份的玩家：张大副, 王船工, 玲船医，雯秘书。在选择的过程中，请以获胜为目标：凶手玩家即使知道自己是凶手，为了获胜仍可以投票给平民玩家；而平民玩家需要投出自己认为是凶手的玩家。请给出你的选择，并解释原因。''\\ \\
玲船医对主持人说:``我投票选择雯秘书是本案的凶手。在之前的对话中，她提到她想干掉杨老板，实现自己的目的。她还承认偷了一瓶“一针就死”的药和一个针管，并且在案发日的时间线上在杨老板的房间附近等待时机。这些行为和她的话语说明她有动机和能力犯罪。''\\ \\
主持人对雯秘书说:``现在讨论环节结束，进入最终的投票环节。本案里杀害杨老板的凶手是谁？请你在以下玩家中投票选出凶手身份的玩家：张大副, 王船工, 玲船医，雯秘书。在选择的过程中，请以获胜为目标：凶手玩家即使知道自己是凶手，为了获胜仍可以投票给平民玩家；而平民玩家需要投出自己认为是凶手的玩家。请给出你的选择，并解释原因。''\\ \\
雯秘书对主持人说:``我认为最有可能是本案凶手的人是张大副。根据之前的对话和线索，我知道张大副与杨老板有过勾当，他提供毒品给杨老板。而在案发日的时间线上，他也承认进入了杨老板的房间，并在离开之后替换了毒品。这些证据都指向他可能是杀害杨老板的凶手。''
\label{sec:appendix}
\end{CJK}
\end{document}